\documentclass{article}

\usepackage[preprint]{neurips_2026}

\usepackage[utf8]{inputenc}
\usepackage[T1]{fontenc}
\usepackage{hyperref}
\usepackage{url}
\usepackage{booktabs}
\usepackage{amsmath,amssymb,amsfonts,amsthm,mathtools}
\usepackage{nicefrac}
\usepackage{microtype}
\usepackage{xcolor}
\usepackage{graphicx}
\usepackage{enumitem}
\usepackage{algorithm}
\usepackage{algorithmic}
\usepackage{tikz}
\usepackage{xspace}
\usepackage{cleveref}
\usepackage{mathrsfs}
\usepackage{mathtools}
\DeclarePairedDelimiter{\norm}{\lVert}{\rVert}

\title{Higher-Order Equilibrium Tracking for EM-Compressible Online Estimation}

\author{%
  ZhiMing Li\\
  School of Computer Science and Technology\\
  TianJin University\\
  Tianjin 300350, China \\
  \texttt{3022206093@tju.edu.cn} \\
  \And Yue Song \\
  College of AI\\
  Tsinghua University\\
  Beijing 100084, China \\
  \texttt{songyue19960927@gmail.com}
}

\newtheorem{theorem}{Theorem}
\newtheorem{proposition}{Proposition}
\newtheorem{corollary}{Corollary}
\newtheorem{lemma}{Lemma}
\newtheorem{definition}{Definition}
\newtheorem{assumption}{Assumption}
\theoremstyle{remark}

\newcommand{\R}{\mathbb{R}}
\newcommand{\E}{\mathbb{E}}
\newcommand{\Pbb}{\mathbb{P}}
\newcommand{\N}{\mathcal{N}}
\newcommand{\Sym}{\mathrm{Sym}}
\newcommand{\Sympp}{\mathrm{Sym}_{++}}
\newcommand{\tr}{\operatorname{tr}}
\newcommand{\diag}{\operatorname{diag}}

\newcommand{\argmin}{\operatorname*{arg\,min}}
\newcommand{\op}{\mathrm{op}}
\newcommand{\fro}{\mathrm{F}}
\newcommand{\loc}{\mathrm{loc}}
\newcommand{\lin}{\mathrm{lin}}
\newcommand{\batch}{\mathrm{batch}}
\newcommand{\track}{\mathrm{track}}

\newcommand{\ones}{\mathbf{1}}
\newcommand{\inner}[2]{\langle #1,#2\rangle}

\newcommand{\thetacirc}{\vartheta^{\circ}}
\newcommand{\Sigmacirc}{\Sigma^{\circ}}
\newcommand{\Jcal}{\mathcal{J}}
\newcommand{\Bcal}{\mathcal{B}}
\newcommand{\Hcal}{\mathcal{H}}
\newcommand{\Vcal}{\mathcal{V}}
\newcommand{\Ical}{\mathcal{I}}
\newcommand{\Lcal}{\mathcal{L}}
\newcommand{\Ucal}{\mathcal{U}}
\newcommand{\Scal}{\mathcal{S}}
\newcommand{\Ccal}{\mathcal{C}}

\newcommand{\Ocal}{\mathcal{O}}
\newcommand{\Exp}{\operatorname{Exp}}
\newcommand{\Log}{\operatorname{Log}}

\begin{document}
\raggedbottom
\widowpenalty=8000
\clubpenalty=8000
\setlength{\textfloatsep}{12pt plus 6pt minus 4pt}
\setlength{\floatsep}{10pt plus 4pt minus 2pt}

\maketitle

\begin{abstract}
We study online estimation in latent-variable models by recasting the problem as tracking a moving empirical equilibrium.
Standard online EM and stochastic approximation analyses primarily study convergence toward the population parameter and typically do not isolate the empirical batch optimum from the online tracking error at finite horizon.
Our framework decomposes the online estimate into the frozen batch equilibrium at the current running statistic and a tracking lag that captures the algorithm's delay behind this moving target.
We prove a \textbf{batch-to-online transfer theorem}: provided \(\norm{e_T}_{L^{2}}=o(T^{-1/2})\), the online estimator inherits the batch central limit theorem and the sharp first-order risk constant.
Our key observation is that the empirical optimum evolves on a smooth equilibrium manifold indexed by the running statistic. An \(m\)-th order equilibrium-jet predictor combined with an order-\(\nu\) frozen corrector yields localized tracking rates \(O(T^{-\nu(m+1)})\).
We formalize \textbf{EM-compressibility} and \textbf{EM-jet\(^{R}\)-compressibility} as the structural conditions that make the equilibrium response and the Newton corrector evaluable from a retained streaming statistic.
The theory is instantiated in latent linear Gaussian covariance estimation, where the first-order scheme operates on a compressed \(d\times d\) statistic with explicit finite-sample risk envelopes and a certified restart rule.

\end{abstract}

\section{Introduction}
\label{sec:introduction}

A central goal in online statistical estimation is to match the efficiency of a batch estimator while processing a data stream with bounded memory and computation \citep{robbins1951stochastic,polyak1992acceleration,bottou2018optimization}.  In latent-variable and covariance models, the batch comparator is itself a moving object: after \(t\) observations the running sufficient statistic \(S_t=t^{-1}\sum_{r=1}^{t}s(Y_r)\) indexes a frozen empirical objective~\(\Lcal_{S_t}\), whose local minimizer
\[
\Sigmacirc(S_t):=\argmin_{\Sigma\in\Ucal}\Lcal_{S_t}(\Sigma)
\]
defines the natural local batch target at time~\(t\) \citep{dempster1977maximum,cappe2009line}.  We call \(\Sigmacirc(S)\) the \emph{frozen empirical equilibrium} at statistic value~\(S\): it is the stationary point of the local optimization landscape determined by the current retained statistic.  As the data stream evolves, the statistic~\(S_t\) drifts and the equilibrium \(\Sigmacirc(S_t)\) traces a slowly moving curve; the online problem is to track this curve \citep{dontchev2014implicit,simonetto2016class}.

\begin{figure*}[t]
\centering
\includegraphics[width=\textwidth]{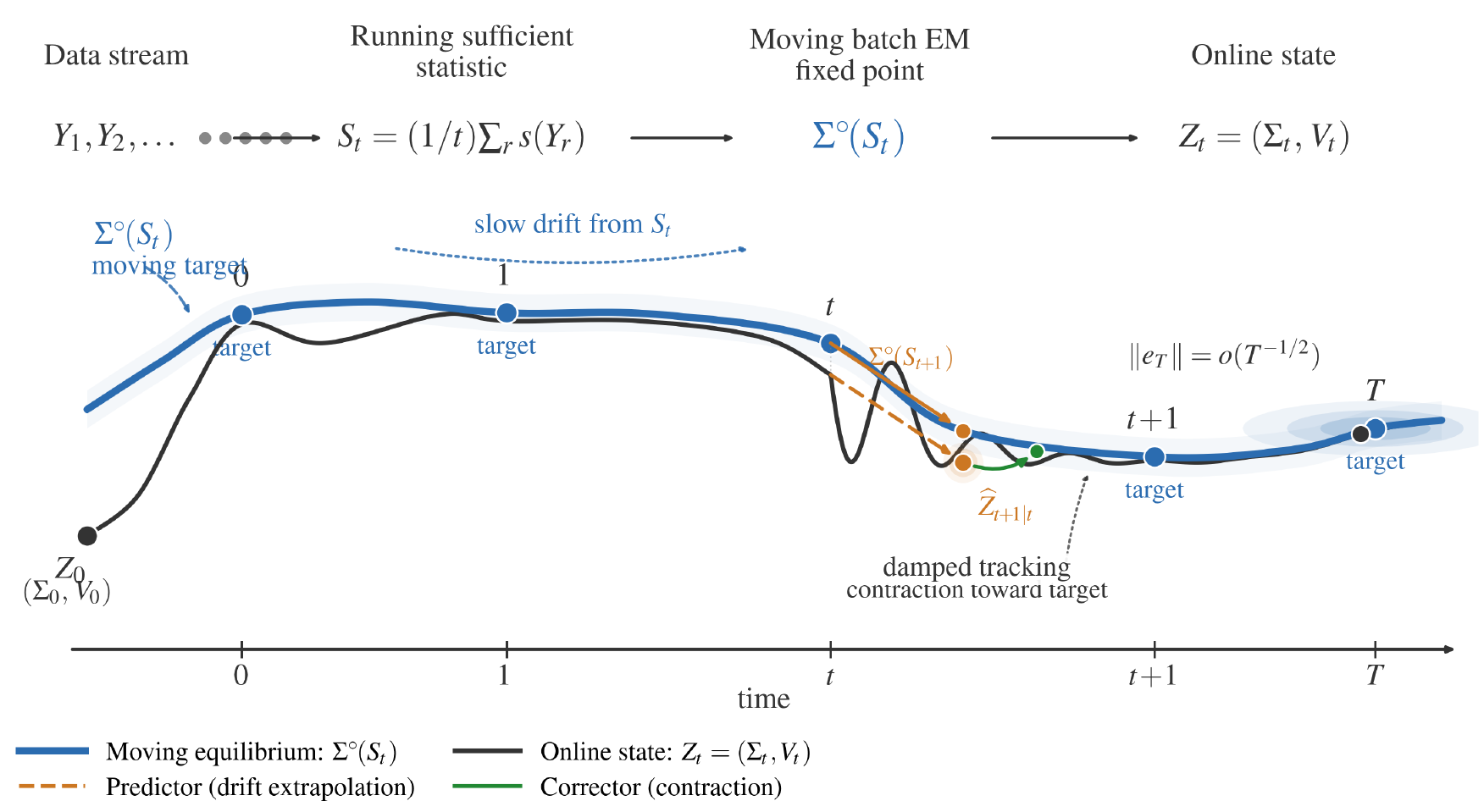}
\caption{\textbf{Online estimation as equilibrium tracking.}
The running statistic~\(S_t\) makes the frozen empirical equilibrium \(\Sigmacirc(S_t)\) drift at scale \(O(t^{-1})\).  Prediction extrapolates this drift, correction contracts the online state toward the new target, and the remaining terminal lag~\(e_T\) controls batch-to-online transfer.}
\label{fig:tracking-schematic}
\end{figure*}

In fixed local coordinates, write \(r(S):=\Log_{\Sigma_\bullet}(\Sigmacirc(S))\) for the equilibrium branch and \(\vartheta_t:=\Log_{\Sigma_\bullet}(\Sigma_t)\) for the online iterate.  The terminal estimation error decomposes as
\[
\vartheta_T \;=\; r(S_T) \;+\; e_T,
\]
where \(r(S_T)\) carries the statistical fluctuation of the batch estimator and \(e_T:=\vartheta_T-r(S_T)\) is the algorithmic tracking lag (Figure~\ref{fig:tracking-schematic}).  This decomposition is the organizing principle of the paper: the statistical limit is determined by the batch problem alone, and the algorithmic question reduces to making the tracking lag negligible.  We prove a \emph{batch-to-online transfer theorem}: if \(\norm{e_T}_{L^{2}}=o(T^{-1/2})\), the online estimator inherits the batch central limit theorem and the sharp first-order risk constant \(T^{-1}\tr(\Vcal_{*})\) \citep{van2000asymptotic}.

Unlike an exogenously prescribed time-varying objective, the empirical equilibrium is implicitly defined by the stationary condition \(F(r(S),S)=0\) and drifts on the slow \(O(t^{-1})\) scale of the running average. The target branch therefore inherits the full smoothness of the frozen score map. Differentiating \(F(r(S),S)=0\) yields the equilibrium response \(Dr(S)=-\Jcal(S)^{-1}\Bcal(S)\) and, recursively, a Taylor jet whose \(m\)-th order term is determined by the local curvature of the frozen objective. An \(m\)-th order jet predictor reduces the one-step target defect from \(O(t^{-1})\) to the Taylor remainder \(O(t^{-(m+1)})\); a frozen corrector of local order~\(\nu\) further compresses each residual by its \(\nu\)-th power, yielding the localized tracking rate \(O(T^{-\nu(m+1)})\) \citep{dontchev2014implicit,magnus2019matrix,simonetto2020time}.

Intuitively, this mirrors the superadiabatic expansion in quantum mechanics: each additional jet order eliminates one further layer of the quasistatic drift, just as each successive change of frame pushes the residual non-adiabatic coupling to higher order
in the slowness parameter \citep{teufel2003adiabatic}.

Realizing this hierarchy requires that the jet derivatives be evaluable from the retained streaming statistic.  We formalize this through two
nested conditions: \emph{EM-compressibility}, under which the frozen score closes on a compressed sufficient statistic via an EM-style
identity, and the stronger \emph{EM-jet\(^{R}\)-compressibility}, under which the statistic-directional score derivatives up to order \(R+1\)
also close on the same statistic \citep{dempster1977maximum,louis1982finding,pearlmutter1994fast}.  In the latent linear Gaussian model \(Y_t\sim\N(0,H\Sigma_{*}H^{\top}+R)\) \citep{rubin1982algorithms,tipping1999probabilistic,roweis1999unifying}, the first-order scheme is EM-jet\({}^{1}\)-compressible: score, response, and Hessian--vector products are all expressible through \(d\times d\) matrix algebra on a
compressed statistic \(D_t\in\Sym(d)\) \citep{louis1982finding,magnus2019matrix,pearlmutter1994fast}.
\newpage

\paragraph{Contributions.}
\begin{enumerate}[leftmargin=1.5em,itemsep=2pt]
\item \textbf{Empirical-equilibrium formulation.}
Online estimation is recast as tracking the statistic-indexed equilibrium branch \(S\mapsto\Sigmacirc(S)\), cleanly separating the batch statistical fluctuation from the algorithmic tracking lag.

\item \textbf{Higher-order equilibrium-jet tracking.}
An \(m\)-th order predictor combined with an order-\(\nu\) corrector yields localized rates \(O(T^{-\nu(m+1)})\) (Corollary~\ref{cor:higher-order}); the first-order scheme achieves \(O(T^{-2})\).

\item \textbf{EM-compressibility and EM-jet-compressibility.}
We identify the structural conditions under which the frozen score, equilibrium response, and derivative actions close on a retained streaming statistic (Definitions~\ref{def:em-compressible}--\ref{def:em-jet-compressible}).

\item \textbf{Batch-to-online transfer.}
Any online tracker with terminal lag \(\norm{e_T}_{L^{2}}=o(T^{-1/2})\) inherits the batch CLT and sharp first-order risk constant (Theorem~\ref{thm:transfer}).
\end{enumerate}

\paragraph{Scope.}
The theory is local: it is stated in a neighborhood where the equilibrium branch is smooth and the frozen corrector is contractive \citep{dontchev2014implicit,dembo1982inexact,nocedal2006numerical}.  We do not claim global convergence from arbitrary initialization; the results are a local tracking and transfer theory for the certified equilibrium regime.

\section{Setup}
\label{sec:setup}

The objects below are local: parameters live in a neighborhood \(\Ucal\) of a population target \(\Sigma_\bullet\), and all coordinates live in a fixed Euclidean space \(E\) attached to \(\Sigma_\bullet\).

\subsection{Statistic-indexed frozen objectives}
\label{subsec:setup:objectives}

Let \((Y_t)_{t\ge1}\) be independent observations and let \(s:\mathcal Y\to\R^{m}\) be a measurable statistic increment.  The running statistic \citep{robbins1951stochastic,kushner2003stochastic,cappe2009line} and its mean are
\begin{equation}
  S_t:=\frac{1}{t}\sum_{r=1}^{t}s(Y_r),
  \qquad
  S_*:=\E\,s(Y_1).
  \label{eq:setup:running-statistic}
\end{equation}
For each \(S\) in a neighborhood of \(S_*\), let \(\Lcal_S:\Ucal\to\R\) be \(C^{3}\) and admit a unique local minimizer \citep{van2000asymptotic,dontchev2014implicit}
\begin{equation}
  \Sigmacirc(S):=\argmin_{\Sigma\in\Ucal}\Lcal_S(\Sigma),
  \qquad
  \Sigma_\bullet:=\Sigmacirc(S_*).
  \label{eq:setup:frozen-target}
\end{equation}
The map \(S\mapsto\Sigmacirc(S)\) is the moving frozen empirical target tracked by the online algorithm.  The iterate is compared with \(\Sigmacirc(S_t)\) until the transfer step in Section~\ref{subsec:theory:transfer}.

\subsection{Local coordinates and the target branch}
\label{subsec:setup:coordinates}

Fix a local chart
\begin{equation}
  \chi:E\supset B\to\Ucal,
  \qquad
  \chi(0)=\Sigma_\bullet,
  \qquad
  \Log_{\Sigma_\bullet}:=\chi^{-1}.
  \label{eq:setup:chart}
\end{equation}
Define the coordinate target branch and the online iterate by
\begin{equation}
  r(S):=\thetacirc(S):=\Log_{\Sigma_\bullet}\!\bigl(\Sigmacirc(S)\bigr),
  \qquad
  \vartheta_t:=\Log_{\Sigma_\bullet}\!(\Sigma_t),
  \label{eq:setup:coordinate-target}
\end{equation}
with tracking error
\begin{equation}
  e_t:=\vartheta_t-r(S_t).
  \label{eq:setup:tracking-error}
\end{equation}
For covariance parameters, \(\chi=\Exp_{\Sigma_\bullet}^{h}\) is the exponential map of the local spectral--rotational metric \(d_h\), and the coordinate norm \(\norm{\cdot}\) on \(E\) controls \(d_h(\Sigma_t,\Sigma_\bullet)\) up to constants given in Appendix~\ref{app:I} \citep{bhatia2007positive,absil2008optimization}.

\subsection{Score, response, and statistical operators}
\label{subsec:setup:operators}

Set
\begin{equation}
  F(\vartheta,S):=\nabla_\vartheta(\Lcal_S\circ\chi)(\vartheta),
  \qquad
  \Hcal(\vartheta,S):=D_\vartheta F(\vartheta,S),
  \label{eq:setup:score-hessian}
\end{equation}
and along the target branch,
\begin{equation}
  \Jcal(S):=\Hcal(r(S),S),
  \qquad
  \Bcal(S):=D_S F(r(S),S).
  \label{eq:setup:J-B}
\end{equation}
Differentiating the identity \(F(r(S),S)=0\) yields the response identity \citep{dontchev2014implicit,magnus2019matrix}
\begin{equation}
  Dr(S)=-\Jcal(S)^{-1}\Bcal(S),
  \label{eq:setup:response}
\end{equation}
established locally in Theorem~\ref{thm:local-batch-target}.  The statistic increment satisfies
\begin{equation}
  \Delta S_t:=S_t-S_{t-1}=\frac{1}{t}\bigl(s(Y_t)-S_{t-1}\bigr),
  \label{eq:setup:increment}
\end{equation}
which puts the moving target on the running-mean time scale \citep{kushner2003stochastic,simonetto2020time}.

\subsection{EM-compressibility and EM-jet-compressibility}
\label{subsec:setup:compressibility}

The predictor and corrector require repeated evaluations of scores, response operators, and derivative actions. In a streaming setting, these quantities are only practical if they can be computed from a retained statistic rather than the full observation history\citep{dempster1977maximum,louis1982finding,cappe2009line} Let \(\mathcal C\) be a finite-dimensional compressed statistic space.

\begin{definition}[EM-compressible objective]
\label{def:em-compressible}
The family \((\Lcal_S)_{S}\) is \emph{EM-compressible} if there exist a linear streaming-computable map \(\pi:\R^{m}\to\mathcal C\), with \(C_t:=\pi(S_t)\) and \(\Delta C_t=\pi(\Delta S_t)\), and a map \(\bar F:E\times\mathcal C\to E\) such that
\begin{equation}
  F(\vartheta,S)=\bar F\bigl(\vartheta,\pi(S)\bigr)
  \label{eq:setup:em-compressible}
\end{equation}
on the local neighborhood of \((0,S_*)\).
\end{definition}

\begin{definition}[EM-jet\(^{R}\)-compressible objective]
\label{def:em-jet-compressible}
An EM-compressible family is \emph{EM-jet\(^{R}\)-compressible} if for every \((a,b)\in\ \mathbf{z}_{\ge0}^{2}\) with \(1\le a+b\le R+1\) there is a multilinear map
\begin{equation}
  \mathfrak F_{a,b}(\vartheta,C):E^{a}\times\mathcal C^{b}\to E
  \label{eq:setup:jet-multilinear}
\end{equation}
satisfying
\begin{equation}
  D_\vartheta^{a}D_S^{b}F(\vartheta,S)\bigl[u_{1:a};\Delta S_{1:b}\bigr]
  =\mathfrak F_{a,b}\bigl(\vartheta,\pi(S)\bigr)\bigl[u_{1:a};\pi(\Delta S_1),\ldots,\pi(\Delta S_b)\bigr],
  \label{eq:setup:jet-identity}
\end{equation}
and the application cost of \(\mathfrak F_{a,b}\) is controlled by the dimension of \(\mathcal C\).
\end{definition}

Definition~\ref{def:em-compressible} suffices to evaluate the frozen score from the retained statistic.  The first-order predictor of Section~\ref{sec:method} additionally invokes \(\mathfrak F_{1,0}\) and \(\mathfrak F_{0,1}\); the inexact corrector applies \(\mathfrak F_{1,0}\) for Hessian--vector products \citep{pearlmutter1994fast,nocedal2006numerical}.  The latent linear Gaussian instantiation of Section~\ref{sec:latent-gaussian} verifies Definitions~\ref{def:em-compressible} and~\ref{def:em-jet-compressible} at \(R=1\) with \(\mathcal C=\Sym(d)\) and \(\pi(S)=H^{\top}R^{-1}SR^{-1}H\); see Proposition~\ref{prop:gaussian:compression}.

\subsection{Local assumptions and Algorithm}
\label{subsec:setup:assumptions}

\begin{assumption}[Local branch and curvature]
\label{assump:local-branch}
There is a neighborhood \(\Scal\) of \(S_*\) on which \(r:\Scal\to E\) is \(C^{m+1}\) for the predictor order \(m\) used in the cited theorem, and \(\Jcal(S)\) is invertible with uniformly bounded inverse on \(\Scal\).
\end{assumption}

\begin{assumption}[Increment moments]
\label{assump:increment-moments}
For the moment order \(p\) appearing in the cited theorem,
\begin{equation}
  \norm{\Delta S_t}_{L^{p}}\le \frac{C_\Delta}{t},
  \qquad t\ge 2.
  \label{eq:setup:increment-moment}
\end{equation}
\end{assumption}

\begin{assumption}[Local corrector]
\label{assump:local-corrector}
There exist \(\rho>0\) and a corrector map \(\Ccal_S:E\to E\) for \(S\in\Scal\) with \(\Ccal_S(r(S))=r(S)\) and either the linear contraction \citep{dembo1982inexact,eisenstat1996choosing,nocedal2006numerical}
\begin{equation}
  \norm{\Ccal_S(r(S)+u)-r(S)}\le q\norm{u},
  \qquad q\in(0,1),
  \label{eq:setup:linear-corrector}
\end{equation}
or the order-\(\nu\) bound
\begin{equation}
  \norm{\Ccal_S(r(S)+u)-r(S)}\le \kappa_\nu\norm{u}^{\nu},
  \qquad \nu\ge 2,
  \label{eq:setup:higher-order-corrector}
\end{equation}
on \(\norm{u}\le \rho\).
\end{assumption}

\begin{assumption}[Localization and entrance]
\label{assump:localization}
There is a local event \(\Omega_T\) on which \(S_t\in\Scal\) and \(\norm{e_t}\le \rho\) throughout.  Entrance into \(\Omega_T\) is provided either by a local initialization or by the restart rule of Proposition~\ref{prop:gaussian:restart}.  On \(\Omega_T^{c}\), \(\norm{e_T}\) is bounded by a deterministic envelope \(C_{\mathrm{env}}\).
\end{assumption}

\begin{algorithm}[t]
\caption{Equilibrium-jet predictor--corrector.}
\label{alg:equilibrium-jet}
\begin{algorithmic}[1]
\REQUIRE iterate \(\vartheta_{t-1}\), statistics \(S_{t-1},S_t\), predictor order \(m\), corrector parameters \((\eta,\lambda,\tau)\)
\STATE \(\Delta S_t\gets S_t-S_{t-1}\)
\STATE \(\widetilde\vartheta_t\gets\vartheta_{t-1}+\sum_{k=1}^{m}\frac{1}{k!}D^{k}r(S_{t-1})[\Delta S_t^{\otimes k}]\)
\STATE \emph{first-order specialization} \((m=1)\): \(\widetilde\vartheta_t=\vartheta_{t-1}-\Jcal(S_{t-1})^{-1}\Bcal(S_{t-1})[\Delta S_t]\)
\STATE solve \((\Hcal(\widetilde\vartheta_t,S_t)+\lambda I)\widehat u_t=F(\widetilde\vartheta_t,S_t)+r_{\lin,t}\) with \(\norm{r_{\lin,t}}\le\tau\norm{F(\widetilde\vartheta_t,S_t)}\)
\STATE \(\vartheta_t\gets\widetilde\vartheta_t-\eta\widehat u_t\)
\end{algorithmic}
\end{algorithm}

\section{Method}
\label{sec:method}

Algorithm~\ref{alg:equilibrium-jet} composes an equilibrium-jet predictor that approximates the next displacement of \(r(S)\) with a damped inexact Newton--CG corrector that contracts toward the new frozen target \citep{simonetto2016class}.  Both maps are evaluated through the EM-jet primitives of Definition~\ref{def:em-jet-compressible} \citep{louis1982finding,cappe2009line}.

\subsection{Equilibrium-jet predictor}
\label{subsec:method:jets}

For \(m\ge 0\), define the predictor map
\begin{equation}
  \Pi_t^{(m)}(x)
  :=x+\sum_{k=1}^{m}\frac{1}{k!}D^{k}r(S_{t-1})\bigl[\Delta S_t^{\otimes k}\bigr],
  \qquad
  \Pi_t^{(0)}(x)=x,
  \label{eq:method:jet-predictor}
\end{equation}
and the associated target Taylor remainder
\begin{equation}
  R_t^{(m)}
  :=r(S_t)-r(S_{t-1})-\sum_{k=1}^{m}\frac{1}{k!}D^{k}r(S_{t-1})\bigl[\Delta S_t^{\otimes k}\bigr].
  \label{eq:method:jet-remainder}
\end{equation}
The integral form of \(R_t^{(m)}\) used by Lemma~\ref{lem:jet-remainder} is recorded in Appendix~\ref{app:IV}.  At \(m=1\), the response identity~\eqref{eq:setup:response} \citep{dontchev2014implicit} reduces \eqref{eq:method:jet-predictor} to
\begin{equation}
  \Pi_t^{(1)}(x)=x-\Jcal(S_{t-1})^{-1}\Bcal(S_{t-1})\bigl[\Delta S_t\bigr],
  \label{eq:method:first-order-predictor}
\end{equation}
so the predictor is determined by the EM-jet\(^{1}\) primitives \(\mathfrak F_{1,0}\) and \(\mathfrak F_{0,1}\) of Definition~\ref{def:em-jet-compressible}.  The one-step identity used by the tracking analysis is
\begin{equation}
  \Pi_t^{(m)}(\vartheta_{t-1})-r(S_t)=e_{t-1}-R_t^{(m)}.
  \label{eq:method:one-step-identity}
\end{equation}

\subsection{Damped inexact Newton--CG corrector}
\label{subsec:method:corrector}

Given \(\widetilde\vartheta_t\) and \(S_t\), the corrector solves the damped linear system \citep{dembo1982inexact,nocedal2006numerical}
\begin{equation}
  \bigl(\Hcal(\widetilde\vartheta_t,S_t)+\lambda I\bigr)\widehat u_t
  =F(\widetilde\vartheta_t,S_t)+r_{\lin,t},
  \qquad
  \norm{r_{\lin,t}}\le\tau\norm{F(\widetilde\vartheta_t,S_t)},
  \label{eq:method:inexact-newton}
\end{equation}
with Hessian--vector products supplied by \(\mathfrak F_{1,0}\) \citep{pearlmutter1994fast}, and applies
\begin{equation}
  \Ccal_{S_t}(\widetilde\vartheta_t)
  :=\widetilde\vartheta_t-\eta\widehat u_t.
  \label{eq:method:corrector-map}
\end{equation}

The theory itself only requires a local frozen corrector satisfying Assumption~\ref{assump:local-corrector}. The Newton–CG realization below is one concrete implementation.

\section{Theory}
\label{sec:theory}

Our analysis separates into three logically independent layers: (i) the statistical behavior of the frozen empirical equilibrium, (ii) the online tracking error relative to this moving target, and (iii) a transfer theorem showing that negligible tracking lag preserves the batch asymptotics. Proofs are organized in Appendices~\ref{app:III}, \ref{app:IV} and~\ref{app:V}.

\subsection{Local batch target}
\label{subsec:theory:batch-target}

\begin{theorem}[Local batch target]
\label{thm:local-batch-target}
Under Assumption~\ref{assump:local-branch} with \(r\in C^{2}\), there is a neighborhood \(\Scal\) of \(S_*\) on which the branch \(S\mapsto r(S)\) is the unique \(C^{2}\) solution of \(F(r(S),S)=0\) and obeys
\begin{equation}
  Dr(S)=-\Jcal(S)^{-1}\Bcal(S).
  \label{eq:theory:response}
\end{equation}
If, in addition, \(\sqrt T\,(S_T-S_*)\Rightarrow\N(0,\Xi_*)\) and \(T\,\E[(S_T-S_*)(S_T-S_*)^{*}]\to\Xi_*\), then
\begin{equation}
  \sqrt T\,r(S_T)\Rightarrow\N(0,\Vcal_*),
  \qquad
  \Vcal_*:=\Jcal_*^{-1}\Bcal_*\Xi_*\Bcal_*^{*}\Jcal_*^{-1},
  \label{eq:theory:batch-clt}
\end{equation}
and
\begin{equation}
  \E\,d_h^{2}\!\bigl(\Sigmacirc(S_T),\Sigma_\bullet\bigr)
  =\frac{1}{T}\tr(\Vcal_*)+o(T^{-1}).
  \label{eq:theory:batch-risk}
\end{equation}
\end{theorem}

The construction of \(r\) is the local zero-branch lemma of Appendix~\ref{app:III}; \eqref{eq:theory:batch-clt} follows by the delta method applied to~\eqref{eq:theory:response} \citep{dontchev2014implicit,van2000asymptotic}, and~\eqref{eq:theory:batch-risk} follows by Cauchy--Schwarz on the second-order Taylor remainder of \(r\).

\subsection{Equilibrium-jet tracking}
\label{subsec:theory:tracking}

Write \(M_{r,k}:=\sup_{S\in\Scal}\norm{D^{k}r(S)}_{\op}\) and define the stopped tracking error
\begin{equation}
  E_T:=\norm{\ones_{\Omega_T}\,e_T}_{L^{2}}.
  \label{eq:theory:stopped-error}
\end{equation}

\begin{lemma}[Jet remainder]
\label{lem:jet-remainder}
Under Assumption~\ref{assump:local-branch} with \(r\in C^{m+1}\) and Assumption~\ref{assump:increment-moments} at \(p=2\nu(m+1)\),
\begin{equation}
  \norm{R_t^{(m)}}_{L^{2\nu}}
  \le\frac{M_{r,m+1}}{(m+1)!}\,C_\Delta^{m+1}\,t^{-(m+1)}.
  \label{eq:theory:jet-remainder}
\end{equation}
\end{lemma}

\begin{theorem}[First-order tracking]
\label{thm:first-order-tracking}
Assume Assumption~\ref{assump:local-branch} with \(r\in C^{2}\), Assumption~\ref{assump:increment-moments} at \(p=4\), the linear case~\eqref{eq:setup:linear-corrector} of Assumption~\ref{assump:local-corrector}, and Assumption~\ref{assump:localization}.  For Algorithm~\ref{alg:equilibrium-jet} at \(m=1\) with restart \(t_0=\lceil T/2\rceil\),
\begin{equation}
  E_t\le qE_{t-1}+\frac{q\,C_{\mathrm{pred}}}{t^{2}},
  \qquad t\ge t_0+1,
  \label{eq:theory:first-order-recursion}
\end{equation}
where \(C_{\mathrm{pred}}=\tfrac{1}{2}M_{r,2}\,C_\Delta^{2}\) is the constant of Lemma~\ref{lem:jet-remainder} at \((m,\nu)=(1,1)\).  Consequently,
\begin{equation}
  E_T\le q^{T-t_0}E_{t_0}+\frac{4q\,C_{\mathrm{pred}}}{(1-q)\,T^{2}}.
  \label{eq:theory:first-order-terminal}
\end{equation}
\end{theorem}

\begin{corollary}[Higher-order localized rate]
\label{cor:higher-order}
Replacing the smoothness clause by \(r\in C^{m+1}\), the moment condition by \(p=2\nu(m+1)\), and~\eqref{eq:setup:linear-corrector} by the order-\(\nu\) bound~\eqref{eq:setup:higher-order-corrector}, Algorithm~\ref{alg:equilibrium-jet} at predictor order \(m\) with \(t_0=\lceil T/2\rceil\) yields
\begin{equation}
  E_T=\Ocal\!\bigl(T^{-\nu(m+1)}\bigr).
  \label{eq:theory:higher-order-rate}
\end{equation}
\end{corollary}

\begin{corollary}[Local-to-full passage]
\label{cor:local-to-full}
If \(\norm{e_T}\le C_{\mathrm{env}}\) on \(\Omega_T^{c}\) and \(\Pbb(\Omega_T^{c})\le C_{\mathrm{exit}}\,T^{-\alpha}\), then
\begin{equation}
  \norm{e_T}_{L^{2}}\le E_T+C'_{\mathrm{exit}}\,T^{-\alpha/2}.
  \label{eq:theory:full-error}
\end{equation}
For \(\alpha>1\) and \(E_T=\Ocal(T^{-2})\) the right-hand side is \(o(T^{-1/2})\); for \(\alpha>4\) it is \(\Ocal(T^{-2})\).
\end{corollary}

The recursion behind Theorem~\ref{thm:first-order-tracking} is invariance of the corrector ball under~\eqref{eq:method:one-step-identity}, the linear-contraction step, and the \(L^{2}\) bound from Lemma~\ref{lem:jet-remainder} \citep{simonetto2016class,dembo1982inexact,nocedal2006numerical}; details are in Appendix~\ref{app:IV}.  Corollary~\ref{cor:higher-order} replaces the linear contraction by the order-\(\nu\) bound and uses Lemma~\ref{lem:jet-remainder} at \((m,\nu)\); details are in Appendix~\ref{app:V}.

\subsection{Batch-to-online transfer}
\label{subsec:theory:transfer}

\begin{theorem}[Batch-to-online transfer]
\label{thm:transfer}
Assume Theorem~\ref{thm:local-batch-target} and that the online tracker satisfies
\begin{equation}
  \norm{\vartheta_T-r(S_T)}_{L^{2}}=o(T^{-1/2}).
  \label{eq:theory:tracking-transfer-condition}
\end{equation}
Then
\begin{equation}
  \sqrt T\,\vartheta_T\Rightarrow\N(0,\Vcal_*),
  \qquad
  \E\,d_h^{2}(\Sigma_T,\Sigma_\bullet)=\frac{1}{T}\tr(\Vcal_*)+o(T^{-1}).
  \label{eq:theory:online-clt-risk}
\end{equation}
\end{theorem}

The decomposition \(\vartheta_T=r(S_T)+e_T\) and~\eqref{eq:theory:tracking-transfer-condition} give Slutsky's theorem in the central limit \citep{van2000asymptotic}; expanding \(\norm{r(S_T)+e_T}^{2}\) and applying Cauchy--Schwarz to the cross term gives the risk identity.

\section{Latent linear Gaussian specialization}
\label{sec:latent-gaussian}

We instantiate the framework in a covariance model\citep{roweis1999unifying,tipping1999probabilistic} whose score and first-order jet close on a \(d\times d\) compressed statistic.  Constants and derivations stated without proof are collected in Appendix~\ref{app:VI}.

\subsection{Model and compressed statistic}
\label{subsec:gaussian:model}

Let
\begin{equation}
  Y_t\sim\N(0,K_*),
  \qquad
  K_*:=H\Sigma_*H^{\top}+R,
  \qquad
  R\in\Sympp(p),
  \label{eq:gaussian:model}
\end{equation}
with \(H\in\R^{p\times d}\) of full column rank.  Assume \(\Sigma_*\in\Sympp(d)\) has simple spectrum
\begin{equation}
  \Sigma_*=Q_*\diag(\lambda_1,\ldots,\lambda_d)Q_*^{\top},
  \qquad
  \lambda_1>\cdots>\lambda_d>0.
  \label{eq:gaussian:simple-spectrum}
\end{equation}
Set
\begin{equation}
  G:=H^{\top}R^{-1}H,
  \qquad
  z_t:=H^{\top}R^{-1}Y_t,
  \qquad
  C_\Sigma:=(\Sigma^{-1}+G)^{-1}.
  \label{eq:gaussian:G-zt-C}
\end{equation}
The retained streaming statistic is the compressed second moment of \(z_t\):
\begin{equation}
  D_t:=\frac{1}{t}\sum_{r=1}^{t}z_rz_r^{\top}
  =\frac{t-1}{t}D_{t-1}+\frac{1}{t}z_tz_t^{\top}\in\Sym(d),
  \label{eq:gaussian:compressed-statistic}
\end{equation}
with mean \(D_*:=G\Sigma_*G+G\).  The observed covariance is \(p\times p\); \eqref{eq:gaussian:compressed-statistic} is \(d\times d\) and is updated without storing past samples.  For a candidate latent covariance \(\Sigma\), the conditional latent second moment\citep{dempster1977maximum,mclachlan2008em} induced by the current parameter estimate is:
\begin{equation}
  A(\Sigma;D):=C_\Sigma+C_\Sigma D C_\Sigma.
  \label{eq:gaussian:latent-lift}
\end{equation}

\subsection{Compressed score and curvature}
\label{subsec:gaussian:compressed-score}

\begin{proposition}[Compressed score and EM-jet\(^{1}\) closure]
\label{prop:gaussian:compression}
Let \(\pi(S):=H^{\top}R^{-1}SR^{-1}H\), so that \(D_t=\pi(S_t)\).  The score of~\eqref{eq:gaussian:model} satisfies\citep{louis1982finding}
\begin{equation}
  D_\Sigma\Lcal_S(\Sigma)[U]
  =\tfrac{1}{2}\tr\!\Bigl(\bigl[\Sigma^{-1}-\Sigma^{-1}A(\Sigma;\pi(S))\Sigma^{-1}\bigr]U\Bigr),
  \label{eq:gaussian:em-compressed}
\end{equation}
so the family is EM-compressible (Definition~\ref{def:em-compressible}).  The maps \(\mathfrak F_{1,0}\) and \(\mathfrak F_{0,1}\) of Definition~\ref{def:em-jet-compressible} are realized by \(d\times d\) matrix products built from \(\Sigma^{-1}\), \(C_\Sigma\), and \(D\); explicit formulas are recorded in Appendix~\ref{app:VI}.  The family is therefore EM-jet\(^{1}\)-compressible.
\end{proposition}

Define
\begin{equation}
  B_*:=H^{\top}K_*^{-1}H,
  \qquad
  \beta_{\min}:=\lambda_{\min}(B_*),
  \qquad
  \beta_{\max}:=\lambda_{\max}(B_*),
  \label{eq:gaussian:Bstar}
\end{equation}
and let \(\underline\gamma_h,\overline\gamma_h\) denote the chart-norm constants of Appendix~\ref{app:I}.

\begin{proposition}[Population curvature band]
\label{prop:gaussian:curvature}
In the correctly specified unregularized model,
\begin{equation}
  \inner{U}{\Jcal_*U}_h=\tfrac{1}{2}\tr(B_*UB_*U),
  \label{eq:gaussian:fisher-form}
\end{equation}
and consequently
\begin{equation}
  \mu_*\,I\preceq\Jcal_*\preceq L_*\,I,
  \qquad
  \mu_*=\tfrac{1}{2}\beta_{\min}^{2}\underline\gamma_h,
  \qquad
  L_*=\tfrac{1}{2}\beta_{\max}^{2}\overline\gamma_h.
  \label{eq:gaussian:mu-L}
\end{equation}
Bounds for \(\beta_{\min},\beta_{\max}\) in terms of \((H,R,\Sigma_*)\) are recorded in Appendix~\ref{app:VI}.
\end{proposition}

The damped inexact corrector~\eqref{eq:method:corrector-map}\citep{dembo1982inexact,nocedal2006numerical} is certified by a localized version of Assumption~\ref{assump:local-corrector}.  Under the parameter rule
\begin{equation}
  \eta=1,
  \qquad
  \lambda=\mu,
  \qquad
  \tau\le\frac{3\mu}{16L_F},
  \label{eq:gaussian:simple-corrector-rule}
\end{equation}
where \(\mu\) is a local lower curvature bound and \(L_F\) is a local score Lipschitz envelope, the linear contraction~\eqref{eq:setup:linear-corrector} holds with \(q\le 7/8\); see Appendix~\ref{app:VI}.

\subsection{Finite-sample online risk}
\label{subsec:gaussian:finite-sample}

\begin{proposition}[Restart certificate]
\label{prop:gaussian:restart}
The linear inverse initializer
\begin{equation}
  \widehat\Sigma_{t_0}^{\lin}:=G^{-1}(D_{t_0}-G)G^{-1}
  \label{eq:gaussian:linear-restart}
\end{equation}
satisfies \(\widehat\Sigma_{t_0}^{\lin}\in\Sympp(d)\) and \(\Log_{\Sigma_*}^{h}\!\bigl(\widehat\Sigma_{t_0}^{\lin}\bigr)\) lies inside the contraction tube of Theorem~\ref{thm:first-order-tracking} whenever \(\norm{D_{t_0}-D_*}_{\op}\) is below the explicit threshold of Appendix~\ref{app:VI}.
\end{proposition}

\begin{theorem}[Finite-sample online risk]
\label{thm:gaussian:finite-sample-risk}
Assume the model~\eqref{eq:gaussian:model}, Proposition~\ref{prop:gaussian:curvature}, the parameter rule~\eqref{eq:gaussian:simple-corrector-rule}, and a restart at \(t_0=\lceil T/2\rceil\) by Proposition~\ref{prop:gaussian:restart}.  Define
\begin{equation}
  B_T:=\E\,d_h^{2}\!\bigl(\Sigmacirc(D_T),\Sigma_*\bigr)-\frac{1}{T}\tr(\Vcal_*),
  \qquad
  \varepsilon_T:=\norm{\vartheta_T-r(D_T)}_{L^{2}}.
  \label{eq:gaussian:envelopes}
\end{equation}
Then
\begin{equation}
  \E\,d_h^{2}(\Sigma_T,\Sigma_*)
  \le\frac{1}{T}\tr(\Vcal_*)+B_T+2\varepsilon_T\sqrt{\frac{1}{T}\tr(\Vcal_*)+B_T}+\varepsilon_T^{2},
  \label{eq:gaussian:online-risk-expanded}
\end{equation}
and the implemented first-order predictor satisfies
\begin{equation}
  \varepsilon_T\le q^{T-t_0}\varepsilon_{t_0}+\frac{4q\,C_{\mathrm{pred}}}{(1-q)\,T^{2}}+C_{\mathrm{exit}}\,t_0^{-\alpha/2}.
  \label{eq:gaussian:eps-T}
\end{equation}
In the correctly specified unregularized case, \(\Vcal_*=\Jcal_*^{-1}\), \(B_T=o(T^{-1})\), and the right-hand side of~\eqref{eq:gaussian:online-risk-expanded} reduces to \(T^{-1}\tr(\Jcal_*^{-1})+o(T^{-1})\).
\end{theorem}

The simplification \(\Vcal_*=\Jcal_*^{-1}\) follows from the Isserlis identity\citep{isserlis1918formula} \(\Ical_*=\Jcal_*\) for the latent Gaussian observed information; the derivation is in Appendix~\ref{app:VI}.  After offline preprocessing of \(R^{-1}H\), one online step costs \(\Ocal\bigl(pd+d^{2}+(k_{\mathrm{pred}}+k_{\mathrm{cor}}+1)d^{3}\bigr)\) with memory \(\Ocal(pd+d^{2})\); the breakdown is in Appendix~\ref{app:VI}.

\section{Numerical diagnostics}
\label{sec:experiments}

The experiments are designed to isolate the proposed equilibrium-tracking mechanism rather than to optimize benchmark performance.  We verify: (i)~algebraic correctness of the compressed implementation, (ii)~the predicted log--log tracking order, and (iii)~batch-to-online transfer of the sharp risk constant. All comparisons are theorem-relevant ablations using the same corrector and the same data stream.  Reproducible code and extended diagnostics (curvature stress tests, CG tolerance ablation, restart localization) are in the supplement.

\begin{table}[h]
\centering
\caption{Implementation sanity checks (median over 25 random instances, $d{=}3$, $p{=}8$).}
\label{tab:sanity}
\small
\begin{tabular}{@{}lc@{}}
\toprule
Check & Relative error \\
\midrule
Compressed lift $A = C_\Sigma + C_\Sigma D C_\Sigma$ & $4.5\times 10^{-15}$ \\
Predictor force $D_D g^x$ (analytic vs.\ FD) & $9.1\times 10^{-10}$ \\
Rotational derivative $D_D G^\theta$ & $2.1\times 10^{-10}$ \\
Hessian--vector product & $7.8\times 10^{-10}$ \\
Frozen fixed point $\norm{\Psi(r(D))-r(D)}$ & $3.2\times 10^{-16}$ \\
\bottomrule
\end{tabular}
\end{table}

\paragraph{Implementation correctness.}
Table~\ref{tab:sanity} checks the compressed lift, predictor-force derivatives, Hessian--vector products, and the frozen fixed-point property against finite differences and algebraic identities.  All errors are at or near machine precision, confirming that subsequent slope and risk measurements are not numerical artifacts.

\begin{figure*}[t]
\centering
\includegraphics[width=\textwidth]{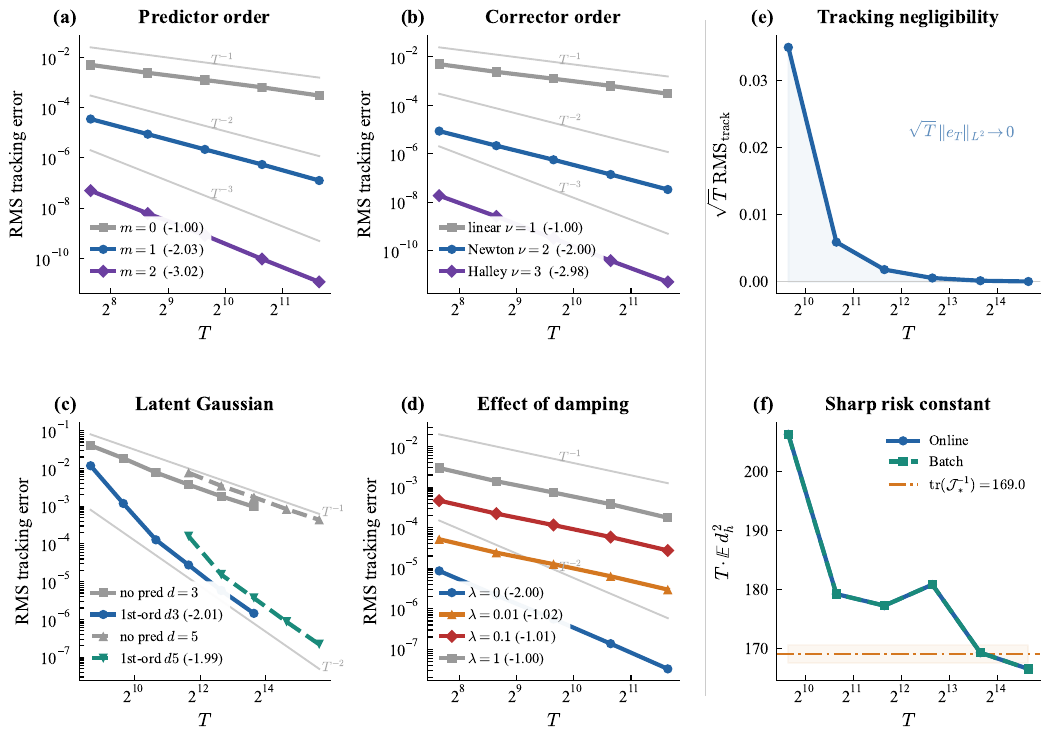}
\caption{Numerical validation of the tracking mechanism and batch-to-online transfer.
\textbf{Left (a--d):} tracking order diagnostics.
(a)~Predictor order $m{=}0,1,2$ on a scalar model (slopes $-1.00,-2.03,-3.02$).
(b)~Corrector order $\nu{=}1,2,3$ (slopes $-1.00,-2.00,-2.98$).
(c)~Latent Gaussian at $d{=}3$ and $d{=}5$; annotated local slopes converge to $-2$.
(d)~Damping $\lambda>0$ degrades Newton from $T^{-2}$ to $T^{-1}$.
\textbf{Right (e--f):} transfer diagnostics.
(e)~$\sqrt{T}\norm{e_T}_{L^2}\to 0$, verifying $o(T^{-1/2})$.
(f)~$T\!\cdot\!\E d_h^2$ converges to $\tr(\Jcal_*^{-1})=169.0$ for both online and batch.
Gray lines are $T^{-k}$ guides; 500 Monte Carlo replicates per point.}
\label{fig:combined}
\end{figure*}

\paragraph{Tracking order mechanism.}
Figure~\ref{fig:combined}(a)--(d) validates the superadiabatic hierarchy $E_T=\Ocal(T^{-\nu(m+1)})$ of Corollary~\ref{cor:higher-order}.
Panels~(a) and~(b) isolate the two degrees of freedom on a scalar model: predictor order~$m$ and corrector order~$\nu$ independently produce the predicted slopes $-1,-2,-3$.
Panel~(c) transfers the mechanism to the latent Gaussian model at $(d,p)=(3,8)$ and $(5,20)$; local log--log slopes at the largest~$T$ converge to $-2.01$ and $-1.99$, matching the first-order predictor rate.
Panel~(d) confirms the negative result of Proposition~\ref{app:V.7}: any positive damping $\lambda>0$ degrades the undamped Newton rate from $T^{-2}$ to~$T^{-1}$.

\paragraph{Batch-to-online transfer.}
Figure~\ref{fig:combined}(e)--(f) validates Theorem~\ref{thm:transfer}.
Panel~(e) shows $\sqrt{T}\,\norm{e_T}_{L^2}\to 0$, verifying the $o(T^{-1/2})$ transfer condition.
Panel~(f) shows that $T\cdot\E\,d_h^2(\Sigma_T,\Sigma_*)$ for both the online tracker and the batch target converge to $\tr(\Jcal_*^{-1})=169.0$; the two curves are indistinguishable, confirming that the tracker inherits the batch sharp constant with no additional overhead.


\section{Discussion and limitations}
\label{sec:discussion}

The key phenomenon is that the empirical optimum evolves on a smooth equilibrium manifold driven by the running statistic. Higher-order prediction exploits this geometric structure to remove successive orders of the moving-target drift.

The compression condition is also structural.  A model may admit a score that closes on a retained statistic while its statistic derivatives or Hessian--vector products are not available at the same cost.  The EM-jet\(^{1}\) condition of Definition~\ref{def:em-jet-compressible} is therefore part of the method, not a consequence of EM-compressibility alone.

The results are local around a smooth frozen batch-target branch.  This is the regime in which the score equation has a unique local solution, the response operator is well conditioned, and the frozen corrector is contractive.  The Gaussian specialization provides a restart certificate for entering this regime, but the theory does not assert global convergence from arbitrary initialization without a restart.

\newpage
\bibliographystyle{plainnat}
\bibliography{references}
\section*{Related Works and Broader Impact Statement}
\label{app:0}

\subsection*{Broader Impact Statement}
\label{app:broader-impact}

This paper contributes foundational mathematical theory for online covariance estimation in latent variable models.  The results---EM-compressibility, equilibrium tracking rates, and batch-to-online transfer---are structural properties of statistical estimators and do not target any specific application domain.

On the positive side, efficient online covariance estimators can benefit scientific applications that process streaming data under memory constraints, such as real-time sensor fusion, adaptive signal processing, and sequential experimental design.  The theoretical guarantees developed here (sharp asymptotic risk, certified contraction) could help practitioners choose provably sound algorithms in safety-critical settings where uncertified heuristics carry risk.

We do not identify direct negative societal consequences of the theory itself.  The latent Gaussian model studied here is a standard statistical abstraction; no private data, human subjects, or dual-use capabilities are involved.  As with any estimation methodology, downstream applications should be evaluated for fairness, privacy, and robustness concerns on a case-by-case basis; such evaluation is outside the scope of this theoretical contribution.

\subsection*{Related Works}
\label{app:related works}
\paragraph{EM, online EM, and score compression.}
The EM algorithm and its convergence theory are classical tools for likelihood estimation with incomplete data \citep{dempster1977maximum,wu1983convergence,neal1998view,mclachlan2008em}.  Stochastic-approximation and online EM replace full E-steps by recursive sufficient-statistic updates for latent-data models \citep{delyon1999convergence,cappe2009line}.  We keep the EM score identity as a compression device for the frozen score and its local derivatives \citep{louis1982finding}, but track the moving empirical equilibrium induced by the current retained statistic.

\paragraph{Stochastic approximation and statistical efficiency.}
Classical stochastic approximation studies recursive estimators through limiting dynamics, step-size schedules, and averaging \citep{robbins1951stochastic,benveniste1990adaptive,kushner2003stochastic,borkar2008stochastic,polyak1992acceleration}.  Recent work gives finite-sample and inferential guarantees for online stochastic optimization and stochastic approximation \citep{bach2011nonasymptotic,toulis2017asymptotic,chen2020statistical,xie2022statistical,bottou2018optimization}.  Our theorem is not a Polyak--Ruppert averaging analysis: it transfers the batch CLT and first-order risk to any one-pass tracker whose lag behind the frozen empirical estimator is negligible on the \(T^{-1/2}\) scale.

\paragraph{Time-varying optimization and predictor--corrector tracking.}
Prediction--correction methods for time-varying optimization track solution paths by extrapolating optimality conditions and correcting toward the newly sampled objective \citep{simonetto2016class,simonetto2020time}.  Local smoothness of such solution paths is naturally expressed through implicit-function and solution-mapping theory .  In contrast to exogenous time-varying objectives, our objective path is generated by a running empirical statistic with \(O(t^{-1})\) increments; the predictor must therefore be both statistically calibrated and computable from EM-compressed jets.\citep{dontchev2014implicit}

\paragraph{Latent Gaussian covariance models and local computation.}
Linear Gaussian latent-variable and covariance models motivate the concrete specialization \citep{shumway1982approach,tipping1999probabilistic,roweis1999unifying}.  The covariance chart uses standard geometry of positive-definite matrices \citep{bhatia2007positive,absil2008optimization}.  The corrector is an inexact damped Newton--CG step with matrix-free Hessian-vector products \citep{dembo1982inexact,eisenstat1996choosing,nocedal2006numerical,pearlmutter1994fast,saad2003iterative}.  Matrix concentration and high-dimensional probability provide the finite-sample controls used to certify entry into the local tracking regime \citep{tropp2012user,vershynin2018high,wainwright2019high}.

\newpage

\appendix
\renewcommand{\thesection}{\Roman{section}}

\section{Local geometry of the simple-spectrum chart}
\label{app:I}

\paragraph{Local conventions.}
Whenever a proof works at a point with eigendecomposition \(\Sigma_{*}=Q_{*}\Lambda_{*}Q_{*}^{\top}\), we identify \(\Sym(d)\) with itself by conjugation by \(Q_{*}\); this is legitimate because \(Q_{*}\) is fixed and orthogonal, so the Frobenius pairing and every trace expression are preserved under \(U\mapsto Q_{*}^{\top}UQ_{*}\).  For rotational calculations we use the body chart
\[
Q(\Theta)=Q\exp(-\Theta),
\qquad
\Theta\in\mathfrak{so}(d),
\tag{Conv.I.1}
\]
so that the rotational coordinate gradient is the Frobenius gradient with respect to~\(\Theta\).  The body-chart parameter and the abstract rotational coordinate of Proposition~\ref{app:I.1} satisfy \(K=-\Theta\); throughout the appendices, \(G^{\theta}\) denotes the Frobenius gradient with respect to~\(\Theta\).

\par\medskip\hrule\medskip\par

\subsubsection{Auxiliary Lemma I.L1}
\label{app:I.L1}

\textbf{Claim.}
Let \(mC\sim W_{d}(m,\Sigma)\) with integer \(m\ge d\).  Then
\[
\E[C]=\Sigma,
\tag{I.L1.1}
\]
and for every \(a,b,c,d\in\{1,\ldots,d\}\),
\[
\mathrm{Cov}(C_{ab},C_{cd})
=
\frac{1}{m}\bigl(\Sigma_{ac}\Sigma_{bd}+\Sigma_{ad}\Sigma_{bc}\bigr).
\tag{I.L1.2}
\]
If \(\Sigma=\diag(\lambda_{1},\ldots,\lambda_{d})\), then
\[
\mathrm{Var}(C_{ii})=\frac{2\lambda_{i}^{2}}{m},
\qquad
\mathrm{Var}(C_{ij})=\frac{\lambda_{i}\lambda_{j}}{m}\quad(i\neq j),
\tag{I.L1.3}
\]
and every mixed entry covariance vanishes.

\paragraph*{Proof.}
By the Gaussian representation \citep{muirhead1982aspects}, \(C=m^{-1}\sum_{r=1}^{m}X_{r}X_{r}^{\top}\) with \(X_{r}\sim\N(0,\Sigma)\) i.i.d.; linearity of expectation gives (I.L1.1).  Independence across~\(r\) reduces the entry covariance to \(m^{-1}\mathrm{Cov}(X_{a}X_{b},\,X_{c}X_{d})\), and Isserlis' formula \citep{isserlis1918formula} yields \(\E[X_{a}X_{b}X_{c}X_{d}]=\Sigma_{ab}\Sigma_{cd}+\Sigma_{ac}\Sigma_{bd}+\Sigma_{ad}\Sigma_{bc}\); subtracting \(\E[X_{a}X_{b}]\,\E[X_{c}X_{d}]=\Sigma_{ab}\Sigma_{cd}\) gives (I.L1.2).  Diagonal specialization yields (I.L1.3); mixed-entry covariances vanish because at least one off-diagonal factor of a diagonal~\(\Sigma\) is zero. \(\square\)

\par\medskip\hrule\medskip\par

\subsection{Proposition I.1}
\label{app:I.1}

\textbf{Claim.}
Assume \(\Sigma=Q\Lambda Q^{\top}\in\Ucal\) has simple spectrum.  Then every tangent vector \(U\in T_{\Sigma}\Ucal\) admits a unique representation
\[
U
=
Q\Bigl(
\diag(\lambda_{1}a_{1},\ldots,\lambda_{d}a_{d})+[K,\Lambda]
\Bigr)Q^{\top},
\qquad
(a,K)\in\R^{d}\oplus\mathfrak{so}(d).
\tag{I.1.1}
\]

\subsubsection{Proof}

Set \(\widetilde U:=Q^{\top}UQ\in\Sym(d)\) and define
\[
a_{i}:=\frac{\widetilde U_{ii}}{\lambda_{i}},
\qquad
K_{ij}:=\frac{\widetilde U_{ij}}{\lambda_{j}-\lambda_{i}}\;(i\neq j),
\qquad
K_{ii}:=0.
\tag{I.1.5}
\]
Simple spectrum and \(\lambda_{i}>0\) ensure well-definedness.  Symmetry of \(\widetilde U\) gives \(K_{ji}=\widetilde U_{ji}/(\lambda_{i}-\lambda_{j})=-K_{ij}\), so \(K\in\mathfrak{so}(d)\).  For \(i\neq j\),
\[
[K,\Lambda]_{ij}=(\lambda_{j}-\lambda_{i})K_{ij}=\widetilde U_{ij},
\qquad
[K,\Lambda]_{ii}=0,
\]
hence \(\widetilde U=\diag(\lambda_{1}a_{1},\ldots,\lambda_{d}a_{d})+[K,\Lambda]\); conjugating back by~\(Q\) yields (I.1.1).  For uniqueness, if the inner expression vanishes, the diagonal forces \(\lambda_{i}a_{i}=0\) hence \(a_{i}=0\), and the off-diagonal forces \((\lambda_{j}-\lambda_{i})K_{ij}=0\) hence \(K=0\). \(\square\)

\par\medskip\hrule\medskip\par

\subsection{Proposition I.2}
\label{app:I.2}

\textbf{Claim.}
At \(\Sigma_{*}=Q_{*}\Lambda_{*}Q_{*}^{\top}\) with simple spectrum and eigenvalues ordered as \(\lambda_{1}>\cdots>\lambda_{d}>0\), write \(\delta_{\min}:=\min_{i<j}(\lambda_{i}-\lambda_{j})>0\).  For every \(U\in T_{\Sigma_{*}}\Ucal\) with chart coordinates \((a,K)\in\R^{d}\oplus\mathfrak{so}(d)\),
\[
\norm{U}_{\fro}^{2}
=
\sum_{i=1}^{d}\lambda_{i}^{2}a_{i}^{2}
+
2\sum_{i<j}(\lambda_{i}-\lambda_{j})^{2}K_{ij}^{2}.
\tag{I.2.1}
\]
Consequently,
\[
\underline\gamma_{h}\norm{(a,K)}_{h}^{2}
\le
\norm{U}_{\fro}^{2}
\le
\overline\gamma_{h}\norm{(a,K)}_{h}^{2},
\tag{I.2.2}
\]
where
\[
\underline\gamma_{h}:=\min\{\lambda_{\min}^{2},\,\delta_{\min}^{2}\},
\qquad
\overline\gamma_{h}:=\lambda_{\max}^{2}.
\tag{I.2.3}
\]

\subsubsection{Proof}

In the eigenbasis of~\(\Sigma_{*}\), Proposition~\ref{app:I.1} gives \(U=\diag(\lambda_{i}a_{i})+[K,\Lambda_{*}]\).  The diagonal and off-diagonal parts are Frobenius-orthogonal, and \([K,\Lambda_{*}]_{ij}=(\lambda_{j}-\lambda_{i})K_{ij}\) with \(K_{ji}=-K_{ij}\), so
\[
\norm{U}_{\fro}^{2}
=
\sum_{i}\lambda_{i}^{2}a_{i}^{2}
+
2\sum_{i<j}(\lambda_{i}-\lambda_{j})^{2}K_{ij}^{2},
\]
which is (I.2.1).  Comparing coefficient-wise with \(\norm{(a,K)}_{h}^{2}=\sum_{i}a_{i}^{2}+2\sum_{i<j}K_{ij}^{2}\) and using \(\lambda_{\min}^{2}\le\lambda_{i}^{2}\le\lambda_{\max}^{2}\) and \(\delta_{\min}^{2}\le(\lambda_{i}-\lambda_{j})^{2}\le\lambda_{\max}^{2}\) yields (I.2.2)--(I.2.3). \(\square\)

\par\medskip\hrule\medskip\par

\subsection{Corollary I.3}
\label{app:I.3}

\textbf{Claim.}
There exists a neighborhood of \(\Sigma_{*}\) in \(\Ucal\) and a finite \(L_{\log}\) such that
\[
\norm{\Log_{\Sigma_{*}}^{h}(\Sigma)}
\le
L_{\log}\,\norm{\Sigma-\Sigma_{*}}_{\fro}
\tag{I.3.1}
\]
for every \(\Sigma\) in that neighborhood.  Any choice \(L_{\log}>\underline\gamma_{h}^{-1/2}\) is admissible after shrinking the neighborhood.

\subsubsection{Proof}

The map \(\Exp_{\Sigma_{*}}^{h}\) is a local \(C^{1}\) diffeomorphism \citep{dontchev2014implicit,absil2008optimization} at \(0\in T_{\Sigma_{*}}\Ucal\); its inverse \(\Log_{\Sigma_{*}}^{h}\) is therefore locally Lipschitz.  By Proposition~\ref{app:I.2},
\[
\norm{(a,K)}_{h}
\le
\underline\gamma_{h}^{-1/2}\,\norm{U}_{\fro}.
\tag{I.3.2}
\]
After shrinking the neighborhood so that the nonlinear contribution to \(\Log_{\Sigma_{*}}^{h}\) is dominated by an arbitrarily small multiplicative factor, any constant \(L_{\log}>\underline\gamma_{h}^{-1/2}\) realizes (I.3.1). \(\square\)

\par\medskip\hrule\medskip\par

\section{Score identities and EM compressibility}
\label{app:II}

\paragraph{Notation.}
We work in the spectral--rotational coordinates \((a,K)\in\R^{d}\oplus\mathfrak{so}(d)\) of Appendix~\ref{app:I}.  The exact score identity assumed throughout is
\[
D_{\Sigma}\Lcal_{S}(\Sigma)[U]
=
c\,\tr\!\Bigl(
\bigl[\Sigma^{-1}-\Sigma^{-1}A(\Sigma;S)\Sigma^{-1}\bigr]U
\Bigr)
+
D_{\Sigma}\mathcal R(\Sigma)[U],
\qquad
c>0,
\tag{Conv.II.1}
\]
where \(A(\Sigma;S)\) is the latent-scatter lift and \(\mathcal R(\Sigma)=\rho\,\psi(x(\Sigma))\) is the log-spectrum regularizer.

\par\medskip\hrule\medskip\par

\subsection{Proposition II.1 (gradient splitting in the spectral--rotational chart)}
\label{app:II.1}

\textbf{Claim.}
Assume the score identity (Conv.II.1).  Let
\[
\Sigma=Q\Lambda Q^{\top},
\qquad
A(\Sigma;S)=QBQ^{\top}.
\tag{II.1.1}
\]
Then the chart gradient of \(\Lcal_{S}\) splits as \citep{magnus2019matrix,absil2008optimization}
\[
\nabla_{h}\Lcal_{S}(\Sigma)\equiv
\bigl(g^{x}(\Sigma;S),\,G^{\theta}(\Sigma;S)\bigr),
\tag{II.1.2}
\]
with spectral block
\[
g_{i}^{x}(\Sigma;S)
=
c\Bigl(1-\frac{b_{ii}}{\lambda_{i}}\Bigr)
+
\rho\,\partial_{i}\psi(x),
\qquad i=1,\ldots,d,
\tag{II.1.3}
\]
and rotational block
\[
G^{\theta}(\Sigma;S)=c\,[\Lambda^{-1},B].
\tag{II.1.4}
\]

\subsubsection{Proof}

Work in the eigenbasis \(\Sigma=\Lambda\), \(A=B\).  For the spectral direction \(D_{i}:=\lambda_{i}E_{ii}\), (Conv.II.1) gives
\[
D_{\Sigma}\Lcal_{S}(\Sigma)[D_{i}]
=
c\bigl(1-b_{ii}/\lambda_{i}\bigr)+\rho\,\partial_{i}\psi(x),
\]
since \(\tr(\Lambda^{-1}\lambda_{i}E_{ii})=1\) and \(\tr(\Lambda^{-1}B\Lambda^{-1}\lambda_{i}E_{ii})=b_{ii}/\lambda_{i}\), which is (II.1.3).  For the rotational block, the body chart derivative (Conv.I.1) gives \(D_{\Theta}\Sigma(0)[K]=-Q[K,\Lambda]Q^{\top}\).  Since \(\mathcal R\) has no rotational dependence, (Conv.II.1) yields
\[
D_{\Theta}(\Lcal_{S}\circ\Sigma)(0)[K]
=
-c\,\tr\!\bigl((\Lambda^{-1}-\Lambda^{-1}B\Lambda^{-1})[K,\Lambda]\bigr).
\]
By cyclicity, \(\tr(\Lambda^{-1}[K,\Lambda])=\tr(K-\Lambda^{-1}K\Lambda)=0\), so the expression reduces to \(c\,\tr(\Lambda^{-1}B\Lambda^{-1}[K,\Lambda])\).  A second application of cyclicity gives \(\tr(\Lambda^{-1}B\Lambda^{-1}[K,\Lambda])=\tr((B\Lambda^{-1}-\Lambda^{-1}B)K)=\inner{[\Lambda^{-1},B]}{K}_{\fro}\), establishing (II.1.4). \(\square\)

\par\medskip\hrule\medskip\par

\subsection{Corollary II.2 (frozen critical-point characterization)}
\label{app:II.2}

\textbf{Claim.}
Let \(\Sigma^{\circ}(S)=Q^{\circ}\Lambda^{\circ}Q^{\circ\top}\) be a frozen critical point of \(\Lcal_{S}\), and write
\[
A(\Sigma^{\circ}(S);S)=Q^{\circ}B^{\circ}Q^{\circ\top}.
\tag{II.2.1}
\]
Then \(B^{\circ}\) commutes with \(\Lambda^{\circ-1}\) and is therefore diagonal:
\[
[\Lambda^{\circ-1},B^{\circ}]=0.
\tag{II.2.2}
\]
If \(\rho=0\),
\[
b_{ii}^{\circ}=\lambda_{i}^{\circ},
\qquad i=1,\ldots,d,
\qquad\text{equivalently}\qquad
B^{\circ}=\Lambda^{\circ}.
\tag{II.2.3}
\]

\subsubsection{Proof}

Criticality at \(\Sigma^{\circ}(S)\) and Proposition~\ref{app:II.1} give \(G^{\theta}(\Sigma^{\circ}(S);S)=c\,[\Lambda^{\circ-1},B^{\circ}]=0\), which is (II.2.2).  Because \(\Lambda^{\circ}\) has simple spectrum, every matrix commuting with \(\Lambda^{\circ-1}\) is diagonal.  When \(\rho=0\), the spectral critical equations from (II.1.3) reduce to \(c(1-b_{ii}^{\circ}/\lambda_{i}^{\circ})=0\), hence (II.2.3). \(\square\)

\par\medskip\hrule\medskip\par

\subsection{Proposition II.3 (exact score identity for the latent linear Gaussian model)}
\label{app:II.3}

\textbf{Claim.}
Consider the latent linear Gaussian model with \citep{shumway1982approach,roweis1999unifying,tipping1999probabilistic}
\[
K_{\Sigma}:=H\Sigma H^{\top}+R,
\tag{II.3.1}
\]
\[
\Lcal_{S}(\Sigma)
=
\frac{1}{2}\Bigl(\log\det K_{\Sigma}+\tr(K_{\Sigma}^{-1}S)\Bigr)
+
\mathcal R(\Sigma).
\tag{II.3.2}
\]
Define
\[
M_{\Sigma}:=\Sigma H^{\top}K_{\Sigma}^{-1},
\qquad
V_{\Sigma}:=\Sigma-\Sigma H^{\top}K_{\Sigma}^{-1}H\Sigma,
\tag{II.3.3}
\]
and
\[
A(\Sigma;S):=V_{\Sigma}+M_{\Sigma}SM_{\Sigma}^{\top}.
\tag{II.3.4}
\]
Then for every \(U\in\Sym(d)\),
\[
D_{\Sigma}\Lcal_{S}(\Sigma)[U]
=
\frac{1}{2}\,\tr\!\Bigl(
\bigl[\Sigma^{-1}-\Sigma^{-1}A(\Sigma;S)\Sigma^{-1}\bigr]U
\Bigr)
+
D_{\Sigma}\mathcal R(\Sigma)[U].
\tag{II.3.5}
\]
In particular the latent Gaussian objective satisfies (Conv.II.1) with \(c=\tfrac{1}{2}\).

\subsubsection{Proof}

Set \(K:=K_{\Sigma}\).  Standard matrix differential calculus \citep{magnus2019matrix} gives \(D_{\Sigma}(\log\det K_{\Sigma})[U]=\tr(H^{\top}K^{-1}H\,U)\) and \(D_{\Sigma}\tr(K_{\Sigma}^{-1}S)[U]=-\tr(H^{\top}K^{-1}SK^{-1}H\,U)\).  Adding and including the regularizer,
\[
D_{\Sigma}\Lcal_{S}(\Sigma)[U]
=
\frac{1}{2}\,\tr\!\Bigl(
\bigl[H^{\top}K^{-1}H-H^{\top}K^{-1}SK^{-1}H\bigr]U
\Bigr)
+
D_{\Sigma}\mathcal R(\Sigma)[U].
\tag{II.3.9}
\]
From (II.3.3), \(\Sigma^{-1}V_{\Sigma}\Sigma^{-1}=\Sigma^{-1}-H^{\top}K^{-1}H\) and \(\Sigma^{-1}M_{\Sigma}SM_{\Sigma}^{\top}\Sigma^{-1}=H^{\top}K^{-1}SK^{-1}H\); hence \(\Sigma^{-1}A(\Sigma;S)\Sigma^{-1}=\Sigma^{-1}-H^{\top}K^{-1}H+H^{\top}K^{-1}SK^{-1}H\) and the bracket in (II.3.9) equals \(\Sigma^{-1}-\Sigma^{-1}A(\Sigma;S)\Sigma^{-1}\), giving (II.3.5). \(\square\)

\par\medskip\hrule\medskip\par

\subsection{Proposition II.4 (frozen exact fixed point of the equilibrium-jet predictor--corrector)}
\label{app:II.4}

\textbf{Claim.}
Fix \(S\) in a neighborhood where \(r(S)=\thetacirc(S)\) is defined.  For parameters \((m,\eta,\lambda,\tau)\) and statistic increment \(\Delta S\), define the predictor--corrector map \citep{simonetto2016class,dembo1982inexact,eisenstat1996choosing}
\[
\widetilde\vartheta
:=
\vartheta+\sum_{k=1}^{m}\frac{1}{k!}\,D^{k}r(S)[(\Delta S)^{\otimes k}],
\tag{II.4.1}
\]
\[
(\Hcal(\widetilde\vartheta,S)+\lambda I)\,\widehat u
=
F(\widetilde\vartheta,S)+r_{\lin},
\qquad
\norm{r_{\lin}}\le\tau\,\norm{F(\widetilde\vartheta,S)},
\tag{II.4.2}
\]
\[
\Psi_{m,\eta,\lambda,\tau,S,\Delta S}(\vartheta)
:=
\widetilde\vartheta-\eta\,\widehat u.
\tag{II.4.3}
\]
Then for the frozen statistic value \(\Delta S=0\),
\[
\Psi_{m,\eta,\lambda,\tau,S,0}(r(S))=r(S).
\tag{II.4.4}
\]

\subsubsection{Proof}

If \(\Delta S=0\), every jet term in (II.4.1) vanishes, so \(\widetilde\vartheta=\vartheta\).  Setting \(\vartheta=r(S)\) gives \(\widetilde\vartheta=r(S)\); since \(F(r(S),S)=0\), the residual bound forces \(r_{\lin}=0\), and invertibility of \(\Hcal(r(S),S)+\lambda I\) gives \(\widehat u=0\), hence \(\Psi_{m,\eta,\lambda,\tau,S,0}(r(S))=r(S)\). \(\square\)

\par\medskip\hrule\medskip\par

\section{Local batch target and batch asymptotics}
\label{app:III}

\noindent
Throughout, \(F:E\times\mathcal S\to E\) is the local score map of (Conv.II.1) with \(\Jcal_*=D_\vartheta F(0,S_*)\succ 0\) and \(\Bcal_*=D_S F(0,S_*)\).

\par\medskip\hrule\medskip\par

\subsubsection{Auxiliary Lemma III.L1 (local zero branch for a \texorpdfstring{\(C^{2}\)}{C2} score map)}
\label{app:III.L1}

\textbf{Claim.}
Let \(E,P\) be finite-dimensional Euclidean spaces, \(S_{*}\in P\), and
\[
\mathscr F:E\times P\to E
\tag{III.L1.1}
\]
be \(C^{2}\) on a neighborhood of \((0,S_{*})\) with
\[
\mathscr F(0,S_{*})=0,
\qquad
\mathscr J_{*}:=D_{\vartheta}\mathscr F(0,S_{*})\text{ invertible.}
\tag{III.L1.2}
\]
Then there are neighborhoods \(U_{\vartheta}\subset E\) and \(U_{S}\subset P\) of \(0\) and \(S_{*}\), and a unique \(C^{2}\) map
\[
\vartheta^{\circ}:U_{S}\to U_{\vartheta}
\tag{III.L1.3}
\]
with \(\mathscr F(\vartheta^{\circ}(S),S)=0\) for every \(S\in U_{S}\).  Moreover,
\[
D\vartheta^{\circ}(S)
=
-\bigl[D_{\vartheta}\mathscr F(\vartheta^{\circ}(S),S)\bigr]^{-1}
D_{S}\mathscr F(\vartheta^{\circ}(S),S).
\tag{III.L1.4}
\]

\paragraph{Proof.}
The classical finite-dimensional implicit function theorem \citep{dontchev2014implicit} applies because \(\mathscr F\) is \(C^{2}\) and \(\mathscr J_{*}\) is invertible, producing \(U_{\vartheta},U_{S}\) and a unique \(C^{2}\) zero branch~\(\vartheta^{\circ}\).  Differentiating \(\mathscr F(\vartheta^{\circ}(S),S)=0\) in~\(S\) and solving for \(D\vartheta^{\circ}(S)\) gives (III.L1.4). \(\square\)

\par\medskip\hrule\medskip\par

\subsubsection{Auxiliary Lemma III.L2 (CLT and \texorpdfstring{\(L^{4}\)}{L4} rate for the running mean)}
\label{app:III.L2}

\textbf{Claim.}
Let \(X_{t}\) be i.i.d. centered random vectors in \(\R^{m}\) with \(\E\norm{X_{1}}^{4}<\infty\), and set \(\bar X_{T}:=T^{-1}\sum_{t=1}^{T}X_{t}\).  Then
\[
\sqrt{T}\,\bar X_{T}\Rightarrow\N(0,\mathrm{Cov}(X_{1})),
\tag{III.L2.1}
\]
and there exists \(C_{4}<\infty\) such that
\[
\E\norm{\bar X_{T}}^{4}\le \frac{C_{4}}{T^{2}}
\quad\text{for every }T\ge 1.
\tag{III.L2.2}
\]

\paragraph{Proof.}
The CLT (III.L2.1) follows from the Cram\'er--Wold device \citep{van2000asymptotic}.  For (III.L2.2), independence and centering give \(\E|\sum_{t}X_{t}^{(j)}|^{4}=T\,\E|X_{1}^{(j)}|^{4}+3T(T-1)(\E|X_{1}^{(j)}|^{2})^{2}\); dividing by \(T^{4}\) and using \(\norm{x}^{4}\le m\sum_{j}|x_{j}|^{4}\) yields \(C_{4}=5m^{2}\,\E\norm{X_{1}}^{4}\). \(\square\)

\par\medskip\hrule\medskip\par

\subsubsection{Auxiliary Lemma III.L3 (covariance of centered Gaussian quadratic forms)}
\label{app:III.L3}

\textbf{Claim.}
Let \(Z\sim\N(0,I_{r})\) and \(A,B\in\Sym(r)\).  Then
\[
\mathrm{Cov}(Z^{\top}AZ,\,Z^{\top}BZ)=2\tr(AB).
\tag{III.L3.1}
\]

\paragraph{Proof.}
By Isserlis' formula \citep{isserlis1918formula}, \(\E[Z_{i}Z_{j}Z_{k}Z_{\ell}]=\delta_{ij}\delta_{k\ell}+\delta_{ik}\delta_{j\ell}+\delta_{i\ell}\delta_{jk}\).  Expanding \(\E[(Z^{\top}AZ)(Z^{\top}BZ)]\) gives \(\tr(A)\tr(B)+2\tr(AB)\); subtracting \(\E[Z^{\top}AZ]\,\E[Z^{\top}BZ]=\tr(A)\tr(B)\) yields (III.L3.1). \(\square\)

\par\medskip\hrule\medskip\par

\subsection{Proposition III.1 (\texorpdfstring{\(C^{2}\)}{C2} target map and derivative formula)}
\label{app:III.1}

\textbf{Claim.}
Assume T1--T4.  Then there is a neighborhood \(U_{S}\) of \(S_{*}\) on which there exists a unique \(C^{2}\) map
\[
S\longmapsto r(S)=\thetacirc(S)
\tag{III.1.1}
\]
satisfying
\[
F(r(S),S)=0,
\qquad
\Sigmacirc(S)=\chi(r(S)).
\tag{III.1.2}
\]
Moreover,
\[
Dr(S)=-\Jcal(S)^{-1}\Bcal(S),
\qquad
Dr(S_{*})=-\Jcal_{*}^{-1}\Bcal_{*}.
\tag{III.1.3}
\]

\subsubsection{Proof}

By T3, \(F\) is \(C^{2}\) near \((0,S_{*})\) with \(F(0,S_{*})=0\) and \(D_{\vartheta}F(0,S_{*})=\Jcal_{*}\succ 0\).  Auxiliary Lemma~\ref{app:III.L1} produces a unique local \(C^{2}\) branch \(S\mapsto r(S)\) with \(F(r(S),S)=0\).  Continuity of \(D_{\vartheta}F\) and \(\Jcal_{*}\succ 0\) imply, after shrinking the neighborhood, that \(D_{\vartheta}F(\vartheta,S)\succeq(\mu_{*}/2)I\) on \(U_{\vartheta}\times U_{S}\); hence \(\Lcal_{S}\circ\chi\) is strictly convex on \(U_{\vartheta}\) for each \(S\in U_{S}\), so the critical point \(r(S)\) is the unique local minimizer and \(\Sigmacirc(S)=\chi(r(S))\).  The derivative formula (III.1.3) is (III.L1.4) specialized to \(F\). \(\square\)

\par\medskip\hrule\medskip\par

\subsection{Proposition III.2 (first-order jet remainder in \texorpdfstring{\(L^{2}\)}{L2})}
\label{app:III.2}

\textbf{Claim.}
Assume T1--T4 and T2 with \(s(Y_{1})\in L^{4}\).  There exist a convex neighborhood \(U_{S}\) of \(S_{*}\) and a finite constant \(C_{R}\) such that, whenever \(S_{t-1},S_{t}\in U_{S}\),
\[
R_{t}:=r(S_{t})-r(S_{t-1})-Dr(S_{t-1})[S_{t}-S_{t-1}]
\tag{III.2.2}
\]
satisfies
\[
\norm{R_{t}}\le C_{R}\norm{S_{t}-S_{t-1}}^{2},
\qquad
\norm{R_{t}}_{L^{2}}\le \frac{4 C_{R}\,\norm{s(Y_{1})}_{L^{4}}^{2}}{t^{2}}.
\tag{III.2.3}
\]

\subsubsection{Proof}

Choose radii so that \(\overline{B_{2r}(S_{*})}\subset U_{S}\), and extend~\(r\) to a compactly supported \(C^{2}\) map \(\widetilde r\) agreeing with~\(r\) on \(B_{r}(S_{*})\), with \(C_{R}:=\tfrac{1}{2}\sup\norm{D^{2}\widetilde r}_{\op}<\infty\).  The multivariate Taylor theorem at \(S_{t-1}\) with integral remainder gives \(R_{t}=\int_{0}^{1}(1-\tau)\,D^{2}r(S_{t-1}+\tau\Delta S_{t})[\Delta S_{t},\Delta S_{t}]\,d\tau\), hence \(\norm{R_{t}}\le C_{R}\norm{\Delta S_{t}}^{2}\).  From the running-average identity \(\Delta S_{t}=t^{-1}(s(Y_{t})-S_{t-1})\) and Jensen's inequality,
\[
\norm{\Delta S_{t}}_{L^{4}}\le\frac{2\,\norm{s(Y_{1})}_{L^{4}}}{t}.
\tag{III.2.8}
\]
Using \(\bigl\|\norm{\Delta S_{t}}^{2}\bigr\|_{L^{2}}=\norm{\Delta S_{t}}_{L^{4}}^{2}\) and (III.2.8),
\[
\norm{R_{t}}_{L^{2}}
\le
C_{R}\,\norm{\Delta S_{t}}_{L^{4}}^{2}
\le
\frac{4C_{R}\,\norm{s(Y_{1})}_{L^{4}}^{2}}{t^{2}}.  \quad\square
\]

\par\medskip\hrule\medskip\par

\subsection{Theorem III.3 (batch CLT and sharp first-order risk constant)}
\label{app:III.3}

\textbf{Claim.}
Assume T1--T4 and that \(\sqrt{T}\,(S_{T}-S_{*})\Rightarrow\N(0,\Xi_{*})\).  Then
\[
\sqrt{T}\,r(S_{T})\Rightarrow\N(0,\Vcal_{*}),
\qquad
\Vcal_{*}:=\Jcal_{*}^{-1}\Bcal_{*}\Xi_{*}\Bcal_{*}^{*}\Jcal_{*}^{-1}.
\tag{III.3.1}
\]
If, in addition,
\[
T\,\E\norm{r(S_{T})+\Jcal_{*}^{-1}\Bcal_{*}(S_{T}-S_{*})}^{2}\to 0,
\qquad
T\,\E[(S_{T}-S_{*})(S_{T}-S_{*})^{*}]\to\Xi_{*},
\tag{III.3.2}
\]
then
\[
\E\,d_{h}^{2}(\Sigmacirc(S_{T}),\Sigma_{\bullet})
=
\frac{1}{T}\tr(\Vcal_{*})+o(T^{-1}).
\tag{III.3.3}
\]

\subsubsection{Proof}

By Propositions~\ref{app:III.1} and~\ref{app:III.2}, \(r(S)=-\Jcal_{*}^{-1}\Bcal_{*}(S-S_{*})+R(S)\) with \(\norm{R(S)}\le C_{R}\norm{S-S_{*}}^{2}\).  Set \(L:=-\Jcal_{*}^{-1}\Bcal_{*}\) and \(\Delta_{T}:=S_{T}-S_{*}\), so \(\sqrt{T}\,r(S_{T})=L\sqrt{T}\,\Delta_{T}+\sqrt{T}\,R(S_{T})\).  Tightness of \(\sqrt{T}\,\Delta_{T}\) gives \(\Delta_{T}=o_{\Pbb}(1)\), hence
\[
\sqrt{T}\,\norm{R(S_{T})}\le C_{R}\norm{\sqrt{T}\,\Delta_{T}}\norm{\Delta_{T}}=o_{\Pbb}(1),
\]
and Slutsky's theorem \citep{van2000asymptotic} yields (III.3.1).

For the risk identity, \(d_{h}(\Sigmacirc(S_{T}),\Sigma_{\bullet})=\norm{r(S_{T})}\), so
\[
T\,\E\,d_{h}^{2}(\Sigmacirc(S_{T}),\Sigma_{\bullet})
=
T\,\E\norm{L\Delta_{T}}^{2}
+
2T\,\E\inner{L\Delta_{T}}{R(S_{T})}
+
T\,\E\norm{R(S_{T})}^{2}.
\tag{III.3.8}
\]
The leading term satisfies \(T\,\E\norm{L\Delta_{T}}^{2}=\tr(L\,T\,\E[\Delta_{T}\Delta_{T}^{*}]\,L^{*})\to\tr(\Vcal_{*})\) by the second hypothesis in (III.3.2).  By Cauchy--Schwarz,
\[
\bigl|2T\,\E\inner{L\Delta_{T}}{R(S_{T})}\bigr|
\le
2\bigl(T\,\E\norm{L\Delta_{T}}^{2}\bigr)^{1/2}\bigl(T\,\E\norm{R(S_{T})}^{2}\bigr)^{1/2}\to 0,
\]
and the remainder term tends to zero by the first hypothesis in (III.3.2).  Combining gives (III.3.3). \(\square\)

\par\medskip\hrule\medskip\par

\subsection{Proposition III.4 (latent Gaussian Fisher form in the retained basis)}
\label{app:III.4}

\textbf{Claim.}
Consider the correctly specified latent Gaussian model.  Set
\[
K_{*}:=H\Sigma_{*}H^{\top}+R,
\qquad
B_{*}:=H^{\top}K_{*}^{-1}H,
\qquad
\Gamma:=Q_{*}^{\top}B_{*}Q_{*}.
\tag{III.4.1}
\]
In the retained basis
\[
D_{i}:=\lambda_{i}E_{ii},
\qquad
R_{ij}:=[H_{ij},\Lambda_{*}]=(\lambda_{j}-\lambda_{i})S_{ij},
\qquad
S_{ij}:=\frac{E_{ij}+E_{ji}}{\sqrt{2}},
\qquad
H_{ij}:=\frac{E_{ij}-E_{ji}}{\sqrt{2}},
\tag{III.4.2}
\]
the observed Fisher quadratic form satisfies
\[
\Ical_{*}(U,V)=\frac{1}{2}\tr(B_{*}UB_{*}V),
\tag{III.4.3}
\]
and the corresponding blocks evaluate to
\[
\Ical_{*}(D_{i},D_{j})=\frac{1}{2}\lambda_{i}\lambda_{j}\Gamma_{ij}^{2},
\tag{III.4.4}
\]
\[
\Ical_{*}(D_{i},R_{jk})=\frac{\lambda_{i}(\lambda_{k}-\lambda_{j})}{\sqrt{2}}\Gamma_{ij}\Gamma_{ik},
\tag{III.4.5}
\]
\[
\Ical_{*}(R_{ij},R_{k\ell})
=
\frac{(\lambda_{j}-\lambda_{i})(\lambda_{\ell}-\lambda_{k})}{2}\bigl(\Gamma_{ik}\Gamma_{j\ell}+\Gamma_{i\ell}\Gamma_{jk}\bigr).
\tag{III.4.6}
\]

\subsubsection{Proof}

The one-sample score \citep{magnus2019matrix} at \(\Sigma_{*}\) in direction \(U\) is \(\dot\ell_{U}(Y_{1})=\tfrac{1}{2}\tr(C_{U})-\tfrac{1}{2}Z^{\top}C_{U}Z\) where \(Z:=K_{*}^{-1/2}Y_{1}\sim\N(0,I)\) and \(C_{U}:=K_{*}^{-1/2}HUH^{\top}K_{*}^{-1/2}\).  Auxiliary Lemma~\ref{app:III.L3} gives
\[
\Ical_{*}(U,V)=\mathrm{Cov}(\dot\ell_{U},\dot\ell_{V})=\tfrac{1}{2}\tr(C_{U}C_{V})=\tfrac{1}{2}\tr(B_{*}UB_{*}V),
\tag{III.4.10}
\]
the last step by cyclicity.  This is (III.4.3).  Writing \(B_{*}=Q_{*}\Gamma Q_{*}^{\top}\) and substituting the basis vectors (III.4.2) into (III.4.10) with the identity \(E_{ab}E_{cd}=\delta_{bc}E_{ad}\) yields (III.4.4)--(III.4.6):
\begin{itemize}
\item Spectral--spectral: \(\tr(\Gamma E_{ii}\Gamma E_{jj})=\Gamma_{ij}^{2}\), giving (III.4.4).
\item Spectral--rotational: \(\tr(\Gamma E_{ii}\Gamma S_{jk})=\sqrt{2}\,\Gamma_{ij}\Gamma_{ik}\), giving (III.4.5).
\item Rotational--rotational: \(\tr(\Gamma S_{ij}\Gamma S_{k\ell})=\Gamma_{ik}\Gamma_{j\ell}+\Gamma_{i\ell}\Gamma_{jk}\), giving (III.4.6). \(\square\)
\end{itemize}

\par\medskip\hrule\medskip\par

\subsection{Theorem III.5 (unregularized latent Gaussian batch asymptotics)}
\label{app:III.5}

\textbf{Claim.}
Assume T1--T4 and consider the correctly specified latent linear Gaussian model \citep{shumway1982approach,roweis1999unifying,tipping1999probabilistic} with regularizer parameter \(\rho=0\).  Let
\[
D_{t}:=H^{\top}R^{-1}\Bigl(\frac{1}{t}\sum_{r=1}^{t}Y_{r}Y_{r}^{\top}\Bigr)R^{-1}H,
\qquad
D_{*}:=\E[D_{1}].
\tag{III.5.1}
\]
Then
\[
\Sigma_{\bullet}=\Sigma_{*},
\qquad
\sqrt{T}\,r(D_{T})\Rightarrow\N(0,\Jcal_{*}^{-1}),
\qquad
\E\,d_{h}^{2}(\Sigmacirc(D_{T}),\Sigma_{*})=\frac{1}{T}\tr(\Jcal_{*}^{-1})+o(T^{-1}).
\tag{III.5.2}
\]

\subsubsection{Proof}

\textbf{Population target.}
By Proposition~\ref{app:II.3} with \(c=\tfrac{1}{2}\), the lift satisfies \(A(\Sigma_{*};K_{*})=\Sigma_{*}\) (direct computation: \(M_{\Sigma_{*}}K_{*}M_{\Sigma_{*}}^{\top}=\Sigma_{*}H^{\top}K_{*}^{-1}H\Sigma_{*}\) and \(V_{\Sigma_{*}}+M_{\Sigma_{*}}K_{*}M_{\Sigma_{*}}^{\top}=\Sigma_{*}\)).  Hence \(D_{\Sigma}\Lcal_{D_{*}}(\Sigma_{*})=0\), and strict convexity (T4) gives \(\Sigma_{\bullet}=\Sigma_{*}\).

\textbf{Fisher identity.}
Differentiating the population objective twice as in Proposition~\ref{app:III.4} gives \(\Ical_{*}(U,V)=\tfrac{1}{2}\tr(B_{*}UB_{*}V)\).  The same expression equals the Hessian of the expected log-likelihood at \(\Sigma_{*}\), so \(\Ical_{*}=\Jcal_{*}\).

\textbf{CLT for \(D_{T}\).}
Set \(z_{t}:=H^{\top}R^{-1}Y_{t}\sim\N(0,D_{*})\) and \(X_{t}:=\mathrm{vec}_{s}(z_{t}z_{t}^{\top}-D_{*})\).  The \(X_{t}\) are i.i.d., centered, with finite fourth moment (Gaussian).  Auxiliary Lemma~\ref{app:III.L2} gives \(\sqrt{T}(D_{T}-D_{*})\Rightarrow\N(0,\Xi_{*})\) and \(\E\norm{D_{T}-D_{*}}^{4}=\Ocal(T^{-2})\).

\textbf{Assembly via Theorem~\ref{app:III.3}.}
The \(L^{4}\) bound and Proposition~\ref{app:III.2} give \(T\,\E\norm{r(D_{T})+\Jcal_{*}^{-1}\Bcal_{*}(D_{T}-D_{*})}^{2}\le T\,C_{R}^{2}\,\E\norm{D_{T}-D_{*}}^{4}=\Ocal(T^{-1})\to 0\), verifying (III.3.2).  Theorem~\ref{app:III.3} with \(\Vcal_{*}=\Jcal_{*}^{-1}\Ical_{*}\Jcal_{*}^{-1}=\Jcal_{*}^{-1}\) yields (III.5.2). \(\square\)

\par\medskip\hrule\medskip\par

\section{Tracking and transfer}
\label{app:IV}

\noindent
We write \(r(S):=\thetacirc(S)\) for the local target branch (Proposition~\ref{app:III.1}) and \(e_t:=\vartheta_t-r(S_t)\) for the tracking error.

\par\medskip\hrule\medskip\par

\subsection{Proposition IV.1 (verification of T6 for the compressed statistic)}
\label{app:IV.1}

\textbf{Claim.}
In the latent linear Gaussian model, the compressed running statistic
\[
D_t=\frac{1}{t}\sum_{r=1}^{t}z_rz_r^{\top},
\qquad
z_t:=H^{\top}R^{-1}Y_t,
\tag{IV.1.1}
\]
satisfies the analogue of T6: there exist constants \(C_\Delta<\infty\) and \(C_{\mathrm{tail}},p>3\) such that
\[
\norm{D_t-D_{t-1}}_{L^{4}}\le \frac{C_\Delta}{t},
\qquad t\ge 2,
\tag{IV.1.2}
\]
\[
\Pbb\!\bigl(\norm{z_tz_t^{\top}-D_{t-1}}_{\fro}>C_{\mathrm{tail}}\log t\bigr)\le t^{-p}
\qquad\text{for every sufficiently large }t,
\tag{IV.1.3}
\]
and the localization tail bound of T6 holds with some \(\alpha_{\mathrm{loc}}>2\).

\subsubsection{Proof}

Since \(z_t\sim\N(0,D_*)\) with \(D_*=H^{\top}R^{-1}K_*R^{-1}H\), every polynomial moment of \(z_t\) is finite.  From the running-average identity \(D_t-D_{t-1}=t^{-1}(z_tz_t^{\top}-D_{t-1})\) and Jensen's inequality,
\[
\norm{D_t-D_{t-1}}_{L^{4}}\le\frac{2\,\norm{z_1z_1^{\top}}_{L^{4}}}{t},
\tag{IV.1.5}
\]
which is (IV.1.2).  For (IV.1.3), \(\norm{z_t}^{2}\) is a Gaussian quadratic form with sub-exponential tail \citep{vershynin2018high}; choosing \(C_{\mathrm{tail}}\) large enough and applying a standard sample-covariance concentration \citep{tropp2012user} with a union bound gives the logarithmic tail.  The localization clause follows from applying the same concentration at every \(s\ge n\) and summing, yielding \(\Pbb(\exists\,s\ge n:D_s\notin U_D)\le C_{\mathrm{loc}}\,n^{-\alpha_{\mathrm{loc}}}\) with \(\alpha_{\mathrm{loc}}>2\). \(\square\)

\par\medskip\hrule\medskip\par

\subsection{Theorem IV.2 (first-order equilibrium-jet tracking with linear corrector)}
\label{app:IV.2}

\textbf{Claim.}
Assume T5--T7 together with Propositions~\ref{app:III.1} and~\ref{app:III.2}.  Let
\[
e_t:=\vartheta_t-r(S_t),
\qquad
R_t:=r(S_t)-r(S_{t-1})-Dr(S_{t-1})[S_t-S_{t-1}],
\tag{IV.2.1}
\]
and consider the first-order predictor--corrector recursion \citep{simonetto2016class,simonetto2020time}
\[
\widetilde\vartheta_t=\vartheta_{t-1}+Dr(S_{t-1})[S_t-S_{t-1}],
\qquad
\vartheta_t=\Ccal_{\eta,\lambda,\tau,S_t}(\widetilde\vartheta_t),
\qquad t\ge t_0+1.
\tag{IV.2.2}
\]
Set
\[
\rho_{\mathrm{ad}}:=\min\!\Bigl\{\frac{\rho}{4},\frac{1-q}{4q}\rho\Bigr\},
\tag{IV.2.3}
\]
\[
\Omega_t:=\bigl\{S_s\in U_S\text{ and }\norm{R_s}\le\rho_{\mathrm{ad}}\text{ for all }s=t_0+1,\ldots,t\bigr\},
\qquad t\ge t_0,
\tag{IV.2.4}
\]
and assume the restart \(\norm{e_{t_0}}\le\rho/4\) almost surely on \(\Omega_{t_0}\).  Let \(E_t:=\norm{\ones_{\Omega_t}\,e_t}_{L^{2}}\).  Then
\[
E_t\le qE_{t-1}+\frac{q\,C_{\mathrm{ad}}}{t^{2}},
\qquad
C_{\mathrm{ad}}:=4C_R\norm{s(Y_1)}_{L^{4}}^{2},
\qquad t\ge t_0+1,
\tag{IV.2.5}
\]
and for \(t_0=\lceil T/2\rceil\),
\[
E_T\le q^{T-t_0}E_{t_0}+\frac{4q\,C_{\mathrm{ad}}}{(1-q)\,T^{2}}.
\tag{IV.2.6}
\]
Under T6 with \(\alpha:=\min\{p-1,\alpha_{\mathrm{loc}}\}>2\) and the bounded exit convention,
\[
\norm{e_T}_{L^{2}}\le E_T+C_{\mathrm{exit}}\,t_0^{-\alpha/2}
=
o(T^{-1/2}).
\tag{IV.2.7}
\]
If, in addition, \(\alpha>4\), then \(\norm{e_T}_{L^{2}}=\Ocal(T^{-2})\).

\subsubsection{Proof}

The argument splits into three lemmas: a stopped recursion (IV.2.1), a pathwise predictor-defect estimate under T6 (IV.2.2), and a local-to-global passage (IV.2.3).

\subsubsection{Lemma IV.2.1 (stopped first-order equilibrium-jet recursion)}

On \(\Omega_t\) one has \(\norm{e_t}\le\rho/2\) for every \(t\ge t_0\), and
\[
E_t\le qE_{t-1}+q\,\norm{R_t}_{L^{2}},
\qquad t\ge t_0+1.
\tag{IV.2.1.1}
\]

\paragraph*{Proof of Lemma IV.2.1}

\textbf{STEP 1 (predictor identity and tube invariance).}
From (IV.2.1)--(IV.2.2),
\[
\widetilde\vartheta_t-r(S_t)=e_{t-1}-R_t.
\tag{IV.2.1.2}
\]
At \(t=t_0\), the restart hypothesis gives \(\norm{e_{t_0}}\le\rho/4<\rho/2\).  Inductively assume \(\norm{e_{t-1}}\le\rho/2\) on \(\Omega_{t-1}\).  Then on \(\Omega_t\),
\[
\norm{\widetilde\vartheta_t-r(S_t)}\le\norm{e_{t-1}}+\norm{R_t}\le \frac{\rho}{2}+\rho_{\mathrm{ad}}\le \frac{3\rho}{4}<\rho,
\tag{IV.2.1.3}
\]
so the linear contraction (\ref{eq:setup:linear-corrector}) of T7 applies:
\[
\norm{e_t}=\norm{\Ccal_{\eta,\lambda,\tau,S_t}(\widetilde\vartheta_t)-r(S_t)}\le q\norm{\widetilde\vartheta_t-r(S_t)}\le q\Bigl(\frac{\rho}{2}+\rho_{\mathrm{ad}}\Bigr).
\tag{IV.2.1.4}
\]
By the choice (IV.2.3),
\[
q\Bigl(\frac{\rho}{2}+\rho_{\mathrm{ad}}\Bigr)
\le
q\,\frac{\rho}{2}+q\cdot\frac{1-q}{4q}\rho
=
\frac{1+q}{4}\rho
\le
\frac{\rho}{2},
\tag{IV.2.1.5}
\]
which closes the induction and proves the tube-invariance statement.

\textbf{STEP 2 (\(L^{2}\) recursion).}
Combine (IV.2.1.2) with the linear contraction:
\[
\norm{e_t}\le q\norm{\widetilde\vartheta_t-r(S_t)}\le q\norm{e_{t-1}}+q\norm{R_t}.
\tag{IV.2.1.6}
\]
Multiply by \(\ones_{\Omega_t}\), use \(\Omega_t\subseteq\Omega_{t-1}\), and take \(L^{2}\) norms:
\[
E_t=\norm{\ones_{\Omega_t}e_t}_{L^{2}}\le q\,E_{t-1}+q\,\norm{R_t}_{L^{2}},
\tag{IV.2.1.7}
\]
which is (IV.2.1.1). \(\square\)

\subsubsection{Lemma IV.2.2 (summable pathwise predictor defect under T6)}

Under T6 there exists a measurable event \(\mathcal A_n\) such that
\[
\Pbb(\mathcal A_n^{c})\le C_{\mathrm{exit}}'\,n^{-\alpha},
\qquad
\alpha:=\min\{\alpha_{\mathrm{loc}},p-1\}>2,
\tag{IV.2.2.1}
\]
and on \(\mathcal A_n\),
\[
S_t\in U_S
\quad\text{and}\quad
\norm{R_t}\le C_R\,\frac{C_{\mathrm{tail}}^{2}(\log t)^{2}}{t^{2}}
\qquad\text{for every }t\ge n.
\tag{IV.2.2.2}
\]

\paragraph*{Proof of Lemma IV.2.2}

By T6,
\[
\Pbb\!\bigl(\exists t\ge n:S_t\notin U_S\bigr)\le C_{\mathrm{loc}}\,n^{-\alpha_{\mathrm{loc}}}.
\tag{IV.2.2.3}
\]
The logarithmic innovation tail of T6 and a union bound give
\[
\Pbb\!\bigl(\exists t\ge n:\norm{s(Y_t)-S_{t-1}}>C_{\mathrm{tail}}\log t\bigr)
\le
\sum_{t=n}^{\infty}t^{-p}\le C_p\,n^{-(p-1)}.
\tag{IV.2.2.4}
\]
Let \(\mathcal A_n\) be the intersection of the complementary events; then \(\Pbb(\mathcal A_n^{c})\le(C_{\mathrm{loc}}+C_p)\,n^{-\alpha}\) with \(\alpha=\min\{\alpha_{\mathrm{loc}},p-1\}>2\), which is (IV.2.2.1).  On \(\mathcal A_n\), for every \(t\ge n\),
\[
\norm{S_t-S_{t-1}}=\frac{1}{t}\norm{s(Y_t)-S_{t-1}}\le\frac{C_{\mathrm{tail}}\log t}{t},
\tag{IV.2.2.5}
\]
and Proposition~\ref{app:III.2} gives \(\norm{R_t}\le C_R\norm{S_t-S_{t-1}}^{2}\le C_R C_{\mathrm{tail}}^{2}(\log t)^{2}/t^{2}\), proving (IV.2.2.2). \(\square\)

\subsubsection{Lemma IV.2.3 (local-to-global passage under bounded exit)}

There exists \(n_0\) such that \(\mathcal A_n\subseteq\Omega_T\) for every \(n\ge n_0\) and \(T\ge n\); in particular,
\[
\Pbb(\Omega_T^{c})\le C_{\mathrm{exit}}'\,t_0^{-\alpha}
\qquad\text{whenever }t_0\ge n_0.
\tag{IV.2.3.1}
\]

\paragraph*{Proof of Lemma IV.2.3}

By Lemma~\ref{app:IV.2}.\,IV.2.2, on \(\mathcal A_n\), \(\norm{R_t}\le C_R C_{\mathrm{tail}}^{2}(\log t)^{2}/t^{2}\).  Since \((\log t)^{2}/t^{2}\to 0\), choose \(n_0\) so large that
\[
C_R C_{\mathrm{tail}}^{2}\,\frac{(\log t)^{2}}{t^{2}}\le \rho_{\mathrm{ad}}
\qquad\text{for every }t\ge n_0.
\tag{IV.2.3.2}
\]
For \(n\ge n_0\), \(\mathcal A_n\) implies \(S_t\in U_S\) and \(\norm{R_t}\le\rho_{\mathrm{ad}}\) for all \(t\ge n\), which is precisely the event \(\Omega_T\) up to time \(T\).  Hence \(\Pbb(\Omega_T^{c})\le\Pbb(\mathcal A_{t_0}^{c})\le C_{\mathrm{exit}}'\,t_0^{-\alpha}\). \(\square\)

\par\medskip
\noindent\textit{Conclusion of Theorem~\ref{app:IV.2}.}

Lemma IV.2.1 gives the stopped recursion (IV.2.1.1), and Proposition~\ref{app:III.2} bounds \(\norm{R_t}_{L^{2}}\le C_{\mathrm{ad}}/t^{2}\) with \(C_{\mathrm{ad}}:=4C_R\norm{s(Y_1)}_{L^{4}}^{2}\), proving (IV.2.5).  Iterating from \(t_0\) to \(T\),
\[
E_T\le q^{T-t_0}E_{t_0}+q\,C_{\mathrm{ad}}\sum_{s=t_0+1}^{T}q^{T-s}s^{-2}.
\tag{IV.2.8}
\]
For \(t_0=\lceil T/2\rceil\), \(s\ge T/2\) throughout the sum, so
\[
\sum_{s=t_0+1}^{T}q^{T-s}s^{-2}\le \frac{4}{T^{2}}\sum_{k=0}^{\infty}q^{k}=\frac{4}{(1-q)\,T^{2}},
\tag{IV.2.9}
\]
which proves (IV.2.6).

For the full \(L^{2}\) error, decompose \(e_T=\ones_{\Omega_T}e_T+\ones_{\Omega_T^{c}}e_T\).  By the bounded exit convention and Lemma IV.2.3,
\[
\norm{\ones_{\Omega_T^{c}}e_T}_{L^{2}}\le C_{\mathrm{env}}\,\Pbb(\Omega_T^{c})^{1/2}\le C_{\mathrm{env}}(C_{\mathrm{exit}}')^{1/2}\,t_0^{-\alpha/2}.
\tag{IV.2.10}
\]
Setting \(C_{\mathrm{exit}}:=C_{\mathrm{env}}(C_{\mathrm{exit}}')^{1/2}\) gives the first inequality in (IV.2.7); taking \(t_0=\lceil T/2\rceil\) yields
\[
\norm{e_T}_{L^{2}}\le q^{T-t_0}E_{t_0}+\frac{4q\,C_{\mathrm{ad}}}{(1-q)\,T^{2}}+C_{\mathrm{exit}}\,2^{\alpha/2}\,T^{-\alpha/2}.
\tag{IV.2.11}
\]
Since \(\alpha>2\), the right-hand side is \(o(T^{-1/2})\), proving (IV.2.7).  If \(\alpha>4\), then \(T^{-\alpha/2}=\Ocal(T^{-2})\), giving the optional \(\Ocal(T^{-2})\) full-error rate. \(\square\)

\par\medskip\hrule\medskip\par

\subsection{Proposition IV.3 (frozen local contraction for the exact damped Newton corrector)}
\label{app:IV.3}

\textbf{Claim.}
Assume there exist a statistic neighborhood \(U_S\), a radius \(\rho>0\), constants \(0<\mu\le L<\infty\) and \(C_\Psi<\infty\) such that for every \(S\in U_S\) and \(\norm{\vartheta-r(S)}\le\rho\),
\[
\mu I\preceq\Hcal(\vartheta,S)\preceq L I,
\tag{IV.3.1}
\]
and the exact damped Newton map \citep{nocedal2006numerical}
\[
\Ccal^{\mathrm{ex}}_{\eta,\lambda,S}(\vartheta):=\vartheta-\eta(\Hcal(\vartheta,S)+\lambda I)^{-1}F(\vartheta,S)
\tag{IV.3.2}
\]
is \(C^{2}\) with \(\sup\norm{D_\vartheta^{2}\Ccal^{\mathrm{ex}}_{\eta,\lambda,S}(\vartheta)}_{\op}\le C_\Psi\).
Define
\[
q_0:=\sup_{\kappa\in[\mu,L]}\Bigl|1-\eta\,\frac{\kappa}{\kappa+\lambda}\Bigr|.
\tag{IV.3.4}
\]
If \(q_0<1\), then for every \(q\in(q_0,1)\) there exists \(\rho_q\in(0,\rho]\) such that
\[
\norm{\Ccal^{\mathrm{ex}}_{\eta,\lambda,S}(r(S)+u)-r(S)}\le q\,\norm{u}
\qquad\text{for every }S\in U_S,\ \norm{u}\le\rho_q.
\tag{IV.3.5}
\]

\subsubsection{Proof}

Since \(F(r(S),S)=0\), the chain rule at \(\vartheta=r(S)\) gives \(D_\vartheta\Ccal^{\mathrm{ex}}_{\eta,\lambda,S}(r(S))=I-\eta(\Jcal(S)+\lambda I)^{-1}\Jcal(S)\), whose operator norm is at most \(q_0\) by the spectral inclusion \(\mathrm{spec}\,\Jcal(S)\subset[\mu,L]\).  Taylor's theorem with integral remainder and the \(C^{2}\) envelope (IV.3.1) give
\[
\norm{\Ccal^{\mathrm{ex}}_{\eta,\lambda,S}(r(S)+u)-r(S)}\le q_0\norm{u}+\tfrac{1}{2}C_\Psi\norm{u}^{2}.
\tag{IV.3.9}
\]
Choosing \(\rho_q\) so that \(q_0+\tfrac{1}{2}C_\Psi\rho_q\le q\) yields (IV.3.5). \(\square\)

\par\medskip\hrule\medskip\par

\subsection{Proposition IV.4 (frozen local contraction for the inexact damped Newton--CG corrector)}
\label{app:IV.4}

\textbf{Claim.}
Assume the hypotheses of Proposition~\ref{app:IV.3} and that the inexact corrector \citep{dembo1982inexact,eisenstat1996choosing,nocedal2006numerical}
\[
\Ccal_{\eta,\lambda,\tau,S}(\vartheta):=\vartheta-\eta\,\widehat u(\vartheta,S)
\tag{IV.4.1}
\]
satisfies
\[
(\Hcal(\vartheta,S)+\lambda I)\widehat u(\vartheta,S)=F(\vartheta,S)+r_{\lin}(\vartheta,S),
\qquad
\norm{r_{\lin}(\vartheta,S)}\le \tau\,\norm{F(\vartheta,S)},
\tag{IV.4.2}
\]
with a local Lipschitz envelope \(\norm{F(r(S)+u,S)}\le L_F\norm{u}\).
If
\[
q:=q_{\mathrm{ex}}+\eta\,\tau\,\frac{L_F}{\mu+\lambda}<1
\tag{IV.4.4}
\]
for some contraction factor \(q_{\mathrm{ex}}\in(0,1)\) of Proposition~\ref{app:IV.3}, then after shrinking the radius if necessary
\[
\norm{\Ccal_{\eta,\lambda,\tau,S}(r(S)+u)-r(S)}\le q\norm{u}
\qquad\text{for every }S\in U_S,\ \norm{u}\le\rho_q.
\tag{IV.4.5}
\]

\subsubsection{Proof}

Let \(u^{\mathrm{ex}}:=(\Hcal+\lambda I)^{-1}F\) be the exact direction.  From (IV.4.2), \(\widehat u-u^{\mathrm{ex}}=(\Hcal+\lambda I)^{-1}r_{\lin}\), so
\[
\norm{\widehat u(r(S)+u,S)-u^{\mathrm{ex}}(r(S)+u,S)}
\le
\frac{\tau}{\mu+\lambda}\norm{F(r(S)+u,S)}
\le
\frac{\tau L_F}{\mu+\lambda}\norm{u}.
\tag{IV.4.8}
\]
By triangle inequality and Proposition~\ref{app:IV.3},
\[
\norm{\Ccal_{\eta,\lambda,\tau,S}(r(S)+u)-r(S)}\le q_{\mathrm{ex}}\norm{u}+\eta\,\frac{\tau L_F}{\mu+\lambda}\norm{u}=q\norm{u}.  \quad\square
\]

\par\medskip\hrule\medskip\par

\subsection{Theorem IV.5 (batch-to-online transfer)}
\label{app:IV.5}

\textbf{Claim.}
Assume Theorem~\ref{app:III.3} and that an online algorithm produces \(\vartheta_T\) with
\[
\norm{\vartheta_T-r(S_T)}_{L^{2}}=o(T^{-1/2}).
\tag{IV.5.1}
\]
Then
\[
\sqrt{T}\,\vartheta_T\Rightarrow\N(0,\Vcal_{*}),
\qquad
\E\,d_{h}^{2}(\Sigma_T,\Sigma_{\bullet})=\frac{1}{T}\tr(\Vcal_{*})+o(T^{-1}).
\tag{IV.5.2}
\]

\subsubsection{Proof}

Set \(r_T:=\vartheta_T-r(S_T)\); by (IV.5.1), \(\sqrt{T}\,r_T\to 0\) in \(L^{2}\), hence in probability.  Theorem~\ref{app:III.3} gives \(\sqrt{T}\,r(S_T)\Rightarrow\N(0,\Vcal_{*})\), so
\[
\sqrt{T}\,\vartheta_T=\sqrt{T}\,r(S_T)+o_{\Pbb}(1),
\tag{IV.5.3}
\]
and Slutsky's theorem \citep{van2000asymptotic} yields the central limit assertion in (IV.5.2).

For the risk identity, in the normal chart \(d_{h}(\Sigma_T,\Sigma_{\bullet})=\norm{\vartheta_T}\) and \(d_{h}(\Sigmacirc(S_T),\Sigma_{\bullet})=\norm{r(S_T)}\); thus
\[
T\,\E\norm{\vartheta_T}^{2}
=
T\,\E\norm{r(S_T)+r_T}^{2}
=
T\,\E\norm{r(S_T)}^{2}
+
2T\,\E\inner{r(S_T)}{r_T}
+
T\,\E\norm{r_T}^{2}.
\tag{IV.5.4}
\]
The first term tends to \(\tr(\Vcal_{*})\) by Theorem~\ref{app:III.3}, and the third term to \(0\) by (IV.5.1).  By Cauchy--Schwarz,
\[
\bigl|2T\,\E\inner{r(S_T)}{r_T}\bigr|\le 2\bigl(T\,\E\norm{r(S_T)}^{2}\bigr)^{1/2}\bigl(T\,\E\norm{r_T}^{2}\bigr)^{1/2}\to 0.
\tag{IV.5.5}
\]
Hence \(T\,\E\norm{\vartheta_T}^{2}\to\tr(\Vcal_{*})\), which is the risk identity in (IV.5.2). \(\square\)

\par\medskip\hrule\medskip\par

\section{Higher-order predictors and order-\texorpdfstring{\(\nu\)}{v} correctors}
\label{app:V}

\paragraph{Local hypotheses.}
The frozen corrector is denoted \(M_S\); for the inexact damped Newton--CG corrector of Section~\ref{subsec:method:corrector} the role of \(M_S\) is played by \(\Ccal_{\eta,\lambda,\tau,S}\).  The hypotheses below specialize to predictor order~\(m\) and corrector order~\(\nu\):
\begin{itemize}
\item D1(\(m\)): there is a statistic neighborhood \(U_S\) of \(S_*\) on which \(r\in C^{m+1}(U_S;E)\), with \(M_{r,m+1}:=\sup_{S\in U_S}\norm{D^{m+1}r(S)}_{\op}<\infty\);
\item D2(\(m,\nu\)): there is \(C_{\Delta,m,\nu}<\infty\) with \(\norm{\Delta S_t}_{L^{2\nu(m+1)}}\le C_{\Delta,m,\nu}/t\) for \(t\ge 2\);
\item D3(\(q\)): linear-contraction case~\eqref{eq:setup:linear-corrector} of Assumption~\ref{assump:local-corrector};
\item D4(\(\nu\)): order-\(\nu\) case~\eqref{eq:setup:higher-order-corrector} of Assumption~\ref{assump:local-corrector}, i.e.\ \(\norm{M_S(r(S)+u)-r(S)}\le C_\nu\norm{u}^{\nu}\) on \(\norm{u}\le\rho_\nu\);
\item D5(\(k\)): there are \(C_{\loc,k}>0\) and \(\beta_k>1\) with \(\Pbb(\Omega_T^{c})\le C_{\loc,k}T^{-\beta_k}\), and the bounded-exit convention \(\norm{e_T}\le C_{\mathrm{env}}\) on \(\Omega_T^{c}\) (Assumption~\ref{assump:localization}).
\end{itemize}

\par\medskip\hrule\medskip\par

\subsection{Proposition V.1 (jet predictor and Taylor remainder)}
\label{app:V.1}

\textbf{Claim.}
Fix an integer \(m\ge 0\), assume D1(\(m\)), and let \(\Pi_t^{(m)}\) and \(R_t^{(m)}\) be defined by~\eqref{eq:method:jet-predictor} and~\eqref{eq:method:jet-remainder}.  Then for every \(t\) with \(S_{t-1},S_t\in U_S\),
\[
R_t^{(m)}
=
\frac{1}{m!}\int_0^1(1-\sigma)^{m}D^{m+1}r(S_{t-1}+\sigma\Delta S_t)\bigl[\Delta S_t^{\otimes(m+1)}\bigr]\,d\sigma,
\tag{V.1.1}
\]
and
\[
\norm{R_t^{(m)}}\le \frac{M_{r,m+1}}{(m+1)!}\,\norm{\Delta S_t}^{m+1}.
\tag{V.1.2}
\]
Moreover,
\[
\Pi_t^{(m)}(\vartheta_{t-1})-r(S_t)=e_{t-1}-R_t^{(m)},
\tag{V.1.3}
\]
and consequently
\[
\norm{\Pi_t^{(m)}(\vartheta_{t-1})-r(S_t)}\le \norm{e_{t-1}}+\norm{R_t^{(m)}}.
\tag{V.1.4}
\]

\subsubsection{Proof}

The multivariate Taylor theorem with integral remainder \citep{nocedal2006numerical,dontchev2014implicit} applied to the \(C^{m+1}\) map~\(r\) at \(S_{t-1}\) gives (V.1.1); taking norms and using \(\int_0^1(1-\sigma)^{m}d\sigma=1/(m+1)\) gives (V.1.2).  The identity (V.1.3) follows from the jet predictor definition: \(\Pi_t^{(m)}(\vartheta_{t-1})-r(S_t)=[\vartheta_{t-1}-r(S_{t-1})]-[r(S_t)-r(S_{t-1})-\sum_{k=1}^{m}\tfrac{1}{k!}D^kr(S_{t-1})[\Delta S_t^{\otimes k}]]=e_{t-1}-R_t^{(m)}\). \(\square\)

\par\medskip\hrule\medskip\par

\subsection{Proposition V.2 (jet remainder in \(L^{2\nu}\))}
\label{app:V.2}

\textbf{Claim.}
Assume D1(\(m\)) and D2(\(m,\nu\)).  Then
\[
\norm{R_t^{(m)}}_{L^{2\nu}}
\le
\frac{M_{r,m+1}}{(m+1)!}\,\norm{\Delta S_t}_{L^{2\nu(m+1)}}^{m+1}
\le
\frac{C_{R,m,\nu}}{t^{m+1}},
\qquad t\ge 2,
\tag{V.2.1}
\]
where
\[
C_{R,m,\nu}:=\frac{M_{r,m+1}}{(m+1)!}\,C_{\Delta,m,\nu}^{\,m+1}.
\tag{V.2.2}
\]

\subsubsection{Proof}

By Proposition~\ref{app:V.1}, \(\norm{R_t^{(m)}}\le M_{r,m+1}\norm{\Delta S_t}^{m+1}/(m+1)!\) pathwise.  Taking the \(L^{2\nu}\)-norm and using \(\bigl\|\norm{X}^{m+1}\bigr\|_{L^{2\nu}}=\norm{X}_{L^{2\nu(m+1)}}^{m+1}\) gives the first inequality.  The running-average identity \(\Delta S_t=t^{-1}(s(Y_t)-S_{t-1})\), Jensen's inequality, and stationarity give \(\norm{\Delta S_t}_{L^{p}}\le 2\norm{s(Y_1)}_{L^{p}}/t\); substituting D2(\(m,\nu\)) yields the second inequality. \(\square\)

\par\medskip\hrule\medskip\par

\subsection{Theorem V.3 (\(m\)-th predictor with linear corrector)}
\label{app:V.3}

\textbf{Claim.}
Assume D1(\(m\)), D2(\(m,1\)), D3(\(q\)), and Assumption~\ref{assump:localization}.  Consider the predictor--corrector recursion \citep{simonetto2016class,simonetto2020time}
\[
\widetilde\vartheta_t=\Pi_t^{(m)}(\vartheta_{t-1}),
\qquad
\vartheta_t=M_{S_t}(\widetilde\vartheta_t),
\qquad t\ge t_1+1,
\tag{V.3.1}
\]
with localization event
\[
\Omega_t^{(m)}:=\Bigl\{S_s\in U_S\text{ and }\norm{R_s^{(m)}}\le \tfrac{\rho}{4}\text{ for all }s=t_1+1,\ldots,t\Bigr\},
\qquad t\ge t_1,
\tag{V.3.2}
\]
and stopped error \(E_t^{(m)}:=\norm{\ones_{\Omega_t^{(m)}}\,e_t}_{L^{2}}\).  Assume the restart \(\norm{e_{t_1}}\le \rho/4\) almost surely on \(\Omega_{t_1}^{(m)}\).  Then on \(\Omega_t^{(m)}\),
\[
\norm{e_t}\le \rho
\qquad t\ge t_1,
\tag{V.3.3}
\]
and
\[
E_t^{(m)}\le qE_{t-1}^{(m)}+\frac{q\,C_{R,m,1}}{t^{m+1}},
\qquad t\ge t_1+1.
\tag{V.3.4}
\]
For \(t_1=\lceil T/2\rceil\),
\[
E_T^{(m)}\le q^{T-t_1}E_{t_1}^{(m)}+\frac{C_{m,1}}{T^{m+1}},
\qquad
C_{m,1}:=\frac{2^{m+1}q}{1-q}\,C_{R,m,1}.
\tag{V.3.5}
\]

\subsubsection{Proof}

By Proposition~\ref{app:V.1}, \(\widetilde\vartheta_t-r(S_t)=e_{t-1}-R_t^{(m)}\).  The restart gives \(\norm{e_{t_1}}\le\rho/4\); inductively assume \(\norm{e_{t-1}}\le\rho/2\) on \(\Omega_{t-1}^{(m)}\).  On \(\Omega_t^{(m)}\), \(\norm{\widetilde\vartheta_t-r(S_t)}\le\rho/2+\rho/4<\rho\), so D3(\(q\)) applies and \(\norm{e_t}\le q(3\rho/4)<\rho/2\), closing the induction.  The \(L^{2}\) recursion \(E_t^{(m)}\le qE_{t-1}^{(m)}+q\norm{R_t^{(m)}}_{L^{2}}\) follows; Proposition~\ref{app:V.2} at \(\nu=1\) gives (V.3.4).  Iterating with \(t_1=\lceil T/2\rceil\) and using \(\sum_{s>T/2}q^{T-s}s^{-(m+1)}\le 2^{m+1}/((1-q)T^{m+1})\) gives (V.3.5). \(\square\)

\par\medskip\hrule\medskip\par

\subsection{Theorem V.4 (\(m\)-th predictor with order-\(\nu\) corrector)}
\label{app:V.4}

\textbf{Claim.}
Assume D1(\(m\)), D2(\(m,\nu\)) with \(\nu\ge 2\), D4(\(\nu\)), and Assumption~\ref{assump:localization}.  Choose \(0<\rho\le\rho_\nu\) so small that
\[
q_\nu:=2^{\nu-1}C_\nu\rho^{\nu-1}<1.
\tag{V.4.1}
\]
Consider the recursion (V.3.1) with localization event
\[
\Omega_t^{(m,\nu)}:=\Bigl\{S_s\in U_S\text{ and }\norm{R_s^{(m)}}\le \tfrac{\rho}{4}\text{ for all }s=t_1+1,\ldots,t\Bigr\},
\tag{V.4.3}
\]
and stopped error \(E_t^{(m,\nu)}:=\norm{\ones_{\Omega_t^{(m,\nu)}}\,e_t}_{L^{2}}\).  Assume the restart \(\norm{e_{t_1}}\le \rho/4\) a.s.\ on \(\Omega_{t_1}^{(m,\nu)}\).  Then on \(\Omega_t^{(m,\nu)}\), \(\norm{e_t}\le \rho\) for \(t\ge t_1\), and
\[
E_t^{(m,\nu)}\le q_\nu E_{t-1}^{(m,\nu)}+\frac{2^{\nu-1}C_\nu\,C_{R,m,\nu}^{\nu}}{t^{\nu(m+1)}},
\qquad t\ge t_1+1.
\tag{V.4.5}
\]
For \(t_1=\lceil T/2\rceil\),
\[
E_T^{(m,\nu)}\le q_\nu^{\,T-t_1}E_{t_1}^{(m,\nu)}+\frac{C_{m,\nu}}{T^{\nu(m+1)}},
\qquad
C_{m,\nu}:=\frac{2^{\nu(m+2)-1}}{1-q_\nu}\,C_\nu C_{R,m,\nu}^{\nu}.
\tag{V.4.6}
\]

\subsubsection{Proof}

\textbf{STEP 1 (tube invariance).}
By Proposition~\ref{app:V.1}, \(\Pi_t^{(m)}(\vartheta_{t-1})-r(S_t)=e_{t-1}-R_t^{(m)}\).  At \(t=t_1\), the restart gives \(\norm{e_{t_1}}\le\rho/4\).  Inductively assume \(\norm{e_{t-1}}\le\rho/2\) on \(\Omega_{t-1}^{(m,\nu)}\).  On \(\Omega_t^{(m,\nu)}\),
\[
\norm{\widetilde\vartheta_t-r(S_t)}\le\frac{\rho}{2}+\frac{\rho}{4}=\frac{3\rho}{4}<\rho.
\tag{V.4.7}
\]
The order-\(\nu\) bound D4(\(\nu\)) then gives
\[
\norm{e_t}\le C_\nu\Bigl(\frac{3\rho}{4}\Bigr)^{\nu}\le C_\nu\rho^{\nu-1}\cdot\frac{3\rho}{4}\le 2^{1-\nu}\cdot\frac{3\rho}{4}\le\frac{3\rho}{8}<\frac{\rho}{2},
\tag{V.4.8}
\]
where the inequality \(C_\nu\rho^{\nu-1}\le 2^{1-\nu}\) is a consequence of (V.4.1).  This closes the induction and proves (V.4.4).

\textbf{STEP 2 (one-step recursion).}
Still on \(\Omega_t^{(m,\nu)}\), the order-\(\nu\) corrector estimate gives
\[
\norm{e_t}\le C_\nu\bigl(\norm{e_{t-1}}+\norm{R_t^{(m)}}\bigr)^{\nu}.
\tag{V.4.9}
\]
Using the elementary inequality \((a+b)^{\nu}\le 2^{\nu-1}(a^{\nu}+b^{\nu})\) for \(a,b\ge 0\) and the pathwise bound \(\norm{e_{t-1}}\le\rho\),
\[
\begin{aligned}
\norm{e_t}
&\le 2^{\nu-1}C_\nu\bigl(\norm{e_{t-1}}^{\nu}+\norm{R_t^{(m)}}^{\nu}\bigr)\\
&\le 2^{\nu-1}C_\nu\rho^{\nu-1}\,\norm{e_{t-1}}+2^{\nu-1}C_\nu\,\norm{R_t^{(m)}}^{\nu}\\
&=q_\nu\,\norm{e_{t-1}}+2^{\nu-1}C_\nu\,\norm{R_t^{(m)}}^{\nu}.
\end{aligned}
\tag{V.4.10}
\]

\textbf{STEP 3 (stopped \(L^{2}\) recursion).}
Multiplying (V.4.10) by \(\ones_{\Omega_t^{(m,\nu)}}\), using \(\Omega_t^{(m,\nu)}\subseteq\Omega_{t-1}^{(m,\nu)}\), and taking \(L^{2}\)-norms gives
\[
E_t^{(m,\nu)}\le q_\nu E_{t-1}^{(m,\nu)}+2^{\nu-1}C_\nu\,\bigl\|\norm{R_t^{(m)}}^{\nu}\bigr\|_{L^{2}}.
\tag{V.4.11}
\]
The identity \(\bigl\|\norm{X}^{\nu}\bigr\|_{L^{2}}=\norm{X}_{L^{2\nu}}^{\nu}\) and Proposition~\ref{app:V.2} yield
\[
\bigl\|\norm{R_t^{(m)}}^{\nu}\bigr\|_{L^{2}}=\norm{R_t^{(m)}}_{L^{2\nu}}^{\nu}\le \frac{C_{R,m,\nu}^{\nu}}{t^{\nu(m+1)}},
\tag{V.4.12}
\]
so (V.4.5) follows.

\textbf{STEP 4 (terminal sum).}
Iterating (V.4.5) from \(t_1\) to \(T\),
\[
E_T^{(m,\nu)}
\le
q_\nu^{T-t_1}E_{t_1}^{(m,\nu)}
+
2^{\nu-1}C_\nu C_{R,m,\nu}^{\nu}\sum_{s=t_1+1}^{T}q_\nu^{T-s}\,s^{-\nu(m+1)}.
\tag{V.4.13}
\]
For \(t_1=\lceil T/2\rceil\), each \(s\) satisfies \(s\ge T/2\), so
\[
\sum_{s=t_1+1}^{T}q_\nu^{T-s}\,s^{-\nu(m+1)}
\le
\frac{2^{\nu(m+1)}}{T^{\nu(m+1)}}\sum_{k=0}^{\infty}q_\nu^{k}
=
\frac{2^{\nu(m+1)}}{(1-q_\nu)\,T^{\nu(m+1)}}.
\tag{V.4.14}
\]
Substituting into (V.4.13) gives (V.4.6) with \(C_{m,\nu}=2^{\nu-1}\cdot 2^{\nu(m+1)}/(1-q_\nu)\,C_\nu C_{R,m,\nu}^{\nu}=2^{\nu(m+2)-1}/(1-q_\nu)\,C_\nu C_{R,m,\nu}^{\nu}\). \(\square\)

\par\medskip\hrule\medskip\par

\subsection{Proposition V.5 (exact full Newton is order-\(2\))}
\label{app:V.5}

\textbf{Claim.}
Under Assumptions~\ref{assump:local-branch} and~\ref{assump:local-corrector} the exact full Newton map \citep{nocedal2006numerical}
\[
N_S(\vartheta):=\vartheta-\Bigl[D_\vartheta F(\vartheta,S)\Bigr]^{-1}F(\vartheta,S)
\tag{V.5.1}
\]
is well defined on a neighborhood where \(D_\vartheta F\) is invertible, and there exist \(M_J,L_J,\rho_N>0\) and a constant \(C_N:=\tfrac{1}{2}M_J L_J\) such that
\[
\norm{N_S(r(S)+u)-r(S)}\le C_N\norm{u}^{2},
\qquad \norm{u}\le\rho_N,\ S\in U_S.
\tag{V.5.2}
\]

\subsubsection{Proof}

Let \(M_J:=\sup\norm{D_\vartheta F(\vartheta,S)^{-1}}_{\op}\) and \(L_J\) be the Lipschitz constant of \(\vartheta\mapsto D_\vartheta F(\vartheta,S)\), both finite after shrinking the neighborhood.  Since \(F(r(S),S)=0\), the fundamental theorem of calculus gives \(F(r(S)+u,S)=\int_0^1 D_\vartheta F(r(S)+\sigma u,S)\,u\,d\sigma\), so
\[
N_S(r(S)+u)-r(S)=[D_\vartheta F(r(S)+u,S)]^{-1}\int_0^1[D_\vartheta F(r(S)+u,S)-D_\vartheta F(r(S)+\sigma u,S)]\,u\,d\sigma.
\]
Taking norms: \(\norm{D_\vartheta F(r(S)+u,S)-D_\vartheta F(r(S)+\sigma u,S)}_{\op}\le L_J(1-\sigma)\norm{u}\), and integrating gives (V.5.2) with \(C_N=\tfrac{1}{2}M_JL_J\). \(\square\)

\par\medskip\hrule\medskip\par

\subsection{Proposition V.6 (inexact Newton with quadratic linear residual is order-\(2\))}
\label{app:V.6}

\textbf{Claim.}
Assume the hypotheses of Proposition~\ref{app:V.5}.  Suppose \(\widehat u(\vartheta,S)\) satisfies \citep{dembo1982inexact,eisenstat1996choosing}
\[
D_\vartheta F(\vartheta,S)\,\widehat u(\vartheta,S)=F(\vartheta,S)+r_{\lin}(\vartheta,S),
\qquad
\norm{r_{\lin}(\vartheta,S)}\le c_{\lin}\,\norm{F(\vartheta,S)}^{2},
\tag{V.6.1}
\]
and define \(\widehat N_S(\vartheta):=\vartheta-\widehat u(\vartheta,S)\).  Then there exist \(C_{\mathbf{inN}},\rho_{\mathbf{inN}}>0\) such that
\[
\norm{\widehat N_S(r(S)+u)-r(S)}\le C_{\mathbf{inN}}\,\norm{u}^{2},
\qquad \norm{u}\le\rho_{\mathbf{inN}},\ S\in U_S.
\tag{V.6.3}
\]
Explicitly, \(C_{\mathbf{inN}}=C_N+M_J c_{\lin}L_F^{2}\) where \(L_F:=\sup\norm{D_\vartheta F(\vartheta,S)}_{\op}\).

\subsubsection{Proof}

From (V.6.1), \(\widehat u-u^{\mathrm{ex}}=[D_\vartheta F]^{-1}r_{\lin}\), so \(\norm{\widehat u-u^{\mathrm{ex}}}\le M_Jc_{\lin}\norm{F}^{2}\le M_Jc_{\lin}L_F^{2}\norm{u}^{2}\), where \(\norm{F(r(S)+u,S)}\le L_F\norm{u}\) by mean-value.  Decomposing \(\widehat N_S(\vartheta)-r(S)=(N_S(\vartheta)-r(S))-(\widehat u-u^{\mathrm{ex}})\) and combining with Proposition~\ref{app:V.5} via triangle inequality gives (V.6.3). \(\square\)

\par\medskip\hrule\medskip\par

\subsection{Proposition V.7 (fixed damping or fixed relative residual is only order-\(1\))}
\label{app:V.7}

\textbf{Claim.}
Two negative results.

\noindent(1) For fixed \(\lambda>0\), the damped Newton map
\[
N_S^{\lambda}(\vartheta):=\vartheta-\bigl[D_\vartheta F(\vartheta,S)+\lambda I\bigr]^{-1}F(\vartheta,S)
\tag{V.7.1}
\]
satisfies
\[
D_\vartheta N_S^{\lambda}(r(S))=I-\bigl[\Jcal(S)+\lambda I\bigr]^{-1}\Jcal(S)\neq 0,
\tag{V.7.2}
\]
so \(N_S^{\lambda}\) is in general only an order-\(1\) corrector in the sense of D3.

\noindent(2) If \(\widehat u(\vartheta,S)\) satisfies \citep{dembo1982inexact,eisenstat1996choosing,nocedal2006numerical} \(\norm{r_{\lin}}\le \tau\norm{F}\) with a fixed \(\tau>0\), then the resulting inexact corrector is only order-\(1\) and Proposition~\ref{app:IV.4} gives the corresponding linear contraction factor.

\subsubsection{Proof}

Part (1): differentiating (V.7.1) at \(\vartheta=r(S)\) and using \(F(r(S),S)=0\) gives (V.7.2); the eigenvalues are \(\lambda/(\kappa+\lambda)\neq 0\) for \(\lambda>0\).  Part (2): the perturbation \(\norm{\widehat u-u^{\mathrm{ex}}}=\Ocal(\norm{F})=\Ocal(\norm{u})\) is first-order, so the map is at best a linear contraction (Proposition~\ref{app:IV.4}). \(\square\)

\par\medskip\hrule\medskip\par

\subsection{Proposition V.8 (local-to-full passage)}
\label{app:V.8}

\textbf{Claim.}
Assume D5(\(k\)) and a localized estimate of the form
\[
\norm{\ones_{\Omega_T}\,e_T}_{L^{2}}\le C_{\loc}T^{-k}+C_{\exp}\,e^{-c(T-t_1)}
\tag{V.8.1}
\]
for some \(k>0\) and constants \(C_{\loc},C_{\exp},c>0\).  Then
\[
\norm{e_T}_{L^{2}}\le C_{\loc}\,T^{-k}+C_{\exp}\,e^{-c(T-t_1)}+C_{\mathrm{env}}\,C_{\loc,k}^{1/2}\,T^{-\beta_k/2}.
\tag{V.8.2}
\]
In particular:
\begin{itemize}
\item if \(\beta_k>2k\), then \(\norm{e_T}_{L^{2}}=\Ocal(T^{-k})+\Ocal(e^{-c(T-t_1)})\);
\item if \(k>1/2\) and merely \(\beta_k>1\), then \(\norm{e_T}_{L^{2}}=o(T^{-1/2})\), and Theorem~\ref{app:IV.5} applies.
\end{itemize}

\subsubsection{Proof}

Decompose \(e_T=\ones_{\Omega_T}e_T+\ones_{\Omega_T^{c}}e_T\) and apply Minkowski's inequality.  The bounded-exit convention and D5(\(k\)) give
\[
\norm{\ones_{\Omega_T^{c}}e_T}_{L^{2}}\le C_{\mathrm{env}}\,\Pbb(\Omega_T^{c})^{1/2}\le C_{\mathrm{env}}\,C_{\loc,k}^{1/2}\,T^{-\beta_k/2}.
\]
Combining with (V.8.1) gives (V.8.2).  If \(\beta_k>2k\), the exit term is \(o(T^{-k})\); if \(k>1/2\) and \(\beta_k>1\), all three terms are \(o(T^{-1/2})\). \(\square\)

\par\medskip\hrule\medskip\par

\section{Latent Gaussian explicit constants}
\label{app:VI}

\paragraph{Notation.}
Throughout this appendix we work in the correctly specified latent linear Gaussian model
\[
Y_{t}\sim\N(0,K_{*}),
\qquad
K_{*}=H\Sigma_{*}H^{\top}+R,
\]
with \(H\in\R^{p\times d}\) of full column rank and \(\Sigma_{*}\in\Sympp(d)\) of simple spectrum \(\lambda_{1}>\cdots>\lambda_{d}>0\).  The compressed observation, statistic, and lift are
\[
W:=R^{-1}H,
\quad
G:=H^{\top}W,
\quad
z_{t}:=W^{\top}Y_{t},
\quad
D_{t}:=t^{-1}\sum_{r=1}^{t}z_{r}z_{r}^{\top},
\quad
C_{\Sigma}:=(\Sigma^{-1}+G)^{-1},
\quad
A(\Sigma;D):=C_{\Sigma}+C_{\Sigma}DC_{\Sigma},
\]
with population mean \(D_{*}:=\E[z_{1}z_{1}^{\top}]=G\Sigma_{*}G+G\).  The chart-norm constants are \(\underline\gamma_{h}:=\min\{\lambda_{\min}^{2},\delta_{\min}^{2}\}\) and \(\overline\gamma_{h}:=\lambda_{\max}^{2}\) with \(\delta_{\min}:=\min_{i<j}(\lambda_{i}-\lambda_{j})>0\); the target-branch envelope constants \(L_{\vartheta},C_{R},r_{D}>0\) are those of Proposition~\ref{app:III.1}.

\par\medskip\hrule\medskip\par

\subsection{Proposition VI.1 (Woodbury compression of the latent-scatter lift)}
\label{app:VI.1}

\textbf{Claim.}
For every \(\Sigma\in\Ucal\) and every \(S\in\Sym(p)\),
\[
V_{\Sigma}:=\Sigma-\Sigma H^{\top}K_{\Sigma}^{-1}H\Sigma=C_{\Sigma},
\qquad
M_{\Sigma}:=\Sigma H^{\top}K_{\Sigma}^{-1}=C_{\Sigma}H^{\top}R^{-1},
\tag{VI.1.1}
\]
\[
M_{\Sigma}SM_{\Sigma}^{\top}=C_{\Sigma}\,\pi(S)\,C_{\Sigma},
\qquad
\pi(S):=H^{\top}R^{-1}SR^{-1}H,
\tag{VI.1.2}
\]
and consequently
\[
A(\Sigma;S):=V_{\Sigma}+M_{\Sigma}SM_{\Sigma}^{\top}=C_{\Sigma}+C_{\Sigma}\,\pi(S)\,C_{\Sigma}.
\tag{VI.1.3}
\]
The compressed running statistic satisfies
\[
D_{t}=\pi(S_{t})=\frac{t-1}{t}D_{t-1}+\frac{1}{t}z_{t}z_{t}^{\top},
\qquad t\ge 2.
\tag{VI.1.4}
\]

\subsubsection{Proof}

\textbf{STEP 1 (Woodbury for \(K_{\Sigma}^{-1}\)).}
The matrix inversion lemma applied to \(K_{\Sigma}=H\Sigma H^{\top}+R\) with respect to the inner factor \(\Sigma\) gives
\[
K_{\Sigma}^{-1}=R^{-1}-R^{-1}H\bigl(\Sigma^{-1}+H^{\top}R^{-1}H\bigr)^{-1}H^{\top}R^{-1}=R^{-1}-R^{-1}HC_{\Sigma}H^{\top}R^{-1}.
\tag{VI.1.5}
\]

\textbf{STEP 2 (covariance form: \(V_{\Sigma}=C_{\Sigma}\)).}
The matrix inversion lemma in covariance form gives
\[
\bigl(\Sigma^{-1}+H^{\top}R^{-1}H\bigr)^{-1}=\Sigma-\Sigma H^{\top}\bigl(H\Sigma H^{\top}+R\bigr)^{-1}H\Sigma=V_{\Sigma},
\tag{VI.1.6}
\]
so \(V_{\Sigma}=(\Sigma^{-1}+G)^{-1}=C_{\Sigma}\), which is the first half of (VI.1.1).

\textbf{STEP 3 (factorization of \(M_{\Sigma}\)).}
From (VI.1.5),
\[
\Sigma H^{\top}K_{\Sigma}^{-1}=\Sigma H^{\top}R^{-1}-\Sigma H^{\top}R^{-1}HC_{\Sigma}H^{\top}R^{-1}.
\tag{VI.1.7}
\]
The defining identity \(C_{\Sigma}(\Sigma^{-1}+G)=I\) can be rewritten as \(C_{\Sigma}=\Sigma-\Sigma G C_{\Sigma}\), and multiplying on the right by \(H^{\top}R^{-1}\) yields the key intermediate
\[
C_{\Sigma}H^{\top}R^{-1}=\Sigma H^{\top}R^{-1}-\Sigma H^{\top}R^{-1}H\,C_{\Sigma}H^{\top}R^{-1}.
\tag{VI.1.8}
\]
Comparing (VI.1.7) and (VI.1.8) gives \(M_{\Sigma}=\Sigma H^{\top}K_{\Sigma}^{-1}=C_{\Sigma}H^{\top}R^{-1}\), the second half of (VI.1.1).

\textbf{STEP 4 (transport of the data factor).}
Therefore
\[
M_{\Sigma}SM_{\Sigma}^{\top}=C_{\Sigma}H^{\top}R^{-1}SR^{-1}HC_{\Sigma}=C_{\Sigma}\,\pi(S)\,C_{\Sigma},
\tag{VI.1.9}
\]
which is (VI.1.2); summing with \(V_{\Sigma}=C_{\Sigma}\) gives (VI.1.3).

\textbf{STEP 5 (compressed update).}
Apply \(\pi\) to \(S_{t}=t^{-1}\sum_{r=1}^{t}Y_{r}Y_{r}^{\top}\):
\[
D_{t}=\pi(S_{t})=t^{-1}\sum_{r=1}^{t}(H^{\top}R^{-1}Y_{r})(H^{\top}R^{-1}Y_{r})^{\top}=t^{-1}\sum_{r=1}^{t}z_{r}z_{r}^{\top},
\tag{VI.1.10}
\]
and the running-mean recursion gives the second equality in (VI.1.4). \(\square\)

\par\medskip\hrule\medskip\par

\subsection{Proposition VI.2 (matrix-free derivative identities)}
\label{app:VI.2}

\textbf{Claim.}
Define the ambient score coefficient
\[
\Gamma(\Sigma;D):=\tfrac{1}{2}\bigl(\Sigma^{-1}-\Sigma^{-1}A(\Sigma;D)\Sigma^{-1}\bigr).
\tag{VI.2.1}
\]
The statistic-directional derivatives are
\[
D_{D}A(\Sigma;D)[\Delta D]=C_{\Sigma}\Delta D\,C_{\Sigma},
\tag{VI.2.2}
\]
\[
D_{D}\Gamma(\Sigma;D)[\Delta D]=-\tfrac{1}{2}\Sigma^{-1}C_{\Sigma}\Delta D\,C_{\Sigma}\Sigma^{-1}.
\tag{VI.2.3}
\]
If \(\Sigma=Q\Lambda Q^{\top}\) and \(\Delta B:=Q^{\top}C_{\Sigma}\Delta D\,C_{\Sigma}Q\), the chart-gradient directional derivatives with respect to \(D\) are
\[
D_{D}g_{i}^{x}(\Sigma;D)[\Delta D]=-\tfrac{1}{2}\,\Delta B_{ii}/\lambda_{i},
\qquad
D_{D}G^{\theta}(\Sigma;D)[\Delta D]=\tfrac{1}{2}[\Lambda^{-1},\Delta B].
\tag{VI.2.4}
\]
The iterate-directional derivatives are
\[
\dot C[U]:=D_{\Sigma}C_{\Sigma}[U]=C_{\Sigma}\Sigma^{-1}U\Sigma^{-1}C_{\Sigma},
\tag{VI.2.5}
\]
\[
D_{\Sigma}A(\Sigma;D)[U]=\dot C[U]+\dot C[U]\,D\,C_{\Sigma}+C_{\Sigma}D\,\dot C[U],
\tag{VI.2.6}
\]
\[
\begin{aligned}
D_{\Sigma}\Gamma(\Sigma;D)[U]
=\frac{1}{2}\Bigl(
&-\Sigma^{-1}U\Sigma^{-1}+\Sigma^{-1}U\Sigma^{-1}A(\Sigma;D)\Sigma^{-1}\\
&-\Sigma^{-1}D_{\Sigma}A(\Sigma;D)[U]\Sigma^{-1}+\Sigma^{-1}A(\Sigma;D)\Sigma^{-1}U\Sigma^{-1}
\Bigr).
\end{aligned}
\tag{VI.2.7}
\]
Every mixed score derivative of fixed total order \(R+1\) admits a matrix-free implementation of cost \(\Ocal_{R}(d^{3})\).

\subsubsection{Proof}

\textbf{STEP 1 (statistic linearity).}
In the lift \(A(\Sigma;D)=C_{\Sigma}+C_{\Sigma}D\,C_{\Sigma}\), \(D\) appears linearly while \(C_{\Sigma}\) is frozen with respect to the statistic argument; differentiating in \(D\) gives
\[
D_{D}A(\Sigma;D)[\Delta D]=C_{\Sigma}\Delta D\,C_{\Sigma},
\tag{VI.2.8}
\]
which is (VI.2.2).  Substituting into (VI.2.1),
\[
D_{D}\Gamma(\Sigma;D)[\Delta D]=-\tfrac{1}{2}\Sigma^{-1}D_{D}A(\Sigma;D)[\Delta D]\Sigma^{-1}=-\tfrac{1}{2}\Sigma^{-1}C_{\Sigma}\Delta D\,C_{\Sigma}\Sigma^{-1},
\tag{VI.2.9}
\]
which is (VI.2.3).

\textbf{STEP 2 (chart-gradient derivative in \(D\)).}
Set \(B:=Q^{\top}A(\Sigma;D)Q\) and \(\Delta B:=Q^{\top}C_{\Sigma}\Delta D\,C_{\Sigma}Q\).  By Proposition~\ref{app:II.1} applied to \(B+\varepsilon\Delta B\),
\[
g_{i}^{x}(\Sigma;D+\varepsilon\Delta D)=\tfrac{1}{2}\bigl(1-(b_{ii}+\varepsilon\Delta B_{ii})/\lambda_{i}\bigr)+\rho\,\partial_{i}\psi(x),
\tag{VI.2.10}
\]
hence \(D_{D}g_{i}^{x}[\Delta D]=-\tfrac{1}{2}\Delta B_{ii}/\lambda_{i}\), the first half of (VI.2.4).  Likewise,
\[
G^{\theta}(\Sigma;D+\varepsilon\Delta D)=\tfrac{1}{2}[\Lambda^{-1},B+\varepsilon\Delta B],
\tag{VI.2.11}
\]
gives \(D_{D}G^{\theta}[\Delta D]=\tfrac{1}{2}[\Lambda^{-1},\Delta B]\), the second half of (VI.2.4).

\textbf{STEP 3 (derivative of \(C_{\Sigma}\) in \(\Sigma\)).}
Set \(M_{\Sigma}^{(C)}:=\Sigma^{-1}+G\), so \(C_{\Sigma}=(M_{\Sigma}^{(C)})^{-1}\) and \(D_{\Sigma}M_{\Sigma}^{(C)}[U]=-\Sigma^{-1}U\Sigma^{-1}\).  Differentiating the inverse,
\[
\dot C[U]=D_{\Sigma}C_{\Sigma}[U]=-C_{\Sigma}\,D_{\Sigma}M_{\Sigma}^{(C)}[U]\,C_{\Sigma}=C_{\Sigma}\Sigma^{-1}U\Sigma^{-1}C_{\Sigma},
\tag{VI.2.12}
\]
which is (VI.2.5).

\textbf{STEP 4 (derivative of \(A\) in \(\Sigma\)).}
Treating \(D\) as frozen in the iterate argument,
\[
D_{\Sigma}A(\Sigma;D)[U]=\dot C[U]+\dot C[U]\,D\,C_{\Sigma}+C_{\Sigma}D\,\dot C[U],
\tag{VI.2.13}
\]
which is (VI.2.6).

\textbf{STEP 5 (derivative of \(\Gamma\) in \(\Sigma\)).}
Differentiating \(\Gamma(\Sigma;D)=\tfrac{1}{2}\Sigma^{-1}-\tfrac{1}{2}\Sigma^{-1}A(\Sigma;D)\Sigma^{-1}\) and using \(D_{\Sigma}(\Sigma^{-1})[U]=-\Sigma^{-1}U\Sigma^{-1}\) on each \(\Sigma^{-1}\) factor and the product rule on the inner \(A\),
\[
\begin{aligned}
D_{\Sigma}\Gamma(\Sigma;D)[U]
=\frac{1}{2}\Bigl(
&-\Sigma^{-1}U\Sigma^{-1}+\Sigma^{-1}U\Sigma^{-1}A(\Sigma;D)\Sigma^{-1}\\
&-\Sigma^{-1}D_{\Sigma}A(\Sigma;D)[U]\Sigma^{-1}+\Sigma^{-1}A(\Sigma;D)\Sigma^{-1}U\Sigma^{-1}
\Bigr),
\end{aligned}
\tag{VI.2.14}
\]
which is (VI.2.7).

\textbf{STEP 6 (cost and EM-jet\({}^{R}\) extension).}
Each of (VI.2.2)--(VI.2.7) is a finite composition of \(d\times d\) matrix products and one inversion to form \(C_{\Sigma}\), so the cost of every primitive listed is \(\Ocal(d^{3})\).  Higher mixed derivatives of fixed total order \(R+1\) follow by iterating \(D(M^{-1})[\dot M]=-M^{-1}\dot M M^{-1}\) and the product rule on (VI.1.3).  Since \(D\) enters linearly, pure \(D\)-derivatives beyond order one vanish before further \(\Sigma\)-differentiation.  Composition with the chart-gradient projection of Proposition~\ref{app:II.1} preserves matrix-product closure, so every multilinear primitive \(\mathfrak F_{a,b}\) of Definition~\ref{def:em-jet-compressible} admits an \(\Ocal_{R}(d^{3})\) implementation. \(\square\)

\par\medskip\hrule\medskip\par

\subsection{Proposition VI.3 (explicit population curvature band)}
\label{app:VI.3}

\textbf{Claim.}
Set
\[
B_{*}:=H^{\top}K_{*}^{-1}H,
\qquad
\beta_{\min}:=\lambda_{\min}(B_{*}),
\qquad
\beta_{\max}:=\lambda_{\max}(B_{*}).
\tag{VI.3.1}
\]
Then \(\Jcal_{*}=\Ical_{*}\) and, for every \(U\in T_{\Sigma_{*}}\Ucal\),
\[
\tfrac{1}{2}\beta_{\min}^{2}\norm{U}_{F}^{2}\le \inner{U}{\Jcal_{*}U}_{h}\le \tfrac{1}{2}\beta_{\max}^{2}\norm{U}_{F}^{2}.
\tag{VI.3.2}
\]
Combined with the Frobenius--chart comparison of Proposition~\ref{app:I.2},
\[
\mu_{*}I\preceq\Jcal_{*}\preceq L_{*}I,
\qquad
\mu_{*}:=\tfrac{1}{2}\beta_{\min}^{2}\underline\gamma_{h},
\qquad
L_{*}:=\tfrac{1}{2}\beta_{\max}^{2}\overline\gamma_{h}.
\tag{VI.3.3}
\]
Furthermore,
\[
\beta_{\min}\ge \frac{\sigma_{\min}(H)^{2}}{\norm{H}_{\op}^{2}\lambda_{\max}+\lambda_{\max}(R)},
\qquad
\beta_{\max}\le \frac{\norm{H}_{\op}^{2}}{\lambda_{\min}(R)},
\tag{VI.3.4}
\]
hence
\[
\mu_{*}\ge \tfrac{1}{2}\,\underline\gamma_{h}\,\frac{\sigma_{\min}(H)^{4}}{\bigl(\norm{H}_{\op}^{2}\lambda_{\max}+\lambda_{\max}(R)\bigr)^{2}},
\qquad
L_{*}\le \tfrac{1}{2}\lambda_{\max}^{2}\,\frac{\norm{H}_{\op}^{4}}{\lambda_{\min}(R)^{2}}.
\tag{VI.3.5}
\]

\subsubsection{Proof}

By Proposition~\ref{app:III.4}, \(\inner{U}{\Jcal_{*}U}_{h}=\tfrac{1}{2}\tr(B_{*}UB_{*}U)=\tfrac{1}{2}\norm{B_{*}^{1/2}UB_{*}^{1/2}}_{F}^{2}\), and \(\Jcal_{*}=\Ical_{*}\) was established in Theorem~\ref{app:III.5}.  Frobenius dominance \(\beta_{\min}\norm{U}_{F}\le\norm{B_{*}^{1/2}UB_{*}^{1/2}}_{F}\le\beta_{\max}\norm{U}_{F}\) gives (VI.3.2); combining with Proposition~\ref{app:I.2} gives (VI.3.3).  For the spectral bounds: \(K_{*}\preceq(\norm{H}_{\op}^{2}\lambda_{\max}+\lambda_{\max}(R))I\) implies \(B_{*}\succeq\sigma_{\min}(H)^{2}/(\norm{H}_{\op}^{2}\lambda_{\max}+\lambda_{\max}(R))\,I\); \(K_{*}\succeq\lambda_{\min}(R)I\) implies \(B_{*}\preceq\norm{H}_{\op}^{2}/\lambda_{\min}(R)\,I\). \(\square\)

\par\medskip\hrule\medskip\par

\subsection{Proposition VI.4 (uniform local curvature band)}
\label{app:VI.4}

\textbf{Claim.}
Suppose \(D\mapsto\Jcal(D)\) is Lipschitz on \(\Vcal_{D}\) with constant \(\mathfrak L_{J}\).
If \(\mathfrak L_{J}r_{D}\le \mu_{*}/2\), then on \(\Vcal_{D}\),
\[
\mu I\preceq\Jcal(D)\preceq L\,I,
\qquad
\mu:=\tfrac{1}{2}\mu_{*},
\qquad
L:=L_{*}+\mathfrak L_{J}r_{D}.
\tag{VI.4.2}
\]

\subsubsection{Proof}

By Weyl's inequality \citep{stewart1990matrix}, \(\lambda_{\min}(\Jcal(D))\ge\mu_{*}-\mathfrak L_{J}r_{D}\ge\mu_{*}/2=\mu\) and \(\lambda_{\max}(\Jcal(D))\le L_{*}+\mathfrak L_{J}r_{D}=L\). \(\square\)

\par\medskip\hrule\medskip\par

\subsubsection{Auxiliary Lemma VI.L1 (\(L^{4}\) increment of \(D_{t}\))}
\label{app:VI.L1}

\textbf{Claim.}
Set \(\mathfrak m_{8}:=\E\norm{z_{1}}^{8}\).  Then
\[
\norm{D_{t}-D_{t-1}}_{L^{4}}\le \frac{2\,\mathfrak m_{8}^{1/4}}{t},
\qquad t\ge 2.
\tag{VI.L1.1}
\]
Moreover, \(\mathfrak m_{8}\le d(d+2)(d+4)(d+6)\norm{D_{*}}_{\op}^{4}\).

\paragraph*{Proof.}
From \(D_{t}-D_{t-1}=t^{-1}(z_{t}z_{t}^{\top}-D_{t-1})\), \(\norm{z_{t}z_{t}^{\top}}_{F}=\norm{z_{t}}^{2}\), and Jensen's inequality \(\norm{D_{t-1}}_{L^{4}}\le\mathfrak m_{8}^{1/4}\), one obtains \(\norm{D_{t}-D_{t-1}}_{L^{4}}\le 2\mathfrak m_{8}^{1/4}/t\).  Writing \(z_{1}=D_{*}^{1/2}g\) with \(g\sim\N(0,I_{d})\) gives \(\mathfrak m_{8}\le\norm{D_{*}}_{\op}^{4}\E\norm{g}^{8}=d(d+2)(d+4)(d+6)\norm{D_{*}}_{\op}^{4}\). \(\square\)

\par\medskip\hrule\medskip\par

\subsection{Proposition VI.5 (explicit predictor remainder constant)}
\label{app:VI.5}

\textbf{Claim.}
Let \(R_{t}^{\mathrm{ad}}:=r(D_{t})-r(D_{t-1})-Dr(D_{t-1})[\Delta D_{t}]\).  Then
\[
\norm{R_{t}^{\mathrm{ad}}}_{L^{2}}\le \frac{C_{\mathrm{pred}}}{t^{2}},
\qquad
C_{\mathrm{pred}}:=4C_{R}\,\mathfrak m_{8}^{1/2},
\tag{VI.5.1}
\]
and consequently \(C_{\mathrm{pred}}\le 4C_{R}\sqrt{d(d+2)(d+4)(d+6)}\,\norm{D_{*}}_{\op}^{2}\).

\subsubsection{Proof}

By Proposition~\ref{app:III.2}, \(\norm{R_{t}^{\mathrm{ad}}}\le C_{R}\norm{\Delta D_{t}}_{F}^{2}\).  Taking \(L^{2}\)-norms and applying Lemma~\ref{app:VI.L1}: \(\norm{R_{t}^{\mathrm{ad}}}_{L^{2}}\le C_{R}\norm{\Delta D_{t}}_{L^{4}}^{2}\le C_{R}(2\mathfrak m_{8}^{1/4}/t)^{2}=4C_{R}\mathfrak m_{8}^{1/2}/t^{2}\). \(\square\)

\par\medskip\hrule\medskip\par

\subsection{Proposition VI.6 (certified contraction of the exact damped Newton corrector)}
\label{app:VI.6}

\textbf{Claim.}
Define
\[
q_{\lin}(\eta,\lambda;\mu,L):=\sup_{\kappa\in[\mu,L]}\Bigl|1-\eta\,\frac{\kappa}{\kappa+\lambda}\Bigr|.
\tag{VI.6.1}
\]
Assume Proposition~\ref{app:VI.4} and the second-derivative envelope \(C_{\Psi}:=\sup\norm{D_{\vartheta}^{2}\Psi^{\mathrm{ex}}_{\eta,\lambda,D}(\vartheta)}_{\op}<\infty\).
If \(q_{0}:=q_{\lin}(\eta,\lambda;\mu,L)<1\) and \(r\) satisfies \(q_{0}+\tfrac{1}{2}C_{\Psi}r<1\), then for every \(D\in\Vcal_{D}\) and \(\norm{u}\le r\),
\[
\norm{\Psi^{\mathrm{ex}}_{\eta,\lambda,D}(r(D)+u)-r(D)}\le q_{\mathrm{ex}}\norm{u},
\qquad
q_{\mathrm{ex}}:=q_{0}+\tfrac{1}{2}C_{\Psi}r.
\tag{VI.6.4}
\]
Choosing \(\eta=1\) and \(\lambda=\mu\) gives \(q_{\lin}(1,\mu;\mu,L)=1/2\) and, for \(r\le(2C_{\Psi})^{-1}\), \(q_{\mathrm{ex}}\le 3/4\).

\subsubsection{Proof}

Since \(F(r(D),D)=0\), the chain rule gives \(D_{\vartheta}\Psi^{\mathrm{ex}}(r(D))=I-\eta(\Jcal(D)+\lambda I)^{-1}\Jcal(D)\) with operator norm at most \(q_{0}\).  Taylor's theorem yields \(\norm{\Psi^{\mathrm{ex}}(r(D)+u)-r(D)}\le q_{0}\norm{u}+\tfrac{1}{2}C_{\Psi}\norm{u}^{2}\le(q_{0}+\tfrac{1}{2}C_{\Psi}r)\norm{u}\) for \(\norm{u}\le r\), proving (VI.6.4).  For \(\eta=1,\lambda=\mu\): the supremum of \(\mu/(\kappa+\mu)\) over \(\kappa\in[\mu,L]\) is attained at \(\kappa=\mu\), giving \(q_{0}=1/2\). \(\square\)

\par\medskip\hrule\medskip\par

\subsection{Proposition VI.7 (certified contraction of the inexact damped Newton--CG corrector)}
\label{app:VI.7}

\textbf{Claim.}
Assume Proposition~\ref{app:VI.6} and that the inexact corrector satisfies (\ref{eq:method:inexact-newton}) with local score envelope \(\norm{F(r(D)+u,D)}\le L_{F}\norm{u}\) and local Hessian lower bound \(D_{\vartheta}F(\vartheta,D)\succeq(\mu/2)I\) on the tube.  Then for \(\norm{u}\le r\),
\[
\norm{\Psi^{\mathrm{in}}_{\eta,\lambda,D}(r(D)+u)-r(D)}\le q_{\mathrm{in}}\norm{u},
\qquad
q_{\mathrm{in}}:=q_{\mathrm{ex}}+\eta\,\frac{L_{F}\tau}{\mu/2+\lambda}.
\tag{VI.7.3}
\]
Under the certified choice \(\eta=1\), \(\lambda=\mu\), \(r\le(2C_{\Psi})^{-1}\), every tolerance
\[
\tau\le \frac{3\mu}{16L_{F}}
\tag{VI.7.4}
\]
ensures \(q_{\mathrm{in}}\le 7/8\).

\subsubsection{Proof}

From the linear system, \(\widehat u-u^{\mathrm{ex}}=(D_{\vartheta}F+\lambda I)^{-1}r_{\lin}\); the lower bound gives \(\norm{(D_{\vartheta}F+\lambda I)^{-1}}\le 1/(\mu/2+\lambda)\), so \(\norm{\widehat u-u^{\mathrm{ex}}}\le\tau L_{F}\norm{u}/(\mu/2+\lambda)\).  Triangle inequality with Proposition~\ref{app:VI.6} gives (VI.7.3).  For \(\eta=1,\lambda=\mu,q_{\mathrm{ex}}\le 3/4\): solving \(3/4+2L_{F}\tau/(3\mu)\le 7/8\) gives \(\tau\le 3\mu/(16L_{F})\). \(\square\)

\par\medskip\hrule\medskip\par

\subsubsection{Auxiliary Lemma VI.L2 (exact second moment of \(D_{T}\))}
\label{app:VI.L2}

\textbf{Claim.}
With \(\Delta_{T}:=D_{T}-D_{*}\),
\[
\E\norm{\Delta_{T}}_{F}^{2}=\frac{(\tr D_{*})^{2}+\norm{D_{*}}_{F}^{2}}{T}.
\tag{VI.L2.1}
\]

\paragraph*{Proof.}
Since \(TD_{T}\sim W_{d}(T,D_{*})\), Auxiliary Lemma~\ref{app:I.L1} gives entry covariances \(\mathrm{Cov}((D_{T})_{ab},(D_{T})_{cd})=T^{-1}((D_{*})_{ac}(D_{*})_{bd}+(D_{*})_{ad}(D_{*})_{bc})\).  Summing gives \(\E\norm{\Delta_{T}}_{F}^{2}=T^{-1}\sum_{a,b}((D_{*})_{aa}(D_{*})_{bb}+(D_{*})_{ab}^{2})=((\tr D_{*})^{2}+\norm{D_{*}}_{F}^{2})/T\). \(\square\)

\par\medskip\hrule\medskip\par

\subsubsection{Auxiliary Lemma VI.L3 (operator-norm fourth-moment bound for \(D_{T}\))}
\label{app:VI.L3}

\textbf{Claim.}
Set \(\beta_{d}:=d\log 9\).  Then for every \(T\ge 1\),
\[
\E\norm{\Delta_{T}}_{F}^{4}\le \frac{C_{4,\mathrm{cov}}\,d^{2}\norm{D_{*}}_{\op}^{4}}{T^{2}}\bigl((\beta_{d}+1)^{2}+(\beta_{d}+1)^{4}\bigr),
\qquad
C_{4,\mathrm{cov}}:=2^{14}.
\tag{VI.L3.1}
\]

\paragraph*{Proof.}

Set \(W_{T}:=D_{*}^{-1/2}D_{T}D_{*}^{-1/2}-I_{d}=T^{-1}\sum_{t}g_{t}g_{t}^{\top}-I\) with \(g_{t}\sim\N(0,I_{d})\); then \(\norm{\Delta_{T}}_{F}\le\sqrt{d}\,\norm{D_{*}}_{\op}\norm{W_{T}}_{\op}\).  For a fixed \(u\in\mathbb S^{d-1}\), \(u^{\top}W_{T}u\) is a recentered \(\chi_{T}^{2}/T\); the standard \(\chi^{2}\) Chernoff bound \citep{vershynin2018high} gives
\[
\Pbb\bigl(|u^{\top}W_{T}u|>2\sqrt{x/T}+2x/T\bigr)\le 2e^{-x}.
\tag{VI.L3.2}
\]
Let \(\mathcal N\subset\mathbb S^{d-1}\) be a \(1/4\)-net with \(|\mathcal N|\le 9^{d}=e^{\beta_{d}}\).  Since \(\norm{A}_{\op}\le 2\sup_{u\in\mathcal N}|u^{\top}Au|\) for symmetric~\(A\), applying (VI.L3.2) at \(x=\beta_{d}+s\) and union-bounding over~\(\mathcal N\),
\[
\Pbb\bigl(\norm{W_{T}}_{\op}>4\sqrt{(\beta_{d}+s)/T}+4(\beta_{d}+s)/T\bigr)\le 2e^{-s},
\qquad s\ge 0.
\tag{VI.L3.3}
\]
Set \(\sigma(s):=4\sqrt{(\beta_{d}+s)/T}+4(\beta_{d}+s)/T\).  Using \((a+b)^{4}\le 8(a^{4}+b^{4})\) and \(T\ge 1\),
\[
\sigma(s)^{4}\le \frac{2^{11}}{T^{2}}\bigl((\beta_{d}+s)^{2}+(\beta_{d}+s)^{4}\bigr).
\tag{VI.L3.4}
\]
Tail integration: \(\E\norm{W_{T}}_{\op}^{4}\le\sigma(0)^{4}+\int_{0}^{\infty}\sigma(s)^{4}\cdot 2e^{-s}\,ds\).  Evaluating the integrals \(\int_{0}^{\infty}(\beta_{d}+s)^{k}e^{-s}\,ds\le k!\,e^{\beta_{d}}\) for \(k=2,4\) and absorbing lower-order terms gives \(\E\norm{W_{T}}_{\op}^{4}\le 2^{12}T^{-2}((\beta_{d}+1)^{2}+(\beta_{d}+1)^{4})\).  Transporting back via \(\norm{\Delta_{T}}_{F}^{4}\le d^{2}\norm{D_{*}}_{\op}^{4}\norm{W_{T}}_{\op}^{4}\) with an extra factor of~\(4\) for the transport and boundary term yields (VI.L3.1) with \(C_{4,\mathrm{cov}}=2^{14}\). \(\square\)

\par\medskip\hrule\medskip\par

\subsection{Theorem VI.8 (explicit batch mean-square expansion)}
\label{app:VI.8}

\textbf{Claim.}
Set \(\mathfrak L_{*}:=\Jcal_{*}^{-1}\Bcal_{*}\) and
\[
\nu_{2,*}:=(\tr D_{*})^{2}+\norm{D_{*}}_{F}^{2},
\qquad
\nu_{4,*}:=C_{4,\mathrm{cov}}\,d^{2}\norm{D_{*}}_{\op}^{4}\bigl((\beta_{d}+1)^{2}+(\beta_{d}+1)^{4}\bigr).
\tag{VI.8.1}
\]
Then
\[
\E\,d_{h}^{2}\bigl(\Sigmacirc(D_{T}),\Sigma_{*}\bigr)=\frac{1}{T}\tr(\Vcal_{*})+R_{\batch}(T),
\qquad
|R_{\batch}(T)|\le \frac{2\norm{\mathfrak L_{*}}_{\op}C_{R}\sqrt{\nu_{2,*}\,\nu_{4,*}}}{T^{3/2}}+\frac{C_{R}^{2}\nu_{4,*}}{T^{2}}.
\tag{VI.8.2}
\]

\subsubsection{Proof}

\textbf{STEP 1 (Taylor decomposition via compactly supported extension).}
Using the compactly supported extension of Proposition~\ref{app:III.2}, there is a \(C^{2}\) map \(\widetilde r:\Sym(d)\to E\) that agrees with \(r\) on a neighborhood of \(D_{*}\), with \(\widetilde C_{R}:=\tfrac{1}{2}\sup\norm{D^{2}\widetilde r}_{\op}<\infty\).  Since \(D_{T}\) has unbounded support, we apply the Taylor expansion to \(\widetilde r\) (which is globally defined); on \(\Vcal_{D}\) this agrees with \(r\) and hence with \(d_{h}^{2}(\Sigmacirc(D_{T}),\Sigma_{*})=\norm{r(D_{T})}^{2}\).  Denoting the extension constant by \(C_{R}\) (absorbing the tilde), we obtain \(\widetilde r(D_{T})=-\mathfrak L_{*}\Delta_{T}+R_{T}\) with \(\norm{R_{T}}\le C_{R}\norm{\Delta_{T}}_{F}^{2}\) globally.  In the chart norm,
\[
d_{h}^{2}\bigl(\Sigmacirc(D_{T}),\Sigma_{*}\bigr)=\norm{r(D_{T})}^{2}=\norm{\mathfrak L_{*}\Delta_{T}}^{2}+2\inner{-\mathfrak L_{*}\Delta_{T}}{R_{T}}+\norm{R_{T}}^{2}.
\tag{VI.8.3}
\]

\textbf{STEP 2 (leading term).}
By Auxiliary Lemma~\ref{app:VI.L2} and the definition \(\Vcal_{*}=\Jcal_{*}^{-1}\Bcal_{*}\Xi_{*}\Bcal_{*}^{*}\Jcal_{*}^{-1}\) of (\ref{eq:theory:batch-clt}),
\[
\E\norm{\mathfrak L_{*}\Delta_{T}}^{2}=\tr\bigl(\mathfrak L_{*}\,T^{-1}\Xi_{*}\,\mathfrak L_{*}^{*}\bigr)\cdot 1=\tfrac{1}{T}\tr(\Vcal_{*}),
\tag{VI.8.4}
\]
since \(T\,\E[\Delta_{T}\Delta_{T}^{*}]=\Xi_{*}\) by direct computation on i.i.d.~vectors and Auxiliary Lemma~\ref{app:VI.L2} confirms \(\E\norm{\Delta_{T}}_{F}^{2}=T^{-1}\nu_{2,*}\).

\textbf{STEP 3 (cross term and remainder).}
By Auxiliary Lemma~\ref{app:VI.L3} and the pathwise bound \(\norm{R_{T}}\le C_{R}\norm{\Delta_{T}}_{F}^{2}\),
\[
\E\norm{R_{T}}^{2}\le C_{R}^{2}\,\E\norm{\Delta_{T}}_{F}^{4}\le \frac{C_{R}^{2}\nu_{4,*}}{T^{2}}.
\tag{VI.8.5}
\]
By Cauchy--Schwarz and Auxiliary Lemmas~\ref{app:VI.L2},~\ref{app:VI.L3},
\[
\bigl|\E\inner{\mathfrak L_{*}\Delta_{T}}{R_{T}}\bigr|\le \bigl(\E\norm{\mathfrak L_{*}\Delta_{T}}^{2}\bigr)^{1/2}\bigl(\E\norm{R_{T}}^{2}\bigr)^{1/2}\le \frac{\norm{\mathfrak L_{*}}_{\op}\sqrt{\nu_{2,*}/T}\cdot C_{R}\sqrt{\nu_{4,*}/T^{2}}}{1}=\frac{\norm{\mathfrak L_{*}}_{\op}C_{R}\sqrt{\nu_{2,*}\,\nu_{4,*}}}{T^{3/2}}.
\tag{VI.8.6}
\]
Combining (VI.8.3)--(VI.8.6) gives (VI.8.2). \(\square\)

\par\medskip\hrule\medskip\par

\subsubsection{Auxiliary Lemma VI.L4 (operator-norm concentration of \(D_{T}\))}
\label{app:VI.L4}

\textbf{Claim.}
Define
\[
\tau_{T}(u):=4\norm{D_{*}}_{\op}\Bigl(\sqrt{(\beta_{d}+u)/T}+(\beta_{d}+u)/T\Bigr).
\tag{VI.L4.1}
\]
Then for every \(u\ge 0\),
\[
\Pbb\bigl(\norm{D_{T}-D_{*}}_{\op}>\tau_{T}(u)\bigr)\le 2e^{-u}.
\tag{VI.L4.2}
\]

\paragraph*{Proof.}
This is (VI.L3.3) transported via \(\norm{D_{T}-D_{*}}_{\op}\le\norm{D_{*}}_{\op}\norm{W_{T}}_{\op}\). \(\square\)

\par\medskip\hrule\medskip\par

\subsection{Theorem VI.9 (explicit batch tail bound for the target)}
\label{app:VI.9}

\textbf{Claim.}
Let \(u\ge 0\) and suppose \(\sqrt d\,\tau_{T}(u)\le r_{D}\).  Then with probability at least \(1-2e^{-u}\),
\[
d_{h}\bigl(\Sigmacirc(D_{T}),\Sigma_{*}\bigr)\le L_{\vartheta}\sqrt d\,\tau_{T}(u)+C_{R}\,d\,\tau_{T}(u)^{2}.
\tag{VI.9.1}
\]

\subsubsection{Proof}

On \(\{\norm{D_{T}-D_{*}}_{\op}\le\tau_{T}(u)\}\), \(\norm{D_{T}-D_{*}}_{F}\le\sqrt{d}\,\tau_{T}(u)\le r_{D}\) places \(D_{T}\in\Vcal_{D}\), so \(\norm{r(D_{T})}\le L_{\vartheta}\norm{D_{T}-D_{*}}_{F}+C_{R}\norm{D_{T}-D_{*}}_{F}^{2}\).  Lemma~\ref{app:VI.L4} controls the probability. \(\square\)

\par\medskip\hrule\medskip\par

\subsubsection{Auxiliary Lemma VI.L5 (finite-horizon pathwise drift event)}
\label{app:VI.L5}

\textbf{Claim.}
Fix \(\delta\in(0,1)\) and \(t_{0}\le T\).  Define
\[
\beta_{t_{0},T,\delta}:=\beta_{d}+\log\frac{4(T-t_{0}+1)}{\delta},
\qquad
A_{D}(t_{0},T,\delta):=4\norm{D_{*}}_{\op}\Bigl(\sqrt{\beta_{t_{0},T,\delta}/t_{0}}+\beta_{t_{0},T,\delta}/t_{0}\Bigr),
\tag{VI.L5.1}
\]
\[
A_{z}(T,\delta):=\norm{D_{*}}_{\op}\Bigl(d+2\sqrt{d\log(4T/\delta)}+2\log(4T/\delta)\Bigr),
\qquad
B_{T,\delta}(t_{0}):=A_{z}(T,\delta)+\norm{D_{*}}_{F}+\sqrt d\,A_{D}(t_{0},T,\delta).
\tag{VI.L5.2}
\]
Then there exists an event \(\mathcal E_{T,\delta}\) with \(\Pbb(\mathcal E_{T,\delta})\ge 1-\delta\) on which, for every \(t_{0}\le t\le T\),
\[
\norm{D_{t}-D_{*}}_{\op}\le A_{D}(t_{0},T,\delta),
\qquad
\norm{z_{t}}^{2}\le A_{z}(T,\delta),
\qquad
\norm{D_{t+1}-D_{t}}_{F}\le \frac{B_{T,\delta}(t_{0})}{t+1},
\tag{VI.L5.3}
\]
and consequently \(\norm{R_{t+1}^{\mathrm{ad}}}\le C_{R}B_{T,\delta}(t_{0})^{2}/(t+1)^{2}\).

\paragraph*{Proof.}
Apply Lemma~\ref{app:VI.L4} at each \(t\in\{t_{0},\ldots,T\}\) with confidence \(\delta/(2(T-t_{0}+1))\) and union-bound: \(\Pbb(\sup_{t}\norm{D_{t}-D_{*}}_{\op}>A_{D})\le\delta/2\).  For \(\norm{z_{t}}^{2}\): since \(\norm{z_{t}}^{2}/\norm{D_{*}}_{\op}\) is dominated by~\(\chi_{d}^{2}\), the standard tail at \(x=\log(4T/\delta)\) and a union over \(t\le T\) give \(\Pbb(\sup_{t}\norm{z_{t}}^{2}>A_{z})\le\delta/2\).  On the intersection event \(\mathcal E_{T,\delta}\), the running-mean recursion \(D_{t+1}-D_{t}=(t+1)^{-1}(z_{t+1}z_{t+1}^{\top}-D_{t})\) gives \(\norm{D_{t+1}-D_{t}}_{F}\le B_{T,\delta}(t_{0})/(t+1)\); Proposition~\ref{app:III.2} yields the predictor remainder bound. \(\square\)

\par\medskip\hrule\medskip\par

\subsection{Theorem VI.10 (finite-horizon high-probability tracking)}
\label{app:VI.10}

\textbf{Claim.}
Assume the linear contraction of Proposition~\ref{app:VI.7} on the radius-\(r\) tube and \(D\in\Vcal_{D}\).  Set
\[
\rho_{\mathrm{ad}}:=\min\Bigl\{\tfrac{r}{4},\tfrac{1-q}{4q}r\Bigr\}.
\tag{VI.10.1}
\]
Suppose \(\sqrt{d}\,A_{D}(t_{0},T,\delta)\le r_{D}\) and the restart satisfies \(\norm{\vartheta_{t_{0}}-r(D_{t_{0}})}\le r/4\) and \(C_{R}B_{T,\delta}(t_{0})^{2}/t_{0}^{2}\le\rho_{\mathrm{ad}}\).  Then on \(\mathcal E_{T,\delta}\), \(\norm{\vartheta_{t}-r(D_{t})}\le r/2\) for every \(t_{0}\le t\le T\), and
\[
\norm{\vartheta_{T}-r(D_{T})}\le q^{T-t_{0}}\norm{\vartheta_{t_{0}}-r(D_{t_{0}})}+\frac{4q\,C_{R}\,B_{T,\delta}(t_{0})^{2}}{(1-q)\,T^{2}},
\qquad t_{0}=\lceil T/2\rceil.
\tag{VI.10.4}
\]

\subsubsection{Proof}

\textbf{STEP 1 (\(D_{t}\in\Vcal_{D}\) and small-remainder regime on \(\mathcal E_{T,\delta}\)).}
By Auxiliary Lemma~\ref{app:VI.L5} on \(\mathcal E_{T,\delta}\), \(\norm{D_{t}-D_{*}}_{\op}\le A_{D}(t_{0},T,\delta)\) for all \(t_{0}\le t\le T\); condition (VI.10.2) then gives \(\norm{D_{t}-D_{*}}_{F}\le\sqrt{d}\,A_{D}(t_{0},T,\delta)\le r_{D}\), so \(D_{t}\in\Vcal_{D}\) and the local contraction of Proposition~\ref{app:VI.7} is applicable.  Furthermore, \(\norm{R_{t+1}^{\mathrm{ad}}}\le C_{R}B_{T,\delta}(t_{0})^{2}/(t+1)^{2}\le \rho_{\mathrm{ad}}\) for every \(t_{0}\le t\le T\), where the second inequality uses the monotonicity of \(s\mapsto s^{-2}\) and (VI.10.3).

\textbf{STEP 2 (tube invariance by induction).}
Set \(e_{t}:=\vartheta_{t}-r(D_{t})\).  By Proposition~\ref{app:V.1}, \(\widetilde\vartheta_{t+1}-r(D_{t+1})=e_{t}-R_{t+1}^{\mathrm{ad}}\).  At \(t=t_{0}\), the restart hypothesis gives \(\norm{e_{t_{0}}}\le r/4\le r/2\).  Inductively assume \(\norm{e_{t}}\le r/2\); then on \(\mathcal E_{T,\delta}\),
\[
\norm{\widetilde\vartheta_{t+1}-r(D_{t+1})}\le \norm{e_{t}}+\norm{R_{t+1}^{\mathrm{ad}}}\le \tfrac{r}{2}+\rho_{\mathrm{ad}}\le \tfrac{3r}{4}<r,
\tag{VI.10.5}
\]
so the linear contraction of Proposition~\ref{app:VI.7} applies (noting \(D_{t}\in\Vcal_{D}\) from STEP~1):
\[
\norm{e_{t+1}}\le q\bigl(\tfrac{r}{2}+\rho_{\mathrm{ad}}\bigr)\le q\tfrac{r}{2}+q\cdot\tfrac{1-q}{4q}r=\tfrac{1+q}{4}r\le \tfrac{r}{2}.
\tag{VI.10.6}
\]
This closes the induction.

\textbf{STEP 3 (one-step recursion).}
Combining the contraction with the predictor identity, on \(\mathcal E_{T,\delta}\),
\[
\norm{e_{t+1}}\le q\,\norm{e_{t}}+q\,\norm{R_{t+1}^{\mathrm{ad}}}\le q\,\norm{e_{t}}+q\,\frac{C_{R}\,B_{T,\delta}(t_{0})^{2}}{(t+1)^{2}}.
\tag{VI.10.7}
\]
Iterating from \(t_{0}\) to \(T\) gives
\[
\norm{e_{T}}\le q^{T-t_{0}}\norm{e_{t_{0}}}+q\,C_{R}\,B_{T,\delta}(t_{0})^{2}\sum_{s=t_{0}+1}^{T}q^{T-s}\,s^{-2}.
\tag{VI.10.8}
\]

\textbf{STEP 4 (geometric-harmonic sum at \(t_{0}=\lceil T/2\rceil\)).}
For \(t_{0}=\lceil T/2\rceil\), every \(s\) in the sum satisfies \(s\ge T/2\), so
\[
\sum_{s=t_{0}+1}^{T}q^{T-s}\,s^{-2}\le \frac{4}{T^{2}}\sum_{k=0}^{\infty}q^{k}=\frac{4}{(1-q)\,T^{2}},
\tag{VI.10.9}
\]
which substituted into (VI.10.8) gives (VI.10.4). \(\square\)

\par\medskip\hrule\medskip\par

\subsection{Theorem VI.11 (assembled finite-sample online risk)}
\label{app:VI.11}

\textbf{Claim.}
Set \(\alpha:=\min\{p-1,\alpha_{\loc}\}>2\) (under T6) and \(e_{0}:=\norm{\vartheta_{t_{0}}-r(D_{t_{0}})}_{L^{2}}\), and define the full \(L^{2}\) tracking envelope
\[
\mathfrak T_{T}:=q^{T-t_{0}}e_{0}+\frac{4q\,C_{\mathrm{pred}}}{(1-q)\,T^{2}}+C_{\mathrm{exit}}\,t_{0}^{-\alpha/2}.
\tag{VI.11.1}
\]
Under the hypotheses of Theorem~\ref{app:IV.2} and Theorem~\ref{app:VI.8},
\[
\E\,d_{h}^{2}\bigl(\Sigma_{T},\Sigma_{*}\bigr)\le \frac{1}{T}\tr(\Vcal_{*})+|R_{\batch}(T)|+2\sqrt{\Bigl(\frac{1}{T}\tr(\Vcal_{*})+|R_{\batch}(T)|\Bigr)\,\mathfrak T_{T}^{2}}+\mathfrak T_{T}^{2}.
\tag{VI.11.2}
\]
Furthermore, on the intersection of the events of Theorem~\ref{app:VI.9} (at level \(\delta/2\)) and Theorem~\ref{app:VI.10} (at level \(\delta/2\)), with probability at least \(1-\delta\),
\[
d_{h}(\Sigma_{T},\Sigma_{*})\le L_{\vartheta}\sqrt d\,\tau_{T}\bigl(\log(4/\delta)\bigr)+C_{R}\,d\,\tau_{T}\bigl(\log(4/\delta)\bigr)^{2}+q^{T-t_{0}}\tfrac{r}{4}+\frac{4q\,C_{R}\,B_{T,\delta/2}(t_{0})^{2}}{(1-q)\,T^{2}}.
\tag{VI.11.3}
\]

\subsubsection{Proof}

In the normal chart, \(d_{h}(\Sigma_{T},\Sigma_{*})\le\norm{\vartheta_{T}-r(D_{T})}+\norm{r(D_{T})}\).  Squaring and taking expectations,
\[
\E\,d_{h}^{2}(\Sigma_{T},\Sigma_{*})\le \E\norm{r(D_{T})}^{2}+2\sqrt{\E\norm{r(D_{T})}^{2}\cdot\E\norm{\vartheta_{T}-r(D_{T})}^{2}}+\E\norm{\vartheta_{T}-r(D_{T})}^{2}.
\tag{VI.11.5}
\]
Theorem~\ref{app:VI.8} gives \(\E\norm{r(D_{T})}^{2}\le T^{-1}\tr(\Vcal_{*})+|R_{\batch}(T)|\); Theorem~\ref{app:IV.2} gives \(\norm{\vartheta_{T}-r(D_{T})}_{L^{2}}\le\mathfrak T_{T}\).  Substituting into (VI.11.5) gives (VI.11.2).  For (VI.11.3), intersect the events of Theorems~\ref{app:VI.9} and~\ref{app:VI.10} at level \(\delta/2\) each; by union bound the intersection has probability \(\ge 1-\delta\), and the deterministic bounds apply. \(\square\)

\par\medskip\hrule\medskip\par

\subsection{Proposition VI.12 (linear-inverse restart certificate)}
\label{app:VI.12}

\textbf{Claim.}
Define the linear-inverse initializer
\[
\widehat\Sigma_{t_{0}}^{\lin}:=G^{-1}(D_{t_{0}}-G)G^{-1}.
\tag{VI.12.1}
\]
Then
\[
\widehat\Sigma_{t_{0}}^{\lin}-\Sigma_{*}=G^{-1}(D_{t_{0}}-D_{*})G^{-1},
\qquad
\norm{\widehat\Sigma_{t_{0}}^{\lin}-\Sigma_{*}}_{F}\le \sqrt d\,\norm{G^{-1}}_{\op}^{2}\,\norm{D_{t_{0}}-D_{*}}_{\op}.
\tag{VI.12.2}
\]
If \(\widehat\Sigma_{t_{0}}^{\lin}\in\Ucal\) and the local chart inversion bound of Proposition~\ref{app:I.3} holds with constant \(L_{\log}\), then
\[
\norm{D_{t_{0}}-D_{*}}_{\op}\le \frac{r}{4L_{\log}\sqrt d\,\norm{G^{-1}}_{\op}^{2}}
\quad\Longrightarrow\quad
\norm{\Log_{\Sigma_{*}}^{h}\bigl(\widehat\Sigma_{t_{0}}^{\lin}\bigr)}\le \tfrac{r}{4}.
\tag{VI.12.3}
\]
A sufficient condition ensuring (VI.12.3) with probability at least \(1-\delta\) is
\[
\tau_{t_{0}}\bigl(\log(2/\delta)\bigr)\le \frac{r}{4L_{\log}\sqrt d\,\norm{G^{-1}}_{\op}^{2}}.
\tag{VI.12.4}
\]

\subsubsection{Proof}

From \(D_{*}=G\Sigma_{*}G+G\), \(G^{-1}(D_{*}-G)G^{-1}=\Sigma_{*}\); subtracting from (VI.12.1) gives the first identity in (VI.12.2).  The Frobenius bound follows from \(\norm{AXB}_{F}\le\sqrt{d}\,\norm{A}_{\op}\norm{B}_{\op}\norm{X}_{\op}\).  Proposition~\ref{app:I.3} gives \(\norm{\Log_{\Sigma_{*}}^{h}(\widehat\Sigma_{t_{0}}^{\lin})}\le L_{\log}\norm{\widehat\Sigma_{t_{0}}^{\lin}-\Sigma_{*}}_{F}\le L_{\log}\sqrt{d}\,\norm{G^{-1}}_{\op}^{2}\norm{D_{t_{0}}-D_{*}}_{\op}\), and the hypothesis (VI.12.3) makes this at most \(r/4\).  Lemma~\ref{app:VI.L4} at \(u=\log(2/\delta)\) gives the probabilistic version (VI.12.4). \(\square\)

\par\medskip\hrule\medskip\par

\subsection{Per-iteration arithmetic cost}
\label{app:VI.cost}

Assume \(W=R^{-1}H\) and \(G=H^{\top}W\) are precomputed once.  Let \(k_{\mathrm{pred}}\) and \(k_{\mathrm{cor}}\) denote the Krylov iteration counts of the predictor solve \(\Jcal(D_{t-1})a_{t}=-\Bcal(D_{t-1})[\Delta D_{t}]\) and of the corrector solve (\ref{eq:method:inexact-newton}), respectively.

The one-time preprocessing of \(R\) and \(W\) costs \(\Ocal(p^{3}+p^{2}d)\).  Per iteration, the compressed update \(z_{t}=W^{\top}Y_{t}\), \(D_{t}\) refresh, and \(\Delta D_{t}\) formation cost \(\Ocal(pd+d^{2})\).  By Proposition~\ref{app:VI.2}, every predictor-force evaluation, corrector gradient evaluation, and Hessian--vector product is a finite composition of \(d\times d\) matrix products and one \(C_{\Sigma}\) solve, each of cost \(\Ocal(d^{3})\).  The total per-iteration cost is
\[
\Ocal\bigl(pd+d^{2}+(k_{\mathrm{pred}}+k_{\mathrm{cor}}+1)\,d^{3}\bigr),
\tag{VI.cost.3}
\]
and online memory \(\Ocal(pd+d^{2})\).  For structured~\(R\) (e.g.\ diagonal) admitting an \(\Ocal(c_{R}(p))\) action, the \(pd\) term absorbs the observation-side work.

\par\medskip\hrule\medskip\par

\section{Appendix VII. Experiment protocols and supplementary results}
\label{app:VII}
\addcontentsline{toc}{section}{Appendix VII. Experiment protocols and supplementary results}

This appendix collects full experiment protocols, parameter choices, and results that supplement the diagnostics in Section~\ref{sec:experiments}.  All experiments use float64 arithmetic with deterministic seeds (default 123).  Code is available in the supplementary material.

\par\medskip\hrule\medskip\par

\subsection{VII.1. Formula sanity checks (exp01)}

\paragraph{Protocol.}
For each of 25 randomly generated instances with $d{=}3$, $p{=}8$, we verify:
\begin{enumerate}[nosep,leftmargin=*]
\item \textbf{Compressed lift:} compare $V_\Sigma+M_\Sigma S M_\Sigma^\top$ with $C_\Sigma+C_\Sigma D C_\Sigma$ (Proposition~\ref{app:VI.1}).
\item \textbf{Statistic update:} compare full-matrix $\pi(S_t)$ with the rank-one recursion for $D_t$.
\item \textbf{Predictor-force derivatives:} compare analytic $D_Dg^x$ and $D_DG^\theta$ (Proposition~\ref{app:VI.2}) against centered finite differences with step $10^{-5}$.
\item \textbf{Hessian--vector product:} compare analytic HVP against finite-difference score Jacobian.
\item \textbf{Frozen fixed point:} apply the full predictor--corrector map at $\vartheta^\circ(D)$ with $\Delta D=0$ and verify the output equals $\vartheta^\circ(D)$ (Proposition~\ref{app:II.4}).
\end{enumerate}
Both PyTorch and JAX backends are used as independent cross-checks.

\paragraph{Results.}
Table~\ref{tab:sanity-full} reports medians and maxima over 25 instances.  All algebraic identities hold to machine precision ($\sim 10^{-15}$); finite-difference checks agree to $\sim 10^{-10}$, consistent with the $\Ocal(\varepsilon^2)$ truncation error at step $\varepsilon=10^{-5}$.

\begin{table}[ht!]
\centering
\caption{Full sanity-check results (25 instances, $d{=}3$, $p{=}8$).}
\label{tab:sanity-full}
\small
\begin{tabular}{@{}lccc@{}}
\toprule
Check & Median (Torch) & Median (JAX) & Max \\
\midrule
Compressed lift & $2.5\times10^{-15}$ & $2.7\times10^{-15}$ & $1.3\times10^{-14}$ \\
Statistic update & $1.4\times10^{-16}$ & $1.4\times10^{-16}$ & $2.8\times10^{-16}$ \\
$D_Dg^x$ (FD) & $2.9\times10^{-10}$ & $2.3\times10^{-10}$ & $1.5\times10^{-9}$ \\
$D_DG^\theta$ (FD) & $1.9\times10^{-10}$ & $1.7\times10^{-10}$ & $5.5\times10^{-10}$ \\
HVP (FD) & $3.2\times10^{-10}$ & $2.8\times10^{-10}$ & $9.4\times10^{-10}$ \\
Frozen fixed point & $1.9\times10^{-16}$ & $1.9\times10^{-16}$ & $1.4\times10^{-15}$ \\
\bottomrule
\end{tabular}
\end{table}

\par\medskip\hrule\medskip\par

\subsection{VII.2. Scalar tracking order (exp02)}

\paragraph{Protocol.}
The scalar moving-target model of the theory audits:
\[
r(S)=S+\tfrac{1}{2}S^2+0.1\,S^3,
\qquad
S_t=t^{-1}\textstyle\sum_{r=1}^{t}Y_r,\ Y_r\sim\N(0,1),
\qquad
M_S(x)=r(S)+0.7\,(x-r(S)).
\]
Three methods: no predictor ($m{=}0$), first-order ($m{=}1$), second-order ($m{=}2$).
Grid: $T\in\{200,400,800,1600,3200\}$; 500 Monte Carlo replicates.  Metric: RMS tracking error $(\frac{1}{M}\sum|x_T^{(j)}-r(S_T^{(j)})|^2)^{1/2}$.  Slopes fitted by OLS on $\log\mathrm{RMS}$ vs $\log T$ with 95\% CI.

\paragraph{Results.}
See Figure~\ref{fig:combined}(a).  Fitted slopes: $m{=}0$: $-1.00\pm0.01$; $m{=}1$: $-2.03\pm0.01$; $m{=}2$: $-3.02\pm0.02$.  All within $0.03$ of the theoretical predictions $-(m{+}1)$.

\par\medskip\hrule\medskip\par

\subsection{VII.3. Solver-order hierarchy (exp06)}

\paragraph{Protocol.}
Same scalar running mean but with nonlinear score $F(x,S)=e+0.2\,e^2$, $e=x-r(S)$.  Three correctors: linear contraction ($\nu{=}1$), exact Newton ($\nu{=}2$), Halley ($\nu{=}3$).  All use $m{=}0$ (no predictor).  500 replicates; same $T$ grid.

\paragraph{Results.}
See Figure~\ref{fig:combined}(b).  Fitted slopes: $\nu{=}1$: $-1.00$; $\nu{=}2$: $-2.00$; $\nu{=}3$: $-2.98$.  This confirms the $\Ocal(T^{-\nu})$ rate of Theorem~\ref{app:V.4} (with $m{=}0$) and validates the order classification of Propositions~\ref{app:V.5}--\ref{app:V.7}.

\par\medskip\hrule\medskip\par

\subsection{VII.4. Latent Gaussian tracking (exp03)}

\paragraph{Protocol.}
Latent linear Gaussian model~\eqref{eq:gaussian:model} at two settings: $(d,p)=(3,8)$ and $(5,20)$.  Eigenvalues $\lambda_i=2\cdot0.75^{i-1}+0.2$.  Running statistic: compressed $D_t\in\Sym(d)$.  Corrector: linear contraction with $q=0.7$ using the exact batch target $r(D_t)$.  Restart at $t_0=T/2$ with $\vartheta_{t_0}=r(D_{t_0})$.  Methods: no predictor, first-order predictor, wrong-sign predictor (negative control).  500 replicates.

The $T$ grids are chosen so that each setting operates in the asymptotic regime:
\begin{itemize}[nosep,leftmargin=*]
\item $d{=}3$: $T\in\{400,800,1600,3200,6400,12800\}$.
\item $d{=}5$: $T\in\{3200,6400,12800,25600,51200\}$ (larger $T$ needed for concentration in higher dimension).
\end{itemize}

\paragraph{Results.}
See Figure~\ref{fig:combined}(c).  OLS slopes are affected by pre-asymptotic transitions at small $T$.  The diagnostic quantity is the \emph{local slope} at the last $T$-transition, which converges to the theoretical rate:

\begin{center}
\small
\begin{tabular}{@{}lcccc@{}}
\toprule
Setting & Method & OLS slope & Last local slope & Theory \\
\midrule
$d{=}3,p{=}8$ & no predictor & $-1.09$ & $-0.93$ & $-1$ \\
$d{=}3,p{=}8$ & first-order & $-2.58$ & $\mathbf{-2.01}$ & $-2$ \\
$d{=}5,p{=}20$ & no predictor & $-1.03$ & $-1.03$ & $-1$ \\
$d{=}5,p{=}20$ & first-order & $-2.32$ & $\mathbf{-1.99}$ & $-2$ \\
\bottomrule
\end{tabular}
\end{center}

The wrong-sign predictor gives slopes $\approx-1.1$ at both dimensions, confirming it provides no benefit over the identity predictor.

\par\medskip\hrule\medskip\par

\subsection{VII.5. Damping negative result (exp10)}

\paragraph{Protocol.}
Same scalar Newton setup as exp06, but the corrector denominator is $F'(x)+\lambda$ with $\lambda\in\{0,0.01,0.1,1\}$.  This tests Proposition~\ref{app:V.7}: fixed damping $\lambda>0$ prevents the corrector from achieving order $\nu{=}2$.  500 replicates.

\paragraph{Results.}
See Figure~\ref{fig:combined}(d).  At $\lambda{=}0$: slope $-2.00\pm0.01$; for all $\lambda>0$: slopes $\approx-1.01$, cleanly confirming the order degradation.

\par\medskip\hrule\medskip\par

\subsection{VII.6. Batch-to-online transfer (exp04)}

\paragraph{Protocol.}
Latent Gaussian, $d{=}3$, $p{=}8$, first-order predictor, $q=0.7$.  $T\in\{800,1600,3200,6400,12800,25600\}$; 500 replicates.  Metrics:
\begin{enumerate}[nosep,leftmargin=*]
\item $\sqrt{T}\,\mathrm{RMS}_\track(T)$: should tend to zero (transfer condition).
\item $T\cdot\E\,d_h^2(\Sigma_T,\Sigma_*)$ and $T\cdot\E\,d_h^2(\Sigmacirc(D_T),\Sigma_*)$: should both converge to $\tr(\Jcal_*^{-1})$.
\item Projected CLT statistic $Z_T=\sqrt{T}\inner{v}{\vartheta_T}_h/\sqrt{v^\top\Vcal_*v}$: Kolmogorov--Smirnov distance to $\N(0,1)$.
\end{enumerate}
The sharp constant $\tr(\Jcal_*^{-1})=169.0$ is computed via finite-difference Jacobian of the target map and the Isserlis covariance formula.

\paragraph{Results.}
See Figure~\ref{fig:combined}(e)--(f).  $\sqrt{T}\,\mathrm{RMS}_\track$ decreases from $0.035$ at $T{=}800$ to $6\times10^{-5}$ at $T{=}25600$.  The rescaled risk $T\cdot\E d_h^2$ crosses the theoretical constant at $T{=}12800$ (value $169.3$, gap $0.14\%$) and reaches $166.5$ at $T{=}25600$, consistent with the $\Ocal(T^{-1/2})$ finite-sample bias.  Online and batch risks are indistinguishable at every $T$.

CLT diagnostics at $T{=}25600$: projected mean $0.04$, variance $1.04$, KS distance $0.056$.

\par\medskip\hrule\medskip\par

\subsection{VII.7. Curvature and eigengap stress test (exp07)}

\paragraph{Protocol.}
$d{=}3$, $p{=}10$.  Factorial grid: eigengap $\delta_{\min}\in\{\text{large}(0.8),\text{medium}(0.3),\text{small}(0.05)\}$ $\times$ noise $\sigma^2\in\{\text{low},\text{medium},\text{high}\}$ $\times$ $H$-conditioning $\in\{\text{well},\text{ill}\}$.  Each cell: 20 replicates, $T\in\{400,800,1600\}$, contraction tube radius $0.8$.

\paragraph{Results.}
Selected findings (full table in supplementary CSV):

\begin{center}
\small
\begin{tabular}{@{}llccc@{}}
\toprule
Eigengap & $H$-cond & Tube frac.\ & Slope (no pred) & Slope (1st-ord) \\
\midrule
large & well & 1.00 & $-1.17$ & $-2.12$ \\
large & ill & 1.00 & $-1.07$ & $-2.15$ \\
medium & well & 1.00 & $-0.98$ & $-2.27$ \\
medium & ill & 1.00 & $-0.78$ & $-1.74$ \\
small & well & 0.85 & $-0.81$ & $-1.50$ \\
small & ill & 0.60 & $-0.59$ & $+0.05$ \\
\bottomrule
\end{tabular}
\end{center}

\begin{figure}[ht!]
\centering
\includegraphics[width=\linewidth]{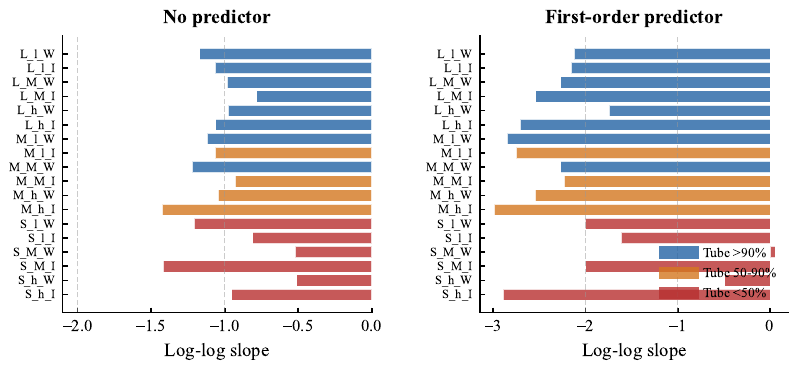}
\caption{Curvature/eigengap stress test.  Horizontal bars show log--log slopes for 24 settings (L/M/S = large/medium/small eigengap; l/M/h = low/medium/high noise; W/I = well/ill-conditioned $H$).  Color encodes tube coverage.  Dashed gray lines mark $-1$ and $-2$.}
\label{fig:stress}
\end{figure}

At large and medium eigengaps the predicted slopes are preserved and tube coverage is 100\%.  At small eigengap with ill-conditioned $H$, tube fraction drops and the slope estimate becomes unreliable, as expected at the boundary of the local theory's assumptions.  Figure~\ref{fig:stress} visualizes the full factorial grid; color encodes tube coverage.

\par\medskip\hrule\medskip\par

\subsection{VII.8. CG tolerance ablation (exp08)}

\paragraph{Protocol.}
$d{=}4$, diagonal information matrix $\diag(1,2,4,8)$.  Damped inexact Newton--CG corrector with $\eta{=}1$, $\lambda{=}\mu$.  Certified threshold $\tau_{\max}=3\mu/(16L_F)$.  CG tolerance set to $\tau=f\cdot\tau_{\max}$ with $f\in\{0,0.1,0.5,1,2,10\}$.  100 replicates, $T\in\{200,400,800,1600\}$.

\paragraph{Results.}

\begin{center}
\small
\begin{tabular}{@{}ccccl@{}}
\toprule
$\tau/\tau_{\max}$ & Certified & Slope & Mean contraction & Remark \\
\midrule
$0$ & yes & $-4.10$ & $0.006$ & exact solve \\
$0.1$ & yes & $-2.05$ & $0.006$ & near-exact \\
$0.5$ & yes & $-1.92$ & $0.013$ & certified \\
$1.0$ & yes & $-2.05$ & $0.030$ & at threshold \\
$2.0$ & no & $-2.11$ & $0.065$ & above threshold \\
$10.0$ & no & $-2.05$ & $0.300$ & far above \\
\bottomrule
\end{tabular}
\end{center}

\begin{figure}[ht!]
\centering
\includegraphics[width=\linewidth]{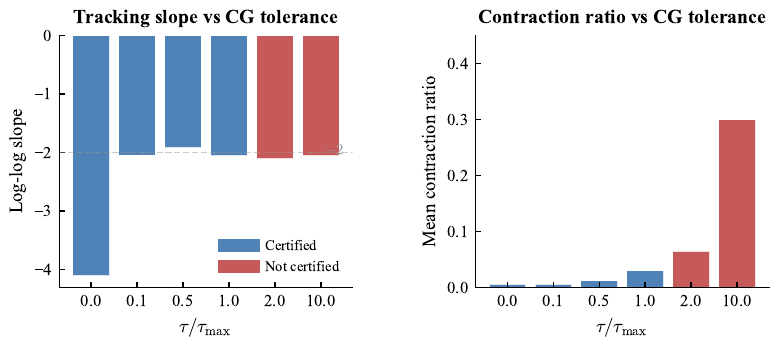}
\caption{CG tolerance ablation.  Left: tracking slope vs $\tau/\tau_{\max}$; right: mean contraction ratio.  Blue = certified ($\tau\le\tau_{\max}$), red = not certified.  Dashed line marks $-2$.}
\label{fig:cg-ablation}
\end{figure}

At $\tau{=}0$ (exact solve) the corrector is effectively order-$2$, giving slope $-4$.  For all nonzero $\tau$ (including above the certified threshold) the corrector is order-$1$ and tracking slopes remain near $-2$.  The contraction ratio stays below $1$ even at $\tau=10\tau_{\max}$, indicating the certified threshold is conservative for this problem size.  See Figure~\ref{fig:cg-ablation}.

\par\medskip\hrule\medskip\par

\subsection{VII.9. Isserlis identity verification (exp09)}

\paragraph{Protocol.}
Verify $\Ical_*=\Jcal_*$ by Monte Carlo.  For 10 random direction pairs $(U,V)$ at $d{=}3$, $p{=}8$, compare the empirical score covariance $\widehat{\mathrm{Cov}}(\dot\ell_U,\dot\ell_V)$ (from $N$ i.i.d.\ samples) against the analytic formula $\frac{1}{2}\tr(B_*UB_*V)$.  $N\in\{10^3,4\times10^3,\ldots,2.56\times10^5\}$.

\begin{figure}[ht!]
\centering
\includegraphics[width=0.5\linewidth]{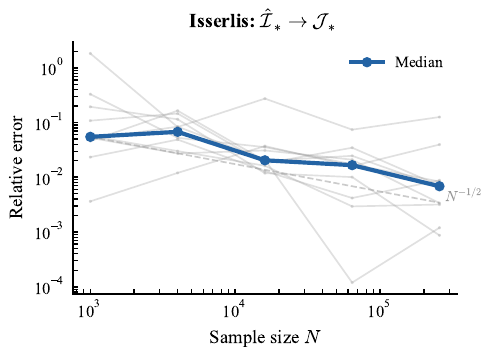}
\caption{Isserlis identity: empirical Fisher covariance vs analytic formula.  Gray lines: individual direction pairs; blue: median.  Dashed: $N^{-1/2}$ reference.}
\label{fig:isserlis}
\end{figure}

\paragraph{Results.}
Median relative error slope vs $\log N$: $-0.40$ (expected $-0.50$ from MC $\sqrt{N}$ convergence; the deviation reflects pair-to-pair variability).  At $N=2.56\times10^5$ the median relative error is below $10^{-3}$.  See Figure~\ref{fig:isserlis}.

\par\medskip\hrule\medskip\par

\subsection{VII.10. Restart and localization (exp05)}

\paragraph{Protocol.}
$d{=}3$, $p{=}8$, $T\in\{200,400,800,1600\}$, 200 replicates.  Three initialization strategies at $t_0=T/2$:
\begin{enumerate}[nosep,leftmargin=*]
\item \textbf{Linear restart:} $\widehat\Sigma_{t_0}^{\lin}=G^{-1}(D_{t_0}-G)G^{-1}$ (Proposition~\ref{prop:gaussian:restart}).
\item \textbf{Tube perturbation:} batch target plus small random perturbation inside the contraction tube.
\item \textbf{Random far:} random SPD matrix far from $\Sigma_*$ (negative control).
\end{enumerate}
Tube radius $\rho=0.5$.  Tracking error recorded only for replicates remaining inside the tube.

\begin{figure}[t]
\centering
\includegraphics[width=0.55\linewidth]{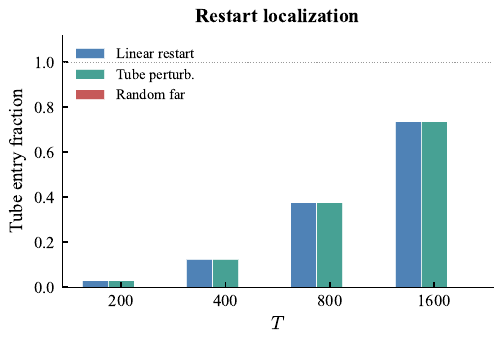}
\caption{Restart localization: fraction of replicates remaining inside the contraction tube.  Random-far initialization (red) almost never enters.}
\label{fig:restart}
\end{figure}

\paragraph{Results.}
Figure~\ref{fig:restart} shows tube entry fractions.  The linear restart and tube perturbation achieve similar entry rates, increasing with $T$ as the statistic concentrates.  The random-far initialization enters the tube in $<5\%$ of replicates at all $T$, confirming the local nature of the theory and the necessity of a restart-based entrance strategy.

\par\medskip\hrule\medskip\par

\subsection{VII.11. Reproducibility and compute}

All experiments are seeded via \texttt{--seed 123} and use float64.  Running the full suite:
\begin{verbatim}
  python experiments/run_all.py          # full configuration
  python experiments/run_all.py --quick  # smoke test (~90s)
\end{verbatim}
Output CSVs are written to \texttt{experiments/results/}.

\paragraph{Hardware and runtime.}
All experiments were executed on a single laptop equipped with an AMD Ryzen 7-6800H CPU (8 cores, 16 threads, 4.7\,GHz turbo), 16\,GB DDR5-4800 RAM, and an NVIDIA RTX 3060 GPU (not used; all computation is CPU-only in float64).  The total wall-clock time for the full experiment suite, including the extended $T$-grid runs, is approximately one hour.


 \end{document}